\documentclass{article}

\PassOptionsToPackage{numbers}{natbib}
\usepackage[final,main]{neurips_2026}

\usepackage[utf8]{inputenc}
\usepackage[T1]{fontenc}

\usepackage{amsmath}
\usepackage{amssymb}
\usepackage{amsfonts}
\usepackage{mathtools}
\usepackage{amsthm}
\usepackage{graphicx}
\usepackage{booktabs}
\usepackage{array}
\usepackage{tabularx}
\usepackage{makecell}
\usepackage{multirow}
\usepackage{longtable}
\usepackage{rotating}
\usepackage{subcaption}
\usepackage{caption}
\usepackage{placeins}
\usepackage{booktabs}

\newcommand{\best}[1]{\textbf{\underline{#1}}}
\newcommand{\tie}[1]{\textbf{#1}}
\newcommand{\statentry}[2]{#1\textpm#2}
\usepackage{standalone}
\usepackage{pgfplots}
\pgfplotsset{compat=1.18}
\usetikzlibrary{calc, positioning, arrows.meta}

\usepackage{algorithm}
\usepackage{algpseudocode}

\usepackage{nicefrac}
\usepackage{microtype}
\usepackage{xcolor}
\usepackage{enumitem}
\usepackage{soul}
\usepackage{textcomp}
\usepackage{etoc}

\usepackage{url}
\usepackage{hyperref}
\usepackage[capitalize,noabbrev]{cleveref}
\hypersetup{hidelinks}

\makeatletter
\providecommand{\theHALG@line}{}
\renewcommand{\theHALG@line}{\thealgorithm.\arabic{ALG@line}}
\makeatother

\definecolor{steelblue}{RGB}{70, 130, 180}
\definecolor{royalblue}{RGB}{65, 105, 225}

\DeclareMathOperator{\softmax}{softmax}
\DeclareMathOperator{\KL}{KL}

\theoremstyle{plain}
\newtheorem{theorem}{Theorem}[section]
\newtheorem{proposition}[theorem]{Proposition}

\theoremstyle{definition}

\theoremstyle{remark}

\title{FRInGe: Distribution-Space Integrated Gradients with Fisher--Rao Geometry}

\author{
   Gabriele Martino \\
  Faculty of Computer Science\\
  Doctoral School Computer Science\\
  University of Vienna, Vienna, Austria \\
  \texttt{gabriele.martino@univie.ac.at} \\
  \And
  Sebastian Tschiatschek \\
  Faculty of Computer Science\\
  ds:UniVie \\
  University of Vienna, Vienna, Austria \\
  \texttt{sebastian.tschiatschek@univie.ac.at} \\
}

\begin{document}

\etocdepthtag.toc{mainmatter}
\maketitle
\begin{abstract}
Although gradient-based attribution methods are model-faithful and scalable, a prominent variant, Integrated Gradients (IG), suffers from high sensitivity to design choices, particularly in baseline selection, path formulation, and discretization, as well as gradient saturation. To address these limitations, we propose Fisher--Rao Integrated Gradients (FRInGe), which defines both the reference and
integration paths in the predictive distribution space. FRInGe replaces input baselines with a maximum-entropy predictive reference and follows a Fisher--Rao geodesic on the probability simplex. The
corresponding input-space trajectory is realized through the pullback Fisher metric and stabilized by
Kullback--Leibler (KL) and Euclidean trust regions; attributions are obtained by integrating input gradients
along this trajectory. We further derive FRInGe-B, a variant that binarizes the predictive probability distribution, removing the computationally expensive Fisher products of FRInGe. Across six ImageNet architectures, FRInGe
most clearly improves Magnitude Aligned Scoring (MAS), while remaining competitive on perturbation
area-under-the-curve (AUC) metrics and infidelity.
\end{abstract}

\section{Introduction}
\label{introduction}

Deep neural networks are increasingly deployed in high-stakes domains such as medical imaging and robotics \cite{SAYRES2019552, saporta2022benchmarking, yu2023identifying},
yet the rationale behind individual predictions often remains opaque \cite{nauta2023anecdotal}. This opacity has motivated research in
explainable AI (XAI), where \emph{feature attribution} methods seek to identify the input components most
relevant to a model's output \cite{montavon2018methods}. Among these, gradient-based approaches are widely adopted because
they leverage exact model gradients directly and can be implemented efficiently via
automatic differentiation.

A prominent representative of this family is Integrated Gradients (IG) \cite{sundararajan2017axiomatic}.
Given an input $x$ and a baseline $x'$, IG attributes importance to each feature by integrating the gradient
of the model output along a path connecting $x'$ to $x$. Under mild regularity assumptions, this
construction satisfies desirable properties such as implementation invariance, sensitivity, and completeness \cite{sundararajan2017axiomatic}.

Despite its theoretical appeal, IG can be sensitive to several design choices. First, the \emph{baseline}
is intended to represent an absence of information, but common choices such as an all-zero image or noise
can induce qualitatively different explanations \cite{sturmfels2020visualizing}. Second, the
\emph{integration path}, typically chosen as a straight line in input space, can traverse regions where
gradients are poorly conditioned or dominated by artifacts \cite{bardhan2024constructing,
kapishnikov2021guided, rahman2022geometrically, zaher2024manifold}. Recent work has therefore explored
replacing straight-line paths with geometry-aware trajectories \cite{salek2025using}. Third, IG approximates
the path integral by a finite Riemann sum, and the resulting discretization error can become substantial
when gradients vary sharply along the path. These issues are further compounded by \emph{saturation}: along
large portions of the standard path, the model output may change rapidly early on and then plateau, so
informative gradients are concentrated in a narrow segment while later integration steps contribute a mostly
noisy or weakly informative signal \cite{walker2024integrated}.

In this paper, we address these issues by defining both the reference and the interpolation schedule in
\emph{predictive distribution space}. Rather than selecting a heuristic input-space baseline, we use a
\emph{distributional} reference point: the maximum-entropy predictive distribution, which
operationalizes ``no information`` directly at the model output level. We then interpolate from the model's
predictive distribution for a given input to this reference along a Fisher--Rao geodesic on the probability simplex. For
categorical predictions---distributions retaining a separate probability for every class---this Fisher--Rao geometry is canonical rather than a tunable design choice: it is uniquely determined by fundamental statistical principles. Formally, up to an overall
scale, it is the unique Riemannian metric invariant under statistically sufficient transformations
\cite{cencov1982statistical,amari2016information}. Moreover, assuming strictly positive endpoints (as guaranteed by our maximum-entropy reference and softmax predictive distributions), the square-root great-circle construction gives the unique constant-speed shortest path. To realize
this trajectory in input space, we use the Fisher--Rao \emph{pullback metric}, which induces a local
geometry on the input space through the model's mapping from inputs to predictive distributions. This yields a
model-aware direction field that avoids committing to a Euclidean straight-line interpolation. To discretize and integrate this path stably, we further use a trust-region step rule combining KL control in predictive space
with a Euclidean step-size control in input space. Feature attributions are then computed by integrating input gradients
along the resulting path.

To the best of our knowledge, FRInGe is the first attribution method that (i) defines the IG reference in
predictive distribution space, (ii) constructs the integration schedule as a Fisher--Rao geodesic in that
space, and (iii) induces the corresponding input-space trajectory via the Fisher--Rao pullback metric. In
contrast to prior path variants that design trajectories directly in input space, or methods that rely on an
explicit generative model of the data manifold, our path is defined by the geometry of the
model's predictive distribution and then mapped back to input space. Defining the reference in predictive
distribution space also reduces dependence on data-type-specific baseline design.

We additionally derive FRInGe-B, a specialization that collapses the categorical prediction into the target class versus all remaining classes. Whereas full FRInGe must iteratively solve a linear system using model-dependent Fisher matrix-vector products at each waypoint, FRInGe-B induces a rank-one pullback metric that eliminates these costly inner-loop evaluations, trading within-rest class geometry for a substantial speedup and competitive attribution quality.

\begin{figure}[t]
    \centering
    \includegraphics[scale=0.7]{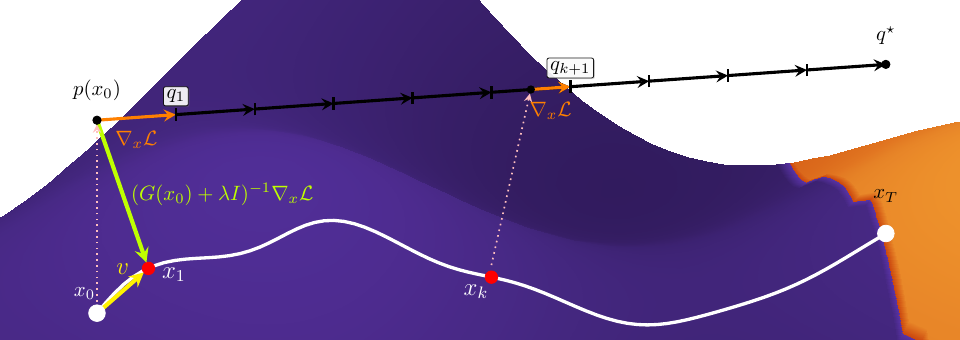}
    \caption{FRInGe constructs a Fisher--Rao geodesic from the initial prediction to a maximum-entropy reference in predictive distribution space and computes an input trajectory whose predictions approximately track its waypoints.}
    \label{fig:fringe_view}
\end{figure}
\section{Related Work}

\subsection{Baseline choice in Integrated Gradients}
A recurring practical challenge in IG is that the baseline is not a benign
implementation detail: different baselines can yield qualitatively different explanations. Prior work has therefore sought to reduce baseline arbitrariness either by analyzing baseline sensitivity or by replacing a single reference with a distribution or a set of baselines. Sturmfels et al.~\cite{sturmfels2020visualizing} provide a detailed analysis of baseline sensitivity and argue that baselines encode different notions of ``missingness,'' motivating the use of \emph{baseline distributions} rather than a single reference input. Expected Gradients operationalizes this idea by averaging IG over baselines sampled from a reference distribution, typically the data distribution, thereby reducing dependence on any specific baseline and improving stability in practice~\cite{erion2021improving}. A complementary line of work connects baselines to Shapley-style attributions. Sundararajan and Najmi~\cite{pmlr-v119-sundararajan20b} formalize how different Shapley operationalizations correspond to different reference conventions and propose \emph{Baseline Shapley} as a principled construction. Building on this perspective, Liu et al.~\cite{liu2023new} propose Shapley Integrated Gradients (SIG), which constructs a \emph{set} of baselines via proportional sampling to better approximate Shapley coalitions. Other approaches tailor baselines to domain semantics; for example, Bardhan et al.~\cite{bardhan2024constructing} show in particle-physics settings that averaging baselines from background events can yield more meaningful attributions than a zero reference. Recent work further argues that candidate baselines should not be treated as equally appropriate and proposes selecting or weighting them according to suitability criteria~\cite{tuan2025weighted}. Overall, these approaches mitigate baseline dependence by averaging across baselines, constructing baseline sets, or weighting candidate references. In contrast, Tan et al.~\cite{tan2023maximum} argue that a principled notion of ''no information'' corresponds to \emph{maximum entropy} in the model's output distribution. We adopt this perspective by defining the reference point in \emph{predictive distribution space}, thereby avoiding the need to specify a heuristic reference directly in input space.

\subsection{Finding a better path}
Beyond the baseline, a substantial body of work improves IG by modifying the integration path to avoid regions where gradients are unstable, saturated, or dominated by artifacts. Guided Integrated Gradients (GuidedIG)~\cite{kapishnikov2021guided} attributes much of IG's noise to the fixed Euclidean straight-line path and proposes an adaptive, model-dependent trajectory that advances preferentially along features with small partial derivatives, helping avoid high-curvature and saturation-dominated segments. Blur Integrated Gradients~\cite{xu2020attribution} replaces linear interpolation with a scale-space path defined by progressively blurring the input, yielding cleaner and more structured attributions through controlled information removal. Other variants modify the path more aggressively to counteract saturation or align the trajectory with decision-relevant directions. IG$^2$~\cite{zhuo2024ig} constructs a counterfactual-guided integration path
and baseline through iterative representation matching, while Adversarial Gradient Integration (AGI)~\cite{pan2021explaining} integrates gradients along adversarially constructed trajectories that follow steepest-ascent directions in input space; related work further refines such adversarial paths to better align gradient accumulation with decision boundaries~\cite{ren2025improving}. More explicitly geometric approaches have also emerged. Geometrically Guided Integrated Gradients (GGIG)~\cite{rahman2022geometrically} perturbs interpolants by ascending the loss landscape to emphasize discriminative directions, while Manifold Integrated Gradients (MIG)~\cite{zaher2024manifold} integrates along geodesics of a learned data manifold induced by a deep generative model. MIG consequently depends on both the classifier and an auxiliary generator, whereas FRInGe is a model-only post-hoc method. Most closely related to our perspective, Salek and Enguehard~\cite{salek2025using} propose Geodesic Integrated Gradients (GeoIG), which replaces the Euclidean straight-line interpolation with geodesics in input space under a model-induced metric defined from the Jacobian, approximated either via shortest paths on a $k$-NN graph or via an energy-based sampling procedure. In contrast, our method does not define the path directly in input space: it defines both the reference and the interpolation schedule in \emph{predictive distribution space} under Fisher--Rao geometry, and then realizes the corresponding input-space trajectory through the pullback metric together with trust-region stabilization. Because of this close conceptual overlap, geometry-aware path methods such as GGIG and GeoIG form especially important empirical comparators for FRInGe.

\begin{figure}[t]
    \centering
    \includegraphics[width=\linewidth]{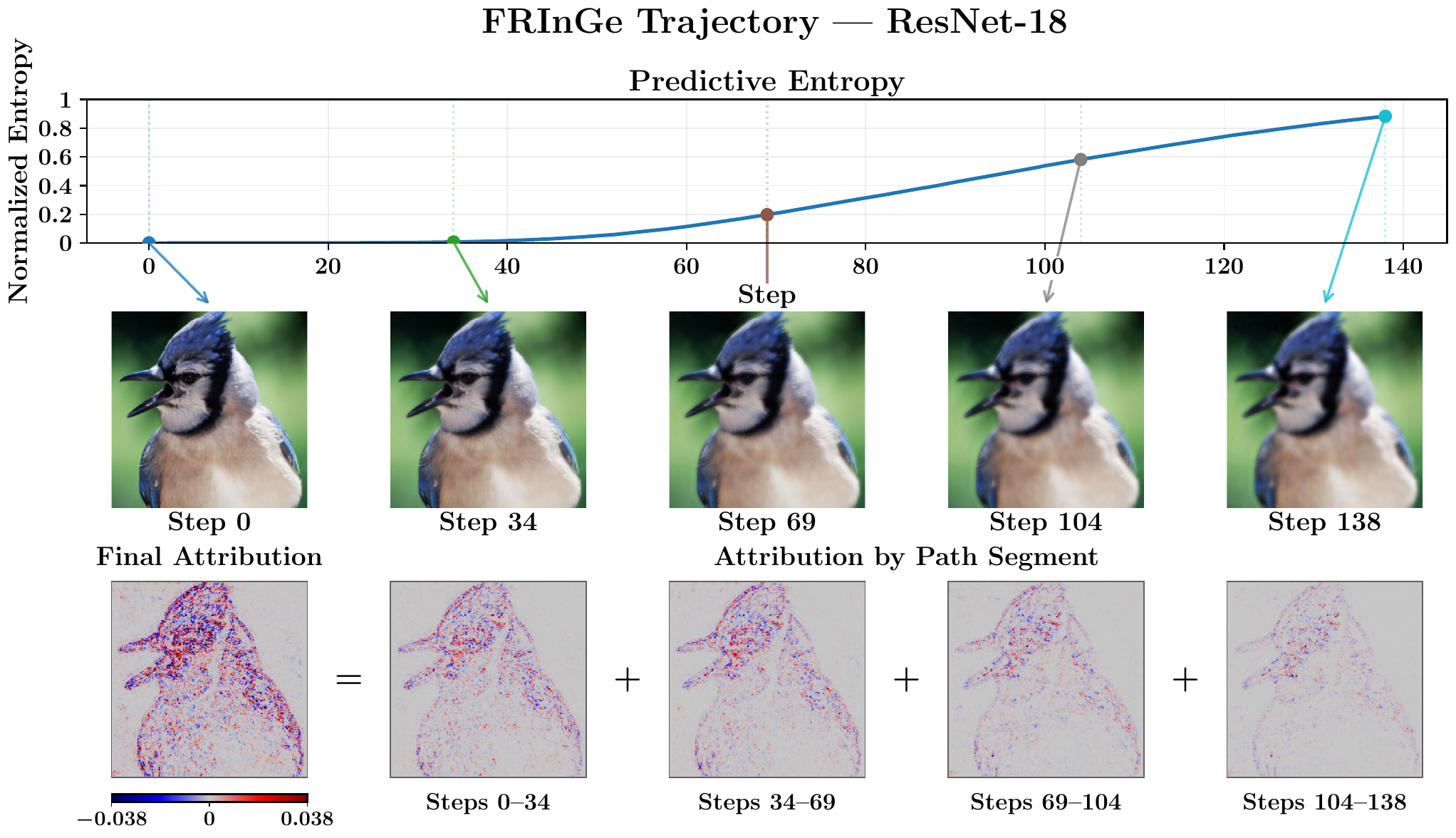}
  \caption{Illustrative FRInGe trajectory on ResNet-18 (protocol in Appendix~\ref{app:ablation}).
\textit{Top}: normalized predictive entropy. \textit{Middle}: five intermediate inputs.
\textit{Bottom}: final attribution (left) and signed contributions accumulated over four consecutive path segments.
All maps share one symmetric colour scale, with grey denoting zero.}
\label{fig:intermediate_steps}
\end{figure}

\section{Background}
\label{sec:background}

Let $F\colon \mathbb{R}^n \to \mathbb{R}^C$ be a differentiable model, let $x\in\mathbb{R}^n$ denote an input, let $C$ be the number of classes, and let $x'\in\mathbb{R}^n$ be a baseline in input space. Fix a target class $t\in\{1,\dots,C\}$ and define the corresponding scalar score $F^t:\mathbb{R}^n\to\mathbb{R}$, for example the $t$-th logit.

\paragraph{Integrated Gradients \cite{sundararajan2017axiomatic}.}
Let $\gamma\colon [0,1]\to\mathbb{R}^n$ be a $C^1$ curve with $\gamma(0)=x'$ and $\gamma(1)=x$. IG along $\gamma$ is defined component-wise by
\begin{equation}
\big[\mathrm{IG}_\gamma(x;x')\big]_i
= \int_{0}^{1}
\frac{\partial F^t}{\partial x_i}\!\big(\gamma(\alpha)\big)\,
\frac{d\gamma_i(\alpha)}{d\alpha}\, d\alpha,
\label{eq:ig_component}
\end{equation}
or equivalently in vector form $ \mathrm{IG}_\gamma(x;x')
= \int_{0}^{1} \nabla F^t\!\big(\gamma(\alpha)\big)\odot\gamma'(\alpha)\, d\alpha.$
By the fundamental theorem for line integrals, IG satisfies \emph{completeness}:
\begin{equation}
\sum_{i=1}^{n} \big[\mathrm{IG}_\gamma(x;x')\big]_i
= F^t(x)-F^t(x').
\label{eq:completeness}
\end{equation}

In practice, the integral is approximated numerically, and the resulting discretization error can become severe when gradients vary sharply along the path or when the realized trajectory undergoes abrupt directional changes between successive steps.

\paragraph{Fisher--Rao geometry and geodesics on the simplex.}
Let $F(x)\in\mathbb{R}^C$ denote the logits and let $ p(x)=\softmax(F(x))\in\Delta^{C-1}$
be the predictive distribution, where $\Delta^{C-1}=\bigl\{p\in\mathbb{R}^C_{\ge 0}:\sum_{c=1}^C p_c=1\bigr\}.$ For categorical distributions, the Fisher--Rao geometry is conveniently represented through the elementwise square-root map $\psi(p)=\sqrt{p}$
which embeds the simplex into the positive orthant of the unit sphere $ \mathbb{S}^{C-1}_+=\bigl\{u\in\mathbb{R}^C_{\ge 0}:\|u\|_2=1\bigr\}.$
For this embedding, Fisher--Rao geodesics on the simplex correspond to great-circle geodesics on the sphere, up to a constant rescaling of arc length. Hence, given two predictive distributions $p_0,p_1\in\Delta^{C-1}$, we define
\[
\psi_0=\sqrt{p_0},\qquad \psi_1=\sqrt{p_1},
\qquad
\theta=\arccos(\langle \psi_0,\psi_1\rangle).
\]
So a constant-speed shortest path on the sphere from $\psi_0$ to $\psi_1$ is given by the spherical linear interpolation \cite{shoemake1985animating}:
\begin{equation}
s(\alpha)
=
\frac{\sin((1-\alpha)\theta)}{\sin\theta}\,\psi_0
+
\frac{\sin(\alpha\theta)}{\sin\theta}\,\psi_1,
\qquad \alpha\in[0,1].
\label{eq:slerp_simple}
\end{equation}
Mapping this path back to the simplex yields $p(\alpha)=s(\alpha)\odot s(\alpha)$, which is the Fisher--Rao geodesic between $p_0$ and $p_1$. The corresponding Fisher--Rao distance is $d_{\mathrm{FR}}(p_0,p_1)=2\,\arccos\!\big(\langle \sqrt{p_0},\sqrt{p_1}\rangle\big).$

\paragraph{Pullback Fisher metric in input space.}
Let $J_F(x)=\partial F(x)/\partial x \in \mathbb{R}^{C\times n}$ denote the Jacobian of logits with respect to the input. For a softmax model, the Fisher information in logit coordinates is
\begin{equation}
S(x) = \mathrm{diag}\!\big(p(x)\big) - p(x)p(x)^\top \in \mathbb{R}^{C\times C}.
\label{eq:fisher_logits}
\end{equation}
The induced \emph{pullback Fisher metric} \cite{arvanitidis2021pulling} in the input space is
\begin{equation}
G(x) = J_F(x)^\top\, S(x)\, J_F(x) \in \mathbb{R}^{n\times n}.
\label{eq:pullback_metric}
\end{equation}
This matrix is positive semidefinite and may be rank-deficient or ill-conditioned. It provides a local quadratic approximation of the KL divergence between nearby predictive distributions:
\begin{equation}
\mathrm{KL} \big(p(x+\mathrm{d}x)\,\|\,p(x)\big)\approx \tfrac12\,\mathrm{d}x^\top G(x)\,\mathrm{d}x.
\label{eq:kl_quadratic}
\end{equation}

\section{Method}
\label{sec:method}



Given an input $x$ and a fixed target class $t$, FRInGe computes attributions
of the target logit $F^t$ along an input trajectory guided by predictive
geometry. We first choose a reference distribution $q^\star$ and construct a
Fisher--Rao geodesic connecting the initial prediction $p(x)$ to this
reference. We discretize that geodesic into predictive waypoints and define
a spherical alignment loss to guide their approximate tracking in input
space. We then derive damped natural-gradient directions using the pullback
Fisher tensor. Next, we introduce a KL-based step-size bound using a local
quadratic approximation and couple the number of tracking steps to the
predictive budget. A complementary Euclidean displacement cap controls the
lengths of the input-space segments used for numerical integration. For the
image models studied here, we further regularize spatial smoothness of both the update directions and the inputs. Finally, we use trapezoidal quadrature to accumulate target-logit gradients along the realized trajectory, where completeness holds relative to the reached endpoint $x_T$ and is evaluated separately from whether $p(x_T)$ reaches $q^\star$ (Section~\ref{sec:attr} and Appendix~\ref{app:euc_tr_quant}).

Unless stated otherwise, FRInGe denotes the full $C$-class formulation with both smoothness terms active (penalizing high-frequency variation in the update direction and in the current input). Variants disabling either term are reported as ablations.

\paragraph{Reference distribution.}
\label{sec:refdist}
The reference distribution $q^\star\in\Delta^{C-1}$ represents a problem-dependent notion of ``no evidence'' or ``uninformative prediction.'' In this work, we adopt the maximum-entropy choice, i.e.,
\begin{equation}
q^\star = \tfrac{1}{C}\mathbf{1},
\label{eq:maxent_ref}
\end{equation}
motivated by the maximum entropy principle \cite{tan2023maximum}. This yields a model-agnostic reference specified directly in predictive distribution space. Uniformity represents no predictive preference among classes. Although we use the uniform reference, other choices may be appropriate depending on domain knowledge or experimental design. Beyond the interpretation as ``no evidence'', the reference can also encode a targeted contrast, for example by favoring a selected alternative class or assigning equal maximal probability to two selected classes to represent a tie in the classification prediction.

\paragraph{Inducing an input-space trajectory that tracks predictive waypoints.}
\label{sec:input_tracking}
Specifying a reference in predictive distribution space avoids the need to prescribe a particular input baseline. Computing path-integrated attributions requires an input trajectory
along which target-logit gradients can be evaluated. We therefore seek an input-space trajectory whose predictions track successive waypoints along the Fisher--Rao geodesic from $p(x)$ to $q^\star$.

We initialize the input trajectory at the original input, $x_0=x$.
Let $p_0=p(x)$ denote the predictive distribution of the input $x$, and let $s(\alpha)$ be the Fisher--Rao geodesic in square-root coordinates defined in Eq.~\eqref{eq:slerp_simple}. For $\alpha_k=k/T$, $k=0,\dots,T$, we define the waypoint sequence $s_k=s(\alpha_k)$. Defining $q_k=s_k\odot s_k$, we have $q_0=p_0$ and $q_T=q^\star$. At state $x_k$, $k=0,\dots,T-1$,
we target the \emph{next} waypoint using the spherical alignment loss
\begin{equation}
\mathcal{L}_{k+1}(x)
=
1-\Big\langle \sqrt{p(x)},\, s_{k+1} \Big\rangle,
\label{eq:spherical_loss}
\end{equation}

which is minimized when the current predictive distribution aligns with that
target. This indexing ensures that $s_0=\sqrt{p(x_0)}$ is the starting
prediction, whereas $s_T=\sqrt{q^\star}$ is the final predictive-space target. How we choose the number of steps $T$ is described in the paragraphs below.

A Euclidean descent direction for $\mathcal{L}_{k+1}$ would track the predictive
waypoints, but it ignores the model-induced geometry and can enter poorly
conditioned input-space regions. We instead use the pullback Fisher metric to
obtain geometry-aware directions, as described in the paragraph below; this optimization only realizes the path,
while attribution is still computed by path-integrated gradients.

\paragraph{Geometry-aware steps via a damped natural gradient.}
\label{sec:natural_grad}
Let $G(x_k)$ denote the pullback Fisher tensor defined in
Eq.~\eqref{eq:pullback_metric}. Since it may be singular,
it defines the Fisher seminorm
\begin{equation}
\|v\|_{G(x_k)}^2 := v^\top G(x_k)\,v.
\label{eq:fr_norm}
\end{equation}
At state $x_k$, let
\[
g_k=\nabla_x \mathcal{L}_{k+1}(x)\rvert_{x=x_k}.
\]
For a candidate displacement $-v$, the first-order change in the
tracking loss is $-g_k^\top v$, while
$\tfrac12 v^\top G(x_k)v$ approximates the KL divergence induced
by the displacement, cf.~Eq.~\eqref{eq:kl_quadratic}.
We therefore balance predicted loss reduction against local
predictive change using the natural-gradient construction
\cite{amari1998natural}.

Because $G(x_k)$ may be singular or ill-conditioned, we additionally
penalize the Euclidean size of the displacement through Tikhonov
damping \cite{martens2020new}. Without any additional regularization, the damped natural-gradient
vector is defined, for $\lambda>0$, by
\begin{equation}
v_k
=
\arg\min_{v\in\mathbb{R}^n}
\left\{
\frac{1}{2}v^\top G(x_k)\,v
+ \frac{\lambda}{2}\|v\|_2^2
- g_k^\top v\right\},
\label{eq:damped_ng_obj}
\end{equation}
whose first-order optimality condition is
\begin{equation}
\big(G(x_k)+\lambda I\big)\,v_k = g_k.
\label{eq:damped_solve}
\end{equation}
FRInGe extends this system with both spatial regularization terms,
as specified below in Eq.~\eqref{eq:final_linear}. We solve the resulting
system using conjugate gradients (CG), evaluating Fisher
matrix-vector products implicitly through Jacobian-Vector products (JVPs) and Vector-Jacobian products (VJPs).
Implementation details are given in Appendix~\ref{app:damping}.

The resulting update has the form
\begin{equation}
x_{k+1} = x_k - \eta_k v_k,
\label{eq:update}
\end{equation}
where $\eta_k$ is selected using the step-size bounds below.

\paragraph{KL-based step-size bound.}
\label{sec:trust_region}
Following the use of KL constraints in natural-gradient and trust-region
methods \cite{martens2020new,schulman2015trust}, we introduce a
hyperparameter $\tau>0$ specifying the desired upper bound on the
KL divergence between successive predictive distributions:
\[
\KL\!\big(p(x_k-\eta v_k)\,\|\,p(x_k)\big)\leq\tau.
\]
Thus, $\tau$ controls the permitted predictive change per update.
Smaller values prescribe more conservative updates, whereas larger
values permit greater predictive changes but may reduce the accuracy
of the local quadratic approximation.
Applying this constraint to the quadratic model in
Eq.~\eqref{eq:kl_quadratic} gives the approximate step-size cap
\begin{equation}
\eta_{\mathrm{KL}}
=
\sqrt{\frac{2\tau}
{\|v_k\|_{G(x_k)}^2+\varepsilon}},
\label{eq:eta_kl}
\end{equation}
where $\varepsilon>0$ prevents division by zero.
For a fixed search direction, this rule limits the predictive change
estimated by the local Fisher quadratic model.
Because the approximation is local, the resulting cap does not
by itself guarantee satisfaction of the actual KL constraint.

\paragraph{Matching waypoint spacing to the KL budget.}
\label{sec:synchronization}
The KL budget $\tau$ controls the permitted predictive change per
update, while the number of tracking steps $T$ determines the spacing
of the predictive waypoints. We couple these choices so that the
distance between consecutive waypoints is consistent with the
predictive change allowed by the local step-size rule.

Let $D_{\mathrm{total}} = d_{\mathrm{FR}}\big(p(x),q^\star\big)$
denote the Fisher--Rao distance from the initial prediction to the
reference distribution. Under the local quadratic approximation,
a KL budget of $\tau$ corresponds to a nominal Fisher--Rao
displacement of $\sqrt{2\tau}$. For $D_{\mathrm{total}}>0$, we
therefore set
\begin{equation}
T
=
\left\lceil
\frac{D_{\mathrm{total}}}{\sqrt{2\tau}}
\right\rceil.
\label{eq:step_sync}
\end{equation}
This choice couples the waypoint discretization to the KL budget,
allocating more tracking steps when the reference distribution is
farther from the initial prediction or when a smaller predictive
change is permitted per update.

\paragraph{Euclidean displacement cap.}
The KL-based step-size rule controls local predictive change but
does not directly bound the change in the input space.
In directions of low predictive sensitivity, it may permit large
input updates, across which the gradients used for attribution
can vary substantially. To control the spacing of the input-space
samples used for numerical integration, we introduce a hyperparameter
$\delta_{\mathrm{euc}}>0$ specifying the maximum Euclidean
displacement per update. This gives the additional step-size bound
\begin{equation}
\eta_{\mathrm{Euc}}
=
\frac{\delta_{\mathrm{euc}}}{\|v_k\|_2+\varepsilon},
\label{eq:euclidean_step}
\end{equation}
where also here $\varepsilon>0$ prevents division by zero. The resulting update satisfies
$\|x_{k+1}-x_k\|_2\leq\delta_{\mathrm{euc}}$,
thereby bounding the length of each segment of the integration path.
For a fixed number of updates, a restrictive displacement cap may
reduce progress toward the predictive waypoints, creating a trade-off
between tracking progress and the resolution of the trajectory
used for attribution. Appendices~\ref{app:euc_tr_geometry} and~\ref{app:euc_tr_quant}
analyze how this cap affects waypoint tracking and numerical
integration error.

\paragraph{Bounding the step size.} We combine the Euclidean displacement bound with the predictive-space bound
and a maximum step size $\eta_{\max}>0$ to select
\begin{equation}
\eta_k
=
\min\!\left\{
\eta_{\max},\,\eta_{\mathrm{KL}},\,\eta_{\mathrm{Euc}}
\right\}.
\label{eq:combined_step}
\end{equation}

\paragraph{Spatial regularization of the input trajectory.}
\label{sec:smoothness}
Adversarial examples demonstrate that neural-network
predictions can be altered through visually imperceptible
perturbations \cite{szegedy2013intriguing}, while convolutional
networks can exploit high-frequency image components that are
weakly aligned with human perception \cite{wang2020high}.
These observations motivate introducing a spatial smoothness
prior into the predictive-waypoint tracking procedure.

Let $L$ be a linear spatial differential operator, such as a
discrete Laplacian. We use nonnegative coefficients
$\gamma_{\mathrm{step}},\gamma_{\mathrm{prior}}\geq0$. Combining the pullback Fisher geometry, Euclidean damping, and
spatial regularization yields the FRInGe update direction used in our experiments,
defined by
\begin{equation}
\big(
G(x_k)+\lambda I
+\gamma_{\mathrm{step}}L^\top L
\big)v_k
=
g_k+\gamma_{\mathrm{prior}}L^\top Lx_k.
\label{eq:final_linear}
\end{equation}
The coefficient $\gamma_{\mathrm{step}}$ controls the penalty
$\|Lv_k\|_2^2$ on spatial variation in the update direction.
Defining the input roughness penalty as
$R(x)=\tfrac12\|Lx\|_2^2$, the right-hand side combines the
alignment gradient $g_k$ with
$\gamma_{\mathrm{prior}}\nabla_xR(x_k)$.
The update therefore incorporates both predictive alignment
and smoothness of update directions and inputs. For a symmetric discrete Laplacian, $L^\top L=L^2$ is a
discrete biharmonic operator, corresponding to the
squared-Laplacian smoothness energy
\cite{jacobson2011bounded}.

In Section~\ref{sec:experiments}, we
evaluate FRInGe with both spatial terms active on convolutional neural networks. Figure~\ref{fig:fringe_view} illustrates the overall FRInGe construction.

\paragraph{Attribution computation.}
\label{sec:attr}
We integrate target-logit gradients along the piecewise-linear
input trajectory through $\{x_k\}_{k=0}^{T}$.
Starting from $A_{\mathrm{path}}=0$, we apply the trapezoidal rule
on each segment:
\begin{equation}
\begin{aligned}
A_{\mathrm{path}}
\leftarrow A_{\mathrm{path}}
+
\frac{1}{2}
\Big(\nabla_x F^t(x_k)+\nabla_x F^t(x_{k+1})\Big)
\odot (x_{k+1}-x_k).
\end{aligned}
\label{eq:trapezoid_attr}
\end{equation}
We return $A=-A_{\mathrm{path}}$ to express the attributions
in the conventional endpoint-to-input orientation.
The attributions explain the change in the target logit between the
original input $x$ and the endpoint $x_T(x)$ reached by the algorithm.
Exact integration along any piecewise smooth trajectory recovers this
change, while numerical quadrature introduces a residual in the
completeness relation $\sum_i A_i \approx F^t(x)-F^t(x_T(x))$.
This relation concerns the endpoint actually reached, regardless of
whether its predictive distribution matches $q^\star$. We therefore
evaluate numerical completeness and predictive-target convergence
separately, reporting both the completeness residual and the mismatch
between $p(x_T(x))$ and $q^\star$. Appendix~\ref{app:fringe_algorithm} gives pseudocode for the
full procedure. Figure~\ref{fig:intermediate_steps} illustrates
a representative FRInGe trajectory.

\section{Computational cost and target-versus-rest specialization}
\label{sec:complexity_summary}

FRInGe requires a CG solve at each tracking
step. Each CG iteration evaluates a Fisher matrix-vector product
using one JVP and one VJP.
Appendix~\ref{app:complexity} details the total computational cost.

To reduce this cost, we introduce FRInGe-B, a target-versus-rest
specialization. It applies the FRInGe construction to the aggregated
predictive distribution
$\widetilde p_t(x)=(p_t(x),1-p_t(x))^\top$ and reference
$\widetilde q_t^\star=(1/C,1-1/C)^\top$.
Defining the target-versus-rest log-odds as
\[
\delta_t(x)
=
F^t(x)-\log\!\Big(\sum_{j\ne t}\exp F^j(x)\Big),
\]
the corresponding pullback Fisher tensor is
\begin{equation}
G_{\mathrm B}(x)
=
p_t(x)(1-p_t(x))\,
\nabla_x\delta_t(x)\nabla_x\delta_t(x)^\top.
\label{eq:fringeb_metric_main}
\end{equation}
This tensor has rank at most one, independently of the number
of classes. Substituting $G_{\mathrm B}$ into
Eq.~\eqref{eq:damped_solve} therefore gives a closed-form
unsmoothed direction, eliminating the repeated Fisher JVP/VJP
products and CG solve. The spatially regularized variant uses
solves with the fixed matrix
$\lambda I+\gamma_{\mathrm{step}}L^\top L$,
together with a rank-one correction. Unless stated otherwise, FRInGe-B also keeps both spatial terms active: the suffix B in FRInGe-B identifies target-versus-rest aggregation, not the presence or absence of spatial regularization.

The aggregation of the predictive distribution discards distinctions among non-target
classes, so its trajectories can differ from those of FRInGe.
Appendix~\ref{app:fringeb} provides the detailed derivation, algorithm,
and analysis of attribution quality and computational cost of FRInGe-B.

\section{Experiments}
\label{sec:experiments}
\paragraph{Experimental setup.}
We evaluate FRInGe and its target-versus-rest specialization,
FRInGe-B, on ImageNet validation images. For each image, we explain the
model's top-1 predicted class. We report results on Inception-v3 (Inc-v3),
ResNet-18 (RN18), ResNet-50, ResNet-101, ResNet-152, and VGG-19,
all using standard pretrained weights and their default
preprocessing pipelines. FRInGe and FRInGe-B are tuned
independently. Image cohorts and reference-pool exclusions are
specified in Appendix~\ref{app:benchmark_details}.

We compare against ten attribution baselines:
Integrated Gradients (IG)~\cite{sundararajan2017axiomatic},
IG$^2$~\cite{zhuo2024ig},
SmoothGrad (SG)~\cite{smilkov2017smoothgrad},
BlurIG~\cite{xu2020attribution},
GuidedIG~\cite{kapishnikov2021guided},
Adversarial IG (AGI)~\cite{pan2021explaining},
Expected Gradients (EG)~\cite{erion2021improving},
GGIG~\cite{rahman2022geometrically}, Geodesic Integrated
Gradients (GeoIG)~\cite{salek2025using}, and
Manifold Integrated Gradients (MIG)~\cite{zaher2024manifold}.
Table~\ref{tab:main_quantitative} shows results for FRInGe, FRInGe-B, and a subset of these methods;
Appendix~\ref{app:method_comparison} reports the complete comparison
for all six architectures.

To assess the contributions of pullback geometry and spatial
regularization, we additionally report ablations comparing
Euclidean distribution tracking, Fisher--Rao tracking without spatial
regularization, a Fisher-aware prior-only variant, and FRInGe with
two-term spatial regularization. Implementation details and hyperparameter
settings are provided in the appendix. Code is available at
\url{https://github.com/GabMartino/FRInGe}.

\paragraph{Evaluation protocol and metrics.}
We evaluate each attribution map using four complementary criteria.
First, for \emph{perturbation faithfulness}, we report insertion (Ins-AUC)
and deletion (Del-AUC) area under the curve under blur-based
perturbations~\cite{petsiuk2018rise}, together with
Infidelity~\cite{yeh2019fidelity}, which measures agreement
between attribution-weighted input perturbations and the
corresponding changes in model output.
Second, for \emph{attribution calibration}, we report
Magnitude Aligned Scoring for insertion (MAS-Ins) and
deletion (MAS-Del)~\cite{walker2024MAS}, which assesses whether cumulative
attribution mass aligns with changes in model confidence.
Perturbation AUC primarily evaluates feature ranking, while
MAS additionally accounts for attribution magnitudes.
Third, for \emph{local robustness}, we report Max
Sensitivity~\cite{yeh2019fidelity} under small input perturbations
in the appendix.
Fourth, for \emph{explanation focus}, we report Gini
sparseness~\cite{hurley2009comparing} as a descriptive measure
of attribution concentration.
Full metric definitions and perturbation details are given
in Appendix~\ref{app:metric_definitions}.

\subsection{Comparison against baseline methods}

\begin{table*}[t]
\centering
\scriptsize
\setlength{\tabcolsep}{2.5pt}
\renewcommand{\arraystretch}{0.95}
\medmuskip=1mu 
\caption{Attribution performance on representative architectures (mean $\pm$ half-width of the 95\% bootstrap interval). The best mean score in each row is \textbf{\underline{bolded}}; methods whose displayed mean $\pm$ half-width ranges overlap with that of the best are \textbf{bold}.}
\label{tab:main_quantitative}
\begin{tabular}{@{}llccccccc@{}}
\toprule
\textbf{Metric} & \textbf{Arch.} & \textbf{FRInGe (Ours)} & \textbf{FRInGe-B (Ours)} & \textbf{IG} & \textbf{SG} & \textbf{BlurIG} & \textbf{GGIG} & \textbf{GeoIG} \\
\midrule
\multirow{2}{*}{MAS-Ins $\uparrow$}
& RN18
& $\mathbf{0.597 \pm .015}$
& $\mathbf{\underline{0.601 \pm .014}}$
& $0.443 \pm .015$
& $0.488 \pm .014$
& $\mathbf{0.577 \pm .014}$
& $0.426 \pm .016$
& $0.453 \pm .014$ \\
& Inc-v3
& $\mathbf{\underline{0.761 \pm .009}}$
& $0.743 \pm .008$
& $0.673 \pm .010$
& $0.660 \pm .009$
& $0.735 \pm .009$
& $0.631 \pm .011$
& $0.647 \pm .009$ \\
\midrule
\multirow{2}{*}{MAS-Del $\downarrow$}
& RN18
& $\mathbf{0.264 \pm .010}$
& $\mathbf{\underline{0.259 \pm .008}}$
& $0.359 \pm .011$
& $0.325 \pm .004$
& $0.276 \pm .008$
& $0.373 \pm .008$
& $0.359 \pm .007$ \\
& Inc-v3
& $0.444 \pm .020$
& $0.416 \pm .018$
& $0.485 \pm .019$
& $\mathbf{\underline{0.373 \pm .008}}$
& $\mathbf{0.386 \pm .017}$
& $0.473 \pm .015$
& $0.469 \pm .016$ \\
\midrule
\multirow{2}{*}{Ins-AUC $\uparrow$}
& RN18
& $0.492 \pm .016$
& $\mathbf{0.541 \pm .015}$
& $0.415 \pm .014$
& $0.477 \pm .015$
& $\mathbf{\underline{0.563 \pm .015}}$
& $0.404 \pm .016$
& $0.430 \pm .014$ \\
& Inc-v3
& $0.709 \pm .014$
& $0.723 \pm .013$
& $0.647 \pm .014$
& $0.690 \pm .013$
& $\mathbf{\underline{0.758 \pm .013}}$
& $0.658 \pm .015$
& $0.668 \pm .014$ \\
\midrule
\multirow{2}{*}{Del-AUC $\downarrow$}
& RN18
& $0.130 \pm .008$
& $\mathbf{\underline{0.095 \pm .006}}$
& $0.178 \pm .010$
& $0.122 \pm .007$
& $\mathbf{0.096 \pm .007}$
& $0.164 \pm .009$
& $0.135 \pm .008$ \\
& Inc-v3
& $0.240 \pm .012$
& $0.208 \pm .011$
& $0.271 \pm .013$
& $0.206 \pm .011$
& $\mathbf{\underline{0.174 \pm .011}}$
& $0.215 \pm .011$
& $0.216 \pm .011$ \\
\bottomrule
\end{tabular}
\end{table*}

\begin{figure}[!t]
    \centering
    \includegraphics[width=\linewidth]{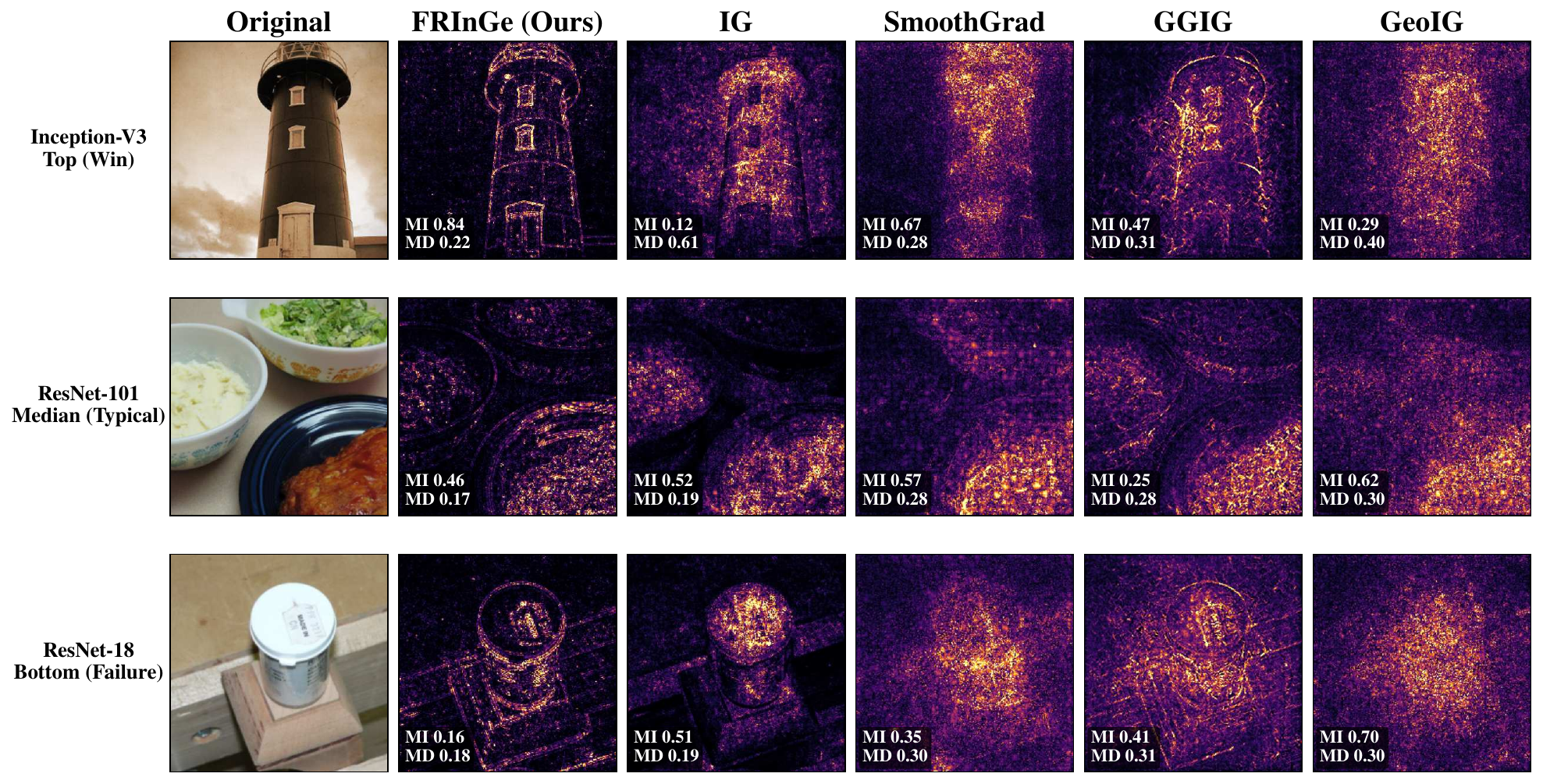}
    \caption{Selected qualitative comparisons on Inception-v3
    (top), ResNet-101 (middle), and ResNet-18 (bottom).
    The ResNet-18 example illustrates reduced boundary detail
    associated with smoothing. Each example reports
    MAS-Insertion (MI) and MAS-Deletion (MD).}
    \label{fig:qualitative_micro_grid}
\end{figure}

Table~\ref{tab:main_quantitative} and the full benchmark in Appendix~\ref{app:method_comparison} (Tables~\ref{tab:full_inception_v3}--\ref{tab:full_vgg19}) highlight FRInGe's strong
performance on attribution magnitude alignment.
Against the ten external baselines, FRInGe achieves the highest
mean MAS-Ins on five of six architectures.
Its gains over geometric competitors are particularly
consistent: FRInGe improves both mean MAS-Ins and MAS-Del
over GGIG and GeoIG on every architecture.
Figure~\ref{fig:qualitative_micro_grid} illustrates selected
qualitative comparisons against these baselines, highlighting differences in attribution concentration and boundary detail.

Among the external baselines, BlurIG leads Ins-AUC and Del-AUC.
A plausible reason is the close alignment between its
Gaussian scale-space path and the blur perturbations used
for evaluation, which may favor its feature ordering under
this protocol.
FRInGe's leading MAS-Ins results under the same perturbation
protocol highlight its strength when evaluation additionally
accounts for attribution magnitudes.

FRInGe-B extends FRInGe with a computationally simpler
target-versus-rest formulation.
When independently tuned, it improves mean
MAS-Del and both Ins-AUC and Del-AUC over FRInGe
on five architectures; FRInGe retains higher mean
MAS-Ins on five architectures
(Table~\ref{tab:fringeb_all_architectures}; see Figure~\ref{fig:fringeb_qualitative} for qualitative comparisons).

\subsection{Ablation study}

We compare Euclidean tracking, Fisher--Rao tracking without spatial
regularization ($\gamma_{\mathrm{step}}=\gamma_{\mathrm{prior}}=0$),
a prior-only variant ($\gamma_{\mathrm{step}}=0$, keeping all
other hyperparameters fixed to FRInGe's tuned values in Table~\ref{tab:fringe_hparams}),
and FRInGe with two-term spatial regularization.
Table~\ref{tab:ablation_study} in the appendix reports quantitative results;
trajectory diagnostics and additional smoothing ablations
appear in Appendix~\ref{app:ablation}.

Among these variants, FRInGe achieves the best mean
MAS-Ins, MAS-Del, and Ins-AUC on all six architectures.
Del-AUC results are mixed between FRInGe and prior-only.
The Euclidean and Fisher variants without spatial regularization produce sparser
maps but weaker MAS and AUC results.
Trajectory diagnostics further show that two-term spatial regularization
suppresses speckled input transformations and yields smoother,
more spatially coherent changes, supporting its role in
the tracking procedure.

\subsection{Hyperparameter sensitivity and local robustness}

We assess FRInGe's hyperparameter sensitivity through one-factor-at-a-time sweeps over
$\delta_{\mathrm{euc}}$, $\tau$, $\eta_{\max}$, $\lambda$, and a tied spatial regularization weight $\gamma_{\mathrm{step}}=\gamma_{\mathrm{prior}}=\gamma$, while keeping the remaining hyperparameters fixed (Appendix~\ref{app:hyper_sens_anal}).
We additionally examine how the Euclidean cap affects
numerical completeness and predictive-space tracking,
and evaluate local robustness against IG on 64 images per
architecture using Max Sensitivity~\cite{yeh2019fidelity}, which measures the maximum relative change in $\ell_2$-normalized attributions under small input perturbations (Appendix~\ref{app:metric_definitions}). The selected architecture-specific hyperparameters are held fixed across images.

Over the tested ranges, the tuning objective (a weighted harmonic mean of Ins-AUC, $1-\text{Del-AUC}$, and Infidelity; Appendix~\ref{app:benchmark_details}) is
comparatively insensitive to $\eta_{\max}$ and stable across
a range of $\tau$ values, with weaker, architecture-dependent
effects from $\lambda$.
The strongest sensitivities arise for $\delta_{\mathrm{euc}}$
and $\gamma$, where moderate nonzero spatial regularization ($\gamma>0$) generally improves
this score.
Activating the Euclidean cap typically reduces the
realized-path completeness residual, but increases endpoint
mismatch and slightly worsens waypoint tracking
(Appendix~\ref{app:euc_tr_quant}).

Local robustness varies between architectures.
FRInGe achieves lower median Max Sensitivity than IG on
ResNet-18 and VGG-19, while IG achieves lower medians on
the other four architectures
(Appendix~\ref{app:max_sens_appendix},
Table~\ref{tab:max_sensitivity_iqr}).

\section{Discussion and Conclusion}

In this paper, we proposed FRInGe which introduces a predictive-space formulation of path-based
attribution, replacing a prescribed baseline input with an
explicit reference distribution.
It constructs a Fisher--Rao geodesic toward a maximum-entropy
reference and computes an input-space trajectory whose
predictions approximately track its waypoints on the path.
The pullback Fisher geometry guides this realization, while
step caps and spatial regularization stabilize and shape
the trajectory.

The clearest empirical advantage of FRInGe appears in MAS-Ins:
FRInGe achieves the highest mean among ten external
baselines on five of six architectures.
It also improves both mean MAS-Ins and MAS-Del over GGIG
and GeoIG across all six architectures we benchmarked it on.
These results support stronger alignment between attribution
magnitudes and model confidence under the tested protocols.
BlurIG leads Ins-AUC and Del-AUC, an advantage that
may partly reflect the alignment between its Gaussian
scale-space path and the evaluation perturbations.

FRInGe-B provides a computationally simpler specialization
through target-versus-rest aggregation.
Its rank-one pullback removes repeated model-dependent
Fisher matrix-vector products from the inner solve.
The two formulations offer distinct performance profiles:
FRInGe retains information about the complete class
distribution and achieves higher MAS-Ins and sparseness, while FRInGe-B reduces the computational burden
of constructing a target-specific trajectory and improves MAS-Del, Ins-AUC, and Del-AUC on most architectures.

Several limitations remain.
Finite-step tracking can leave a predictive endpoint mismatch,
and completeness applies to the realized input endpoint,
up to numerical integration error.
Damping, spatial regularization, and Euclidean displacement
caps also make the realized trajectory dependent on the
input representation.
Applications beyond the image models evaluated here require
appropriate regularizers and separate validation.

More broadly, FRInGe and FRInGe-B establish two instances of a general distribution-guided attribution framework.
In this framework, the target predictive reference and class aggregation become explicit design choices, while the corresponding input endpoint emerges from the trajectory tracking procedure.
Alternative predictive references, class groupings, and modality-specific regularizers provide concrete directions for extending our work.

\begin{ack}
This work has been funded in parts by the Vienna Science and Technology Fund (WWTF) [10.47379/LS23053].
\end{ack}

\bibliographystyle{plain}
\bibliography{bibliography}

@book{cencov1982statistical,
  author    = {{\v{C}}encov, N. N.},
  title     = {Statistical Decision Rules and Optimal Inference},
  series    = {Translations of Mathematical Monographs},
  volume    = {53},
  publisher = {American Mathematical Society},
  address   = {Providence, RI},
  year      = {1982},
  isbn      = {978-0-8218-4502-8}
}

@book{amari2016information,
  author    = {Amari, Shun-ichi},
  title     = {Information Geometry and Its Applications},
  series    = {Applied Mathematical Sciences},
  volume    = {194},
  publisher = {Springer Japan},
  year      = {2016},
  doi       = {10.1007/978-4-431-55978-8}
}

@article{sturmfels2020visualizing,
  author = {Sturmfels, Pascal and Lundberg, Scott and Lee, Su-In},
  title = {Visualizing the Impact of Feature Attribution Baselines},
  journal = {Distill},
  year = {2020},
  note = {https://distill.pub/2020/attribution-baselines},
  doi = {10.23915/distill.00022}
}

@article{erion2021improving,
  title={Improving performance of deep learning models with axiomatic attribution priors and expected gradients},
  author={Erion, Gabriel and Janizek, Joseph D and Sturmfels, Pascal and Lundberg, Scott M and Lee, Su-In},
  journal={Nature Machine Intelligence},
  volume={3},
  number={7},
  pages={620--631},
  year={2021},
  publisher={Nature Publishing Group UK London}
}

@InProceedings{pmlr-v119-sundararajan20b,
  title = 	 {The Many Shapley Values for Model Explanation},
  author =       {Sundararajan, Mukund and Najmi, Amir},
  booktitle = 	 {International Conference on Machine Learning (ICML)},
  pages = 	 {9269--9278},
  year = 	 {2020}
}

@article{liu2023new,
  title={A new baseline assumption of integated gradients based on shaply value},
  author={Liu, Shuyang and Chen, Zixuan and Shi, Ge and Wang, Ji and Fan, Changjie and Xiong, Yu and Hu, Runze Wu Yujing and Ji, Ze and Gao, Yang},
  journal={arXiv preprint arXiv:2310.04821},
  year={2023}
}

@article{bardhan2024constructing,
  title={Constructing sensible baselines for Integrated Gradients},
  author={Bardhan, Jai and Neeraj, Cyrin and Rawat, Mihir and Mitra, Subhadip},
  journal={arXiv preprint arXiv:2412.13864},
  year={2024}
}

@article{tuan2025weighted,
  title={Enhancing Visual Feature Attribution via Weighted Integrated Gradients},
  author={Tuan, Kien Tran Duc and Trong, Tam Nguyen and Hoang, Son Nguyen and Than, Khoat and Duc, Anh Nguyen},
  journal={arXiv preprint arXiv:2505.03201},
  year={2025}
}

@inproceedings{sundararajan2017axiomatic,
  title={Axiomatic attribution for deep networks},
  author={Sundararajan, Mukund and Taly, Ankur and Yan, Qiqi},
  booktitle={International Conference on Machine Learning (ICML)},
  pages={3319--3328},
  year={2017},
}

@article{zhuo2024ig,
  title={{IG}$^2$: Integrated Gradient on Iterative Gradient Path for Feature Attribution},
  author={Zhuo, Yue and Ge, Zhiqiang},
  journal={Transactions on Pattern Analysis and Machine Intelligence (TPAMI)},
  volume={46},
  number={11},
  pages={7173--7190},
  year={2024},
  publisher={IEEE}
}

@article{smilkov2017smoothgrad,
  title={{SmoothGrad}: Removing Noise by Adding Noise},
  author={Smilkov, Daniel and Thorat, Nikhil and Kim, Been and Vi{\'e}gas, Fernanda and Wattenberg, Martin},
  journal={arXiv preprint arXiv:1706.03825},
  year={2017}
}

@inproceedings{kapishnikov2021guided,
  title={Guided integrated gradients: An adaptive path method for removing noise},
  author={Kapishnikov, Andrei and Venugopalan, Subhashini and Avci, Besim and Wedin, Ben and Terry, Michael and Bolukbasi, Tolga},
  booktitle={Conference on Computer Vision and Pattern Recognition (CVPR)},
  pages={5050--5058},
  publisher={IEEE},
  year={2021}
}

@inproceedings{xu2020attribution,
  title={Attribution in scale and space},
  author={Xu, Shawn and Venugopalan, Subhashini and Sundararajan, Mukund},
  booktitle={Conference on Computer Vision and Pattern Recognition (CVPR)},
  pages={9680--9689},
  publisher={IEEE},
  year={2020}
}

@inproceedings{pan2021explaining,
  title={Explaining deep neural network models with adversarial gradient integration},
  author={Pan, Deng and Li, Xin and Zhu, Dongxiao},
  booktitle={International Joint Conference on Artificial Intelligence (IJCAI)},
  year={2021}
}

@inproceedings{petsiuk2018rise,
  title={{RISE}: Randomized Input Sampling for Explanation of Black-Box Models},
  author={Petsiuk, Vitali and Das, Abir and Saenko, Kate},
  booktitle={British Machine Vision Conference (BMVC)},
  pages={151},
  publisher={BMVA Press},
  year={2018},
  url={http://bmvc2018.org/contents/papers/1064.pdf}
}

@inproceedings{walker2024MAS,
  title={Attribution Quality Metrics with Magnitude Alignment},
  author={Walker, Chase and Simon, Dominic and Chen, Kenny and Ewetz, Rickard},
  booktitle={Proceedings of the Thirty-Third International Joint Conference on Artificial Intelligence (IJCAI)},
  pages={530--538},
  year={2024},
  doi={10.24963/ijcai.2024/59}
}

@article{yeh2019fidelity,
  title={On the (In)fidelity and Sensitivity of Explanations},
  author={Yeh, Chih-Kuan and Hsieh, Cheng-Yu and Suggala, Arun and Inouye, David I and Ravikumar, Pradeep K},
  journal={Advances in Neural Information Processing Systems (NeurIPS)},
  volume={32},
  year={2019}
}

@inproceedings{tan2023maximum,
  title={Maximum entropy baseline for integrated gradients},
  author={Tan, Hanxiao},
  booktitle={2023 International Joint Conference on Neural Networks (IJCNN)},
  pages={1--8},
  year={2023},
  organization={IEEE}
}

@inproceedings{ren2025improving,
  title={Improving integrated gradient-based transferable adversarial examples by refining the integration path},
  author={Ren, Yuchen and Zhao, Zhengyu and Lin, Chenhao and Yang, Bo and Zhou, Lu and Liu, Zhe and Shen, Chao},
  booktitle={Conference on Artificial Intelligence (AAAI)},
  volume={39},
  pages={6731--6739},
  year={2025}
}

@inproceedings{rahman2022geometrically,
  title={Geometrically guided saliency maps},
  author={Rahman, Md Mahfuzur and Lewis, Noah and Plis, Sergey},
  booktitle={ICLR 2022 Workshop on PAIR\textsuperscript{2}Struct: Privacy, Accountability, Interpretability, Robustness, Reasoning on Structured Data},
  year={2022}
}

@InProceedings{zaher2024manifold,
  title = 	 {Manifold Integrated Gradients: {R}iemannian Geometry for Feature Attribution},
  author =       {Zaher, Eslam and Trzaskowski, Maciej and Nguyen, Quan and Roosta, Fred},
  booktitle = 	 {International Conference on Machine Learning (ICML)},
  pages = 	 {58090--58104},
  year = 	 {2024},
}

@inproceedings{walker2024integrated,
  title={Integrated decision gradients: Compute your attributions where the model makes its decision},
  author={Walker, Chase and Jha, Sumit and Chen, Kenny and Ewetz, Rickard},
  booktitle={Conference on Artificial Intelligence (AAAI)},
  volume={38},
  pages={5289--5297},
  year={2024}
}

@article{salek2025using,
  title={Using the Path of Least Resistance to Explain Deep Networks},
  author={Salek, Sina and Enguehard, Joseph},
  journal={arXiv preprint arXiv:2502.12108},
  year={2025}
}

@article{hurley2009comparing,
  title={Comparing Measures of Sparsity},
  author={Hurley, Niall and Rickard, Scott},
  journal={IEEE Transactions on Information Theory},
  volume={55},
  number={10},
  pages={4723--4741},
  year={2009},
  doi={10.1109/TIT.2009.2027527}
}

@inproceedings{akiba2019optuna,
  title={Optuna: A Next-Generation Hyperparameter Optimization Framework},
  author={Akiba, Takuya and Sano, Shotaro and Yanase, Toshihiko and Ohta, Takeru and Koyama, Masanori},
  booktitle={Proceedings of the 25th ACM SIGKDD International Conference on Knowledge Discovery and Data Mining},
  pages={2623--2631},
  publisher={ACM},
  year={2019},
  doi={10.1145/3292500.3330701}
}

@article{SAYRES2019552,
title = {Using a Deep Learning Algorithm and Integrated Gradients Explanation to Assist Grading for Diabetic Retinopathy},
journal = {Ophthalmology},
volume = {126},
number = {4},
pages = {552-564},
year = {2019},
issn = {0161-6420},
doi = {https://doi.org/10.1016/j.ophtha.2018.11.016},
url = {https://www.sciencedirect.com/science/article/pii/S0161642018315756},
author = {Rory Sayres and Ankur Taly and Ehsan Rahimy and Katy Blumer and David Coz and Naama Hammel and Jonathan Krause and Arunachalam Narayanaswamy and Zahra Rastegar and Derek Wu and Shawn Xu and Scott Barb and Anthony Joseph and Michael Shumski and Jesse Smith and Arjun B. Sood and Greg S. Corrado and Lily Peng and Dale R. Webster}
}

@article{saporta2022benchmarking,
  title={Benchmarking saliency methods for chest X-ray interpretation},
  author={Saporta, Adriel and Gui, Xiaotong and Agrawal, Ashwin and Pareek, Anuj and Truong, Steven QH and Nguyen, Chanh DT and Ngo, Van-Doan and Seekins, Jayne and Blankenberg, Francis G and Ng, Andrew Y and others},
  journal={Nature Machine Intelligence},
  volume={4},
  number={10},
  pages={867--878},
  year={2022},
  publisher={Nature Publishing Group UK, London}
}

@article{yu2023identifying,
  title={Identifying important sensory feedback for learning locomotion skills},
  author={Yu, Wanming and Yang, Chuanyu and McGreavy, Christopher and Triantafyllidis, Eleftherios and Bellegarda, Guillaume and Shafiee, Milad and Ijspeert, Auke Jan and Li, Zhibin},
  journal={Nature Machine Intelligence},
  volume={5},
  number={8},
  pages={919--932},
  year={2023},
  publisher={Nature Publishing Group UK, London}
}

@article{nauta2023anecdotal,
  title={From anecdotal evidence to quantitative evaluation methods: A systematic review on evaluating explainable ai},
  author={Nauta, Meike and Trienes, Jan and Pathak, Shreyasi and Nguyen, Elisa and Peters, Michelle and Schmitt, Yasmin and Schl{\"o}tterer, J{\"o}rg and Van Keulen, Maurice and Seifert, Christin},
  journal={ACM Computing Surveys},
  volume={55},
  number={13s},
  pages={1--42},
  year={2023},
  publisher={ACM, New York, NY}
}

@article{montavon2018methods,
  title={Methods for interpreting and understanding deep neural networks},
  author={Montavon, Gr{\'e}goire and Samek, Wojciech and M{\"u}ller, Klaus-Robert},
  journal={Digital signal processing},
  volume={73},
  pages={1--15},
  year={2018},
  publisher={Elsevier}
}

@inproceedings{shoemake1985animating,
  title={Animating rotation with quaternion curves},
  author={Shoemake, Ken},
  booktitle={Proceedings of the 12th annual conference on Computer graphics and interactive techniques},
  pages={245--254},
  year={1985}
}

@article{arvanitidis2021pulling,
  title={Pulling back information geometry},
  author={Arvanitidis, Georgios and Gonz{\'a}lez-Duque, Miguel and Pouplin, Alison and Kalatzis, Dimitris and Hauberg, S{\o}ren},
  journal={arXiv preprint arXiv:2106.05367},
  year={2021}
}

@article{martens2020new,
  title={New insights and perspectives on the natural gradient method},
  author={Martens, James},
  journal={Journal of Machine Learning Research},
  volume={21},
  number={146},
  pages={1--76},
  year={2020}
}

@article{amari1998natural,
  title={Natural gradient works efficiently in learning},
  author={Amari, Shun-Ichi},
  journal={Neural computation},
  volume={10},
  number={2},
  pages={251--276},
  year={1998},
  publisher={MIT Press}
}

@inproceedings{schulman2015trust,
  title={Trust region policy optimization},
  author={Schulman, John and Levine, Sergey and Abbeel, Pieter and Jordan, Michael and Moritz, Philipp},
  booktitle={International conference on machine learning},
  pages={1889--1897},
  year={2015},
  organization={Pmlr}
}

@article{szegedy2013intriguing,
  title={Intriguing properties of neural networks},
  author={Szegedy, Christian and Zaremba, Wojciech and Sutskever, Ilya and Bruna, Joan and Erhan, Dumitru and Goodfellow, Ian and Fergus, Rob},
  journal={arXiv preprint arXiv:1312.6199},
  year={2013}
}

@inproceedings{wang2020high,
  title={High-frequency component helps explain the generalization of convolutional neural networks},
  author={Wang, Haohan and Wu, Xindi and Huang, Zeyi and Xing, Eric P},
  booktitle={2020 IEEE/CVF Conference on Computer Vision and Pattern Recognition (CVPR)},
  pages={8681--8691},
  year={2020},
  organization={IEEE}
}

@article{jacobson2011bounded,
  title={Bounded biharmonic weights for real-time deformation.},
  author={Jacobson, Alec and Baran, Ilya and Popovic, Jovan and Sorkine, Olga},
  journal={ACM Trans. Graph.},
  volume={30},
  number={4},
  pages={78},
  year={2011}
}

\clearpage
\appendix
\etocdepthtag.toc{appendixmatter}

\begingroup
\small
\etocsettagdepth{mainmatter}{none}
\etocsettagdepth{appendixmatter}{subsection}
\etocsetnexttocdepth{subsection}
\etocsettocstyle{\section*{Appendix Contents}}{}
\tableofcontents
\endgroup

\clearpage

\section{FRInGe Methodological Details}
\label{app:fringe_impl}

This appendix explains how the predictive-space construction in
Section~\ref{sec:method} is realized numerically and how to interpret the
resulting trajectory. We begin with the complete algorithm, then explain
matrix-free Fisher products, the distinct roles of the spatial penalties,
and the approximate solution of the regularized system. We next examine
how the predictive and Euclidean step-size constraints interact, using
trajectory profiles and quantitative diagnostics to distinguish accuracy
of the path integral from agreement with the prescribed predictive
waypoints. Finally, we give the leading computational cost and measured
latency of categorical FRInGe. Together, these details explain which
approximations affect the update direction, the predictive endpoint,
and the accumulated attributions.

\subsection{FRInGe Pseudocode}
\label{app:fringe_algorithm}

Algorithm~\ref{alg:fringe_small} combines waypoint construction, input-space
tracking, and attribution accumulation. The target class $t$ remains fixed
throughout. At each input state, the direction is computed toward the next
predictive waypoint, and the step size is then restricted before the
target-logit gradients are integrated along the resulting input segment.
The trajectory runs from the original input toward the realized endpoint;
the final sign reversal gives attributions oriented from that endpoint
back to the original input, consistently with Eq.~\eqref{eq:completeness}.

\renewcommand{\algorithmicensure}{\textbf{Returns:}}
\begin{algorithm}[H]
\caption{FRInGe}
\label{alg:fringe_small}
\begin{algorithmic}[1]
\Require Input $x$, target class $t$, logits $F$; damping $\lambda$, KL budget $\tau$,
step cap $\eta_{\max}$, Euclidean cap $\delta_{\mathrm{euc}}$; spatial operator
$L$ and coefficients $\gamma_{\mathrm{step}},\gamma_{\mathrm{prior}}$
\Ensure Attributions $A$
\State Compute $T$ using Eq.~\eqref{eq:step_sync}
\State $p_0\gets\softmax(F(x))$, $q^\star\gets\tfrac{1}{C}\mathbf{1}$
\State Build Fisher--Rao waypoints $\{s_k\}_{k=0}^T$ between $\sqrt{p_0}$ and $\sqrt{q^\star}$
\State $x_0\gets x$, $A_{\mathrm{path}}\gets 0$
\For{$k=0,\dots,T-1$}
    \State $p_k\gets\softmax(F(x_k))$
    \State $\mathcal{L}_{k+1}\gets 1-\langle \sqrt{p_k},s_{k+1}\rangle$
    \State $g_k\gets \nabla_x \mathcal{L}_{k+1}(x)\rvert_{x_k}$
    \State Solve $(G_k+\lambda I+\gamma_{\mathrm{step}}L^\top L)v_k
    =g_k+\gamma_{\mathrm{prior}}L^\top Lx_k$ for $v_k$ by CG
    \State $n_k^2\gets v_k^\top G_kv_k$
    \State $\eta_k\gets\min\!\left(\eta_{\max},
    \sqrt{\frac{2\tau}{n_k^2+\varepsilon}},
    \frac{\delta_{\mathrm{euc}}}{\|v_k\|_2+\varepsilon}\right)$
    \State $x_{k+1}\gets x_k-\eta_kv_k$
    \State $A_{\mathrm{path}}\gets A_{\mathrm{path}}+
    \frac12(\nabla_xF^t(x_k)+\nabla_xF^t(x_{k+1}))
    \odot(x_{k+1}-x_k)$
\EndFor
\State \Return $A=-A_{\mathrm{path}}$
\end{algorithmic}
\end{algorithm}

\subsection{Matrix-Free Pullback Fisher Products}

The pullback tensor in Eq.~\eqref{eq:pullback_metric} measures local
predictive sensitivity, not Euclidean input distance. Its quadratic form
can be small even for a large input displacement if the classifier is
locally insensitive in that direction. This distinction motivates both
damping in the linear system and the separate input-space cap below.

The numerical solver needs only the action of this tensor on a vector,
not the dense $n\times n$ matrix. Using the logit Jacobian $J_F(x)$ and
the softmax Fisher matrix $S(x)$ defined in the background, we evaluate
\[
G(x)v = J_F(x)^\top \bigl(S(x)(J_F(x)v)\bigr).
\]
A Jacobian-vector product maps an input direction into logit space, the
softmax Fisher matrix weights its predictive effect, and a vector-Jacobian
product maps the result back to input space. This avoids storing the dense
input-space tensor, although every such product still requires
model-dependent differentiation. That repeated derivative work, rather
than matrix storage alone, determines much of the computational cost.

\subsection{Spatial Regularization and Numerical Solution}
\label{app:damping}

At state $x_k$, the regularized direction satisfies
\[
\bigl(G(x_k)+\lambda I+\gamma_{\mathrm{step}}L^\top L\bigr)v_k
=
g_k+\gamma_{\mathrm{prior}}L^\top Lx_k,
\]
where
\[
g_k=\nabla_x\mathcal{L}_{k+1}(x)\rvert_{x=x_k},
\]
\(\lambda>0\) supplies Euclidean damping, and $L$ measures spatial
roughness. For the image models, $L$ is a discrete Laplacian, so the
quadratic penalty acts through $L^\top L$. These spatial terms express
an image-specific prior; they are not required by the predictive
Fisher--Rao construction.

The two nonnegative coefficients serve different purposes.
$\gamma_{\mathrm{step}}$ penalizes rough update directions through the
left-hand side, whereas $\gamma_{\mathrm{prior}}$ introduces the current
input's roughness gradient on the right-hand side. With the update
$x_{k+1}=x_k-\eta_k v_k$, the latter term encourages removal of spatial
roughness from the current iterate. Consequently, smoothing the direction
and encouraging a smoother input trajectory are distinct choices.
Setting $\gamma_{\mathrm{step}}=\gamma_{\mathrm{prior}}=\gamma$ is a
tied-coefficient special case, not a requirement of the method.

We approximate the direction using preconditioned conjugate gradients
(PCG). In the spatially regularized setting, the empirical preconditioner
attenuates high-frequency components by Gaussian blurring and scaling:
\[
M^{-1}(v)\approx \frac{\mathrm{Blur}(v)}{\lambda}.
\]
This is a numerical choice for the iterative solve, not an exact inverse
of $L^\top L$ or a replacement for the spatial penalties. When smoothing
is disabled, it reduces to diagonal scaling,
\[
M^{-1}(v)=\frac{v}{\lambda}.
\]

The preceding waypoint's direction initializes the next solve. A finite
iteration ceiling and residual tolerances limit the work per waypoint,
so the computed direction need not solve the displayed system exactly.
Linear-solve accuracy, predictive-waypoint tracking, and path-integral
accuracy are therefore separate numerical considerations.

\subsection{Trust-Region Step Selection}

Once $v_k$ is computed, Eq.~\eqref{eq:combined_step} selects the smallest
of the global step-size ceiling, the local KL-based bound, and the
Euclidean displacement bound. These constraints act on different
quantities. The KL-based bound limits predictive change according to the
local Fisher quadratic model; it is not an exact constraint on the
nonlinear model's KL divergence. The Euclidean bound instead limits the
length of the input segment used for numerical integration. It can be
restrictive even when the predicted change in the output distribution
is small.

The active constraint can change along a trajectory. In regions of low
predictive sensitivity, the Euclidean cap may limit the input displacement;
elsewhere, the Fisher-based bound or the global ceiling may determine the
step. Shorter input segments can improve the resolution of the path
integral, but they need not improve tracking of the prescribed waypoints
within a fixed number of updates. In particular, $T$ is chosen from the
nominal predictive arc length before tracking begins. Neither that choice
nor the step-size rule guarantees that the realized endpoint predicts
the reference distribution.

\subsection{Geometric Effect of the Euclidean Trust-Region Cap}
\label{app:euc_tr_geometry}

\begin{figure}[H]
    \centering
    \includegraphics[width=0.9\textwidth]{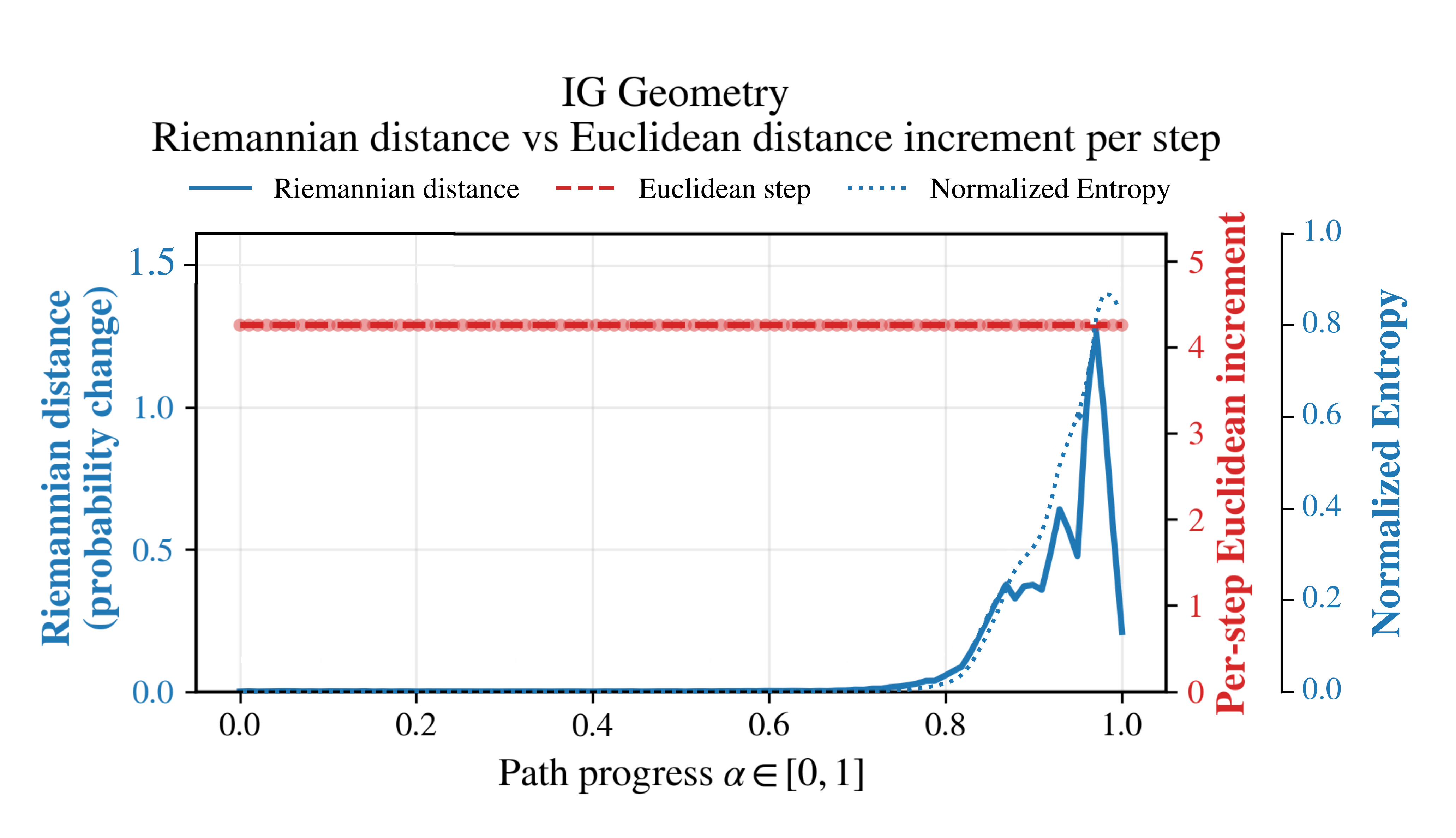}
    \caption{Geometric profile of standard IG. Constant Euclidean steps induce highly non-uniform predictive-space progress, with entropy increasing mainly near the end of the displayed path.}
    \label{fig:geom_ig}
\end{figure}

\begin{figure}[H]
    \centering
    \includegraphics[width=0.9\textwidth]{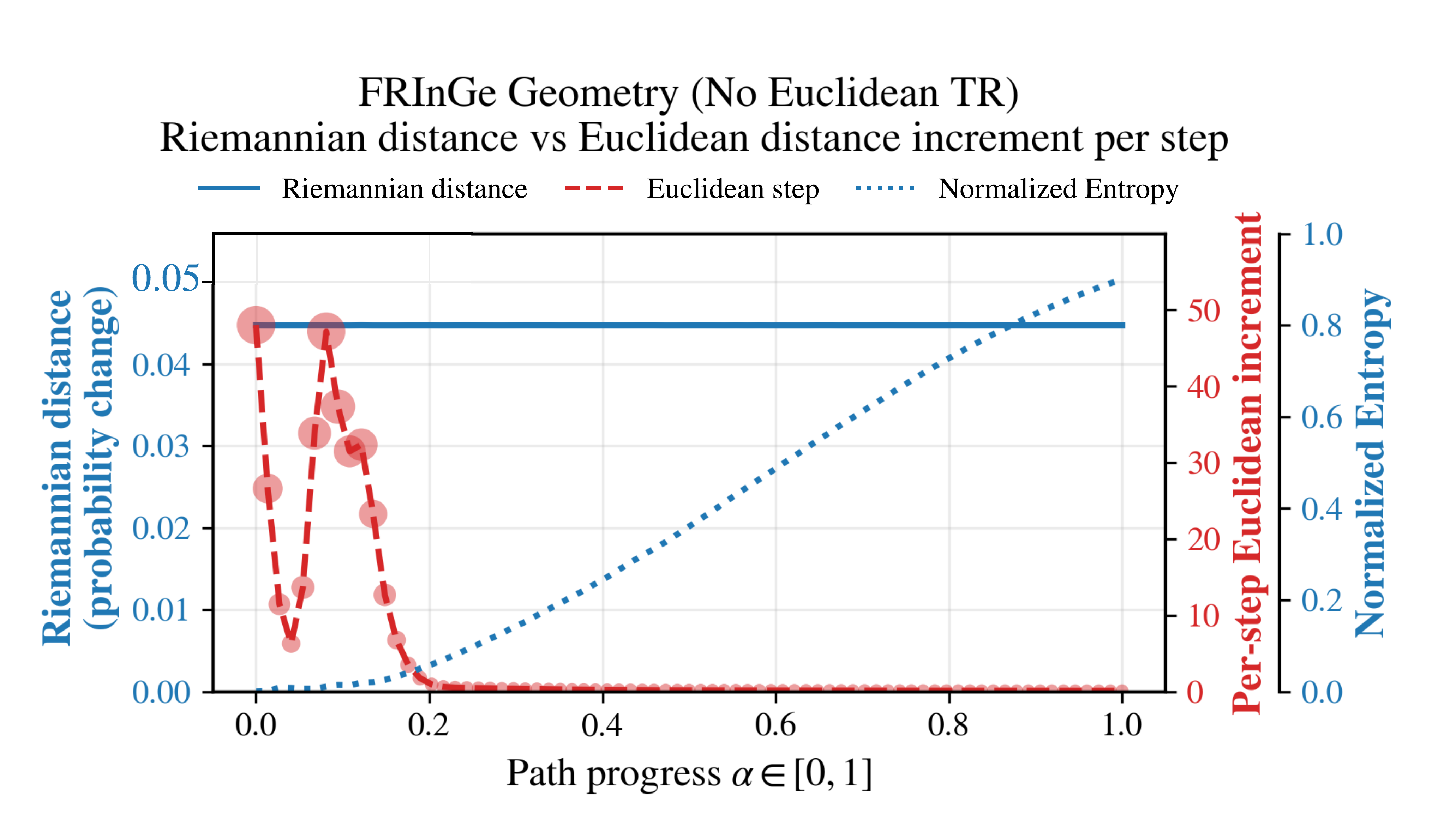}
    \caption{Geometric profile of FRInGe without the Euclidean cap. Approximately constant Fisher--Rao steps coexist with large early displacements in input space; normalized predictive entropy rises more smoothly along the trajectory. Larger Euclidean steps are required in flat geometry regions.}
    \label{fig:geom_fringe_noeuc}
\end{figure}

\begin{figure}[H]
    \centering
    \includegraphics[width=0.9\textwidth]{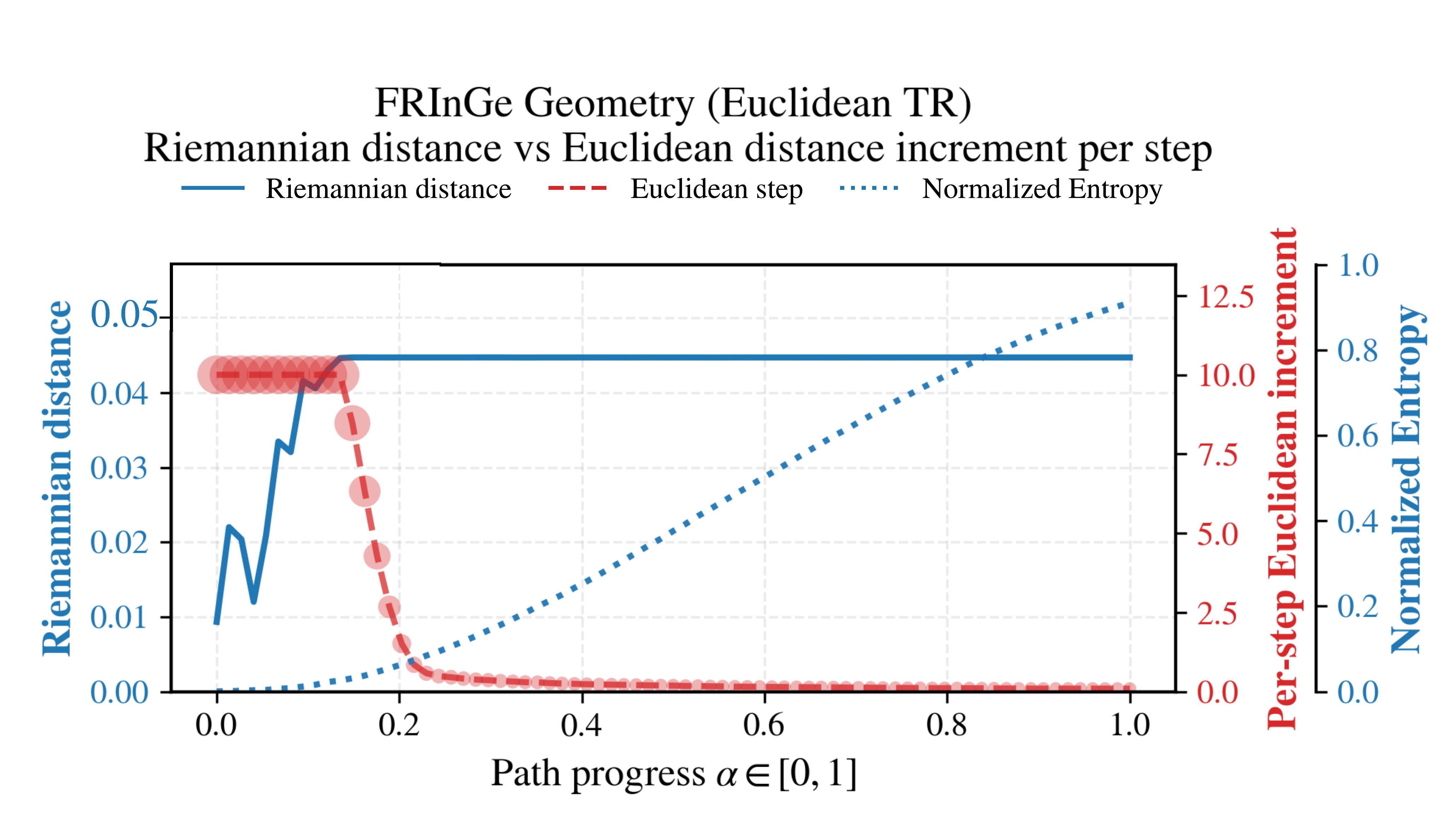}
    \caption{Geometric profile of FRInGe with the Euclidean cap. The cap limits early input-space displacements to approximately $10$; later, the KL constraint limits the step and the Fisher--Rao increment becomes approximately constant.}
    \label{fig:geom_fringe_euc}
\end{figure}

Figures~\ref{fig:geom_ig}, \ref{fig:geom_fringe_noeuc},
and~\ref{fig:geom_fringe_euc} illustrate the interaction between the two
step constraints through three example trajectories: standard IG,
FRInGe without the Euclidean cap, and FRInGe with both bounds. Each
profile shows the local Fisher step, the Euclidean input displacement,
and normalized predictive entropy. These examples illustrate possible
trajectory regimes rather than establishing their prevalence across
images.

For the straight-line IG path in Fig.~\ref{fig:geom_ig}, equal Euclidean
steps produce highly uneven predictive changes. The Fisher step stays
close to zero over much of the path and rises sharply near its endpoint,
while entropy increases mainly late in the trajectory. The displayed
orientation runs from the original input to the baseline, opposite to
the usual attribution-integration direction. This example shows why
uniform sampling in input space need not resolve predictive changes
uniformly.

Without the Euclidean cap, Fig.~\ref{fig:geom_fringe_noeuc} shows a more
uniform local Fisher step and a smoother entropy profile. However,
maintaining the predictive-space step in regions of low sensitivity can
require large input displacements. The Euclidean step norm reaches
approximately $50$ in this example. Such long segments can leave substantial
variation in target-logit gradients unresolved by the quadrature, even
when the local predictive step appears controlled.

With the Euclidean cap, Fig.~\ref{fig:geom_fringe_euc} initially limits
input displacements to approximately $10$. Later, the KL-based bound
becomes more restrictive, and the local Fisher increment is approximately
constant instead. The profile thus illustrates a transition between
input-space and predictive-space step control. The cap's purpose is to
stabilize the numerical integration path, not to guarantee closer waypoint
tracking. The quantitative comparison in Appendix~\ref{app:euc_tr_quant}
examines this distinction across models.

\subsection{Quantitative Effect of the Euclidean Trust-Region Cap}
\label{app:euc_tr_quant}

The Euclidean trust-region radius $\delta_{\mathrm{euc}}$ introduces a
trade-off between numerical integration accuracy and predictive-space
agreement. We assess these separately through realized-path completeness,
endpoint mismatch, and waypoint tracking.

First, we evaluate the \emph{realized-path completeness residual}, measured with respect to the endpoint actually reached by the discrete trajectory:
\[
\varepsilon_{\mathrm{comp}}
=
\frac{\left|\sum_i A_i-\bigl(F^t(x_0)-F^t(x_T)\bigr)\right|}
{\left|F^t(x_0)-F^t(x_T)\right|+\varepsilon}.
\]
Here $\varepsilon>0$ prevents division by zero. This residual measures
numerical integration error relative to the endpoint actually reached,
not agreement with the desired predictive reference. We also report
the endpoint $\mathrm{KL}\!\bigl(p(x_T)\,\|\,u\bigr)$, with $u$
denoting the uniform reference. A small completeness residual can
therefore coexist with a large endpoint mismatch.

Second, we quantify predictive-space waypoint tracking by measuring the Fisher--Rao distance between the realized predictive distribution and the intended geodesic waypoint:
\[
\varepsilon_{\mathrm{track},k}=d_{\mathrm{FR}}(p(x_k),q_k).
\]
We summarize this tracking error using both its mean and its maximum over the trajectory.

Tables~\ref{tab:euc_cap_completeness} and~\ref{tab:euc_cap_tracking} show a consistent pattern across the five models with a restrictive Euclidean cap (Inception-v3 is omitted as its tuned cap $\delta_{\mathrm{euc}}=24.18$ is rarely active). Activating $\delta_{\mathrm{euc}}$ often reduces the completeness residual of the realized input-space integral, but it also tends to increase the final endpoint KL and to worsen predictive-space waypoint tracking. The Euclidean cap should therefore be interpreted as a stabilizer of the realized integration path rather than as a mechanism for exact nonlinear shooting toward the reference distribution.

\begin{table*}[h!]
\centering
\small
\setlength{\tabcolsep}{5pt}
\renewcommand{\arraystretch}{1.1}
\caption{Effect of the Euclidean trust-region cap on realized-path completeness and endpoint mismatch. Lower completeness residual is better. Lower endpoint KL indicates closer agreement with the ideal maximum-entropy target.}
\label{tab:euc_cap_completeness}
\begin{tabular}{lcccc}
\toprule
Model &
\makecell{Completeness error\\$\delta_{\mathrm{euc}}$ on} &
\makecell{Completeness error\\$\delta_{\mathrm{euc}}$ off} &
\makecell{Endpoint KL\\$\delta_{\mathrm{euc}}$ on} &
\makecell{Endpoint KL\\$\delta_{\mathrm{euc}}$ off} \\
\midrule
ResNet-18  & \textbf{0.011 $\pm$ 0.003} & 0.078 $\pm$ 0.034 & 3.697 $\pm$ 0.605 & \textbf{0.148 $\pm$ 0.030} \\
ResNet-50  & \textbf{0.417 $\pm$ 0.174} & 0.535 $\pm$ 0.168 & 2.749 $\pm$ 0.473 & \textbf{2.435 $\pm$ 0.474} \\
ResNet-101 & \textbf{0.184 $\pm$ 0.061} & 0.313 $\pm$ 0.108 & 2.096 $\pm$ 0.530 & \textbf{1.439 $\pm$ 0.383} \\
ResNet-152 & \textbf{0.208 $\pm$ 0.070} & 0.239 $\pm$ 0.071 & 1.235 $\pm$ 0.328 & \textbf{1.161 $\pm$ 0.288} \\
VGG-19     & \textbf{0.014 $\pm$ 0.005} & 0.258 $\pm$ 0.360 & 4.707 $\pm$ 0.603 & \textbf{0.188 $\pm$ 0.088} \\
\bottomrule
\end{tabular}
\end{table*}

\begin{table*}[h!]
\centering
\small
\setlength{\tabcolsep}{5pt}
\renewcommand{\arraystretch}{1.1}
\caption{Effect of the Euclidean trust-region cap on predictive-space waypoint tracking. Lower values indicate closer agreement between the realized trajectory and the intended Fisher--Rao waypoint sequence.}
\label{tab:euc_cap_tracking}
\begin{tabular}{lcccc}
\toprule
Model &
\makecell{Mean tracking error\\$\delta_{\mathrm{euc}}$ on} &
\makecell{Mean tracking error\\$\delta_{\mathrm{euc}}$ off} &
\makecell{Max tracking error\\$\delta_{\mathrm{euc}}$ on} &
\makecell{Max tracking error\\$\delta_{\mathrm{euc}}$ off} \\
\midrule
ResNet-18  & 0.197 $\pm$ 0.022 & \textbf{0.046 $\pm$ 0.008} & 0.977 $\pm$ 0.015 & \textbf{0.850 $\pm$ 0.088} \\
ResNet-50  & 0.154 $\pm$ 0.017 & \textbf{0.141 $\pm$ 0.017} & 0.563 $\pm$ 0.083 & \textbf{0.540 $\pm$ 0.087} \\
ResNet-101 & 0.145 $\pm$ 0.035 & \textbf{0.124 $\pm$ 0.035} & 0.859 $\pm$ 0.073 & \textbf{0.817 $\pm$ 0.086} \\
ResNet-152 & 0.093 $\pm$ 0.014 & \textbf{0.090 $\pm$ 0.013} & 0.711 $\pm$ 0.099 & \textbf{0.700 $\pm$ 0.101} \\
VGG-19     & 0.254 $\pm$ 0.024 & \textbf{0.054 $\pm$ 0.014} & 0.917 $\pm$ 0.035 & \textbf{0.556 $\pm$ 0.119} \\
\bottomrule
\end{tabular}
\end{table*}

\FloatBarrier
\subsection{Computational Cost and Measured Latency}
\label{app:complexity}

\paragraph{Leading-order model evaluations.}
Let $C_{\mathrm{fwd}}$ and $C_{\mathrm{bwd}}$ denote the costs of a model
forward and reverse-mode pass. A pullback-Fisher product requires one JVP and
one VJP, with costs comparable to $C_{\mathrm{fwd}}$ and
$C_{\mathrm{bwd}}$, respectively. Let $N$ be the number of IG integration
steps and $T$ the number of FRInGe tracking steps. Let $M$ denote the average
number of pullback-Fisher products evaluated per categorical waypoint,
including the products required by CG and the trust-region norm. Thus
$M=O(K)$ when $K$ is
the CG iteration ceiling, but $M$ need not equal $K$ exactly.

Table~\ref{tab:operator_complexity} separates classifier-dependent derivative
work from structured spatial-regularizer operations. The accounting treats a
gradient evaluation as its forward--reverse pair and omits lower-order setup,
solver bookkeeping, and opportunities for graph reuse or vectorization.

\begin{table}[H]
\centering
\small
\setlength{\tabcolsep}{4pt}
\caption{Leading classifier-dependent work for one attribution. Spatial
regularizer operations are accounted for separately in the text.}
\label{tab:operator_complexity}
\begin{tabularx}{\linewidth}{@{}l c X >{\raggedleft\arraybackslash}p{0.27\linewidth}@{}}
\toprule
Method & Steps & Work per step & Approximate classifier cost \\
\midrule
IG & $N$ & target-score gradient
& $N(C_{\mathrm{fwd}}+C_{\mathrm{bwd}})$ \\
FRInGe & $T$ & tracking and target-score gradients; $M$ Fisher products
& $T(M+2)(C_{\mathrm{fwd}}+C_{\mathrm{bwd}})$ \\
\bottomrule
\end{tabularx}
\end{table}

Under the rough approximation $C_{\mathrm{fwd}}\approx C_{\mathrm{bwd}}$,
this gives
\begin{equation}
\frac{\mathcal C_{\mathrm{FRInGe}}}{\mathcal C_{\mathrm{IG}}}
\approx \frac{T}{N}(M+2).
\label{eq:complexity_ratios}
\end{equation}
The ratio counts classifier-dependent evaluations, not wall-clock time. With
spatial smoothing, FRInGe also applies the fixed regularizer inside CG
and to the current input; these operations do not evaluate the classifier.

\paragraph{Nominal waypoint bound.}
The largest Fisher--Rao distance to the uniform reference is approached by a
one-hot prediction:
\begin{equation}
D_{\max}=2\arccos\!\left(\frac{1}{\sqrt C}\right)<\pi.
\end{equation}
The nominal
update count is
\begin{equation}
T_{\max}
=
\left\lceil
\frac{2\arccos(1/\sqrt C)}{\sqrt{2\tau}}
\right\rceil
\leq
\left\lceil\frac{\pi}{\sqrt{2\tau}}\right\rceil.
\label{eq:tmax_complexity}
\end{equation}
The expression inside the first ceiling approaches the bound as
$C\to\infty$. Thus the nominal count is bounded with respect to the number of
classes, but still scales as $\tau^{-1/2}$. The realized trajectory may fail to
reach the nominal endpoint when the Euclidean cap is active; this affects
endpoint error, not the precomputed number of waypoints.

\paragraph{Runtime--fidelity controls.}
The KL budget $\tau$ controls waypoint resolution, while $K$ bounds the
categorical Fisher solve. Smaller $\tau$ increases path resolution and runtime.
Larger solver ceilings can reduce linear-system error but do not guarantee
better attribution quality. The benchmark solver uses a previous-waypoint
warm start and stopping tolerances, so the ceiling need not equal the realized
iteration count. The waypoint resolution and linear-solve accuracy thus
control different sources of approximation; neither increased runtime nor
a larger nominal iteration budget alone establishes better attributions.

\paragraph{FRInGe versus IG-300.}
Table~\ref{tab:runtime_sequential_fringe_vs_ig} compares categorical FRInGe
with 300-step Gauss--Legendre IG on the same 64 ImageNet inputs. Both methods
process one image at a time on an NVIDIA H100 GPU under matched preprocessing conditions. Gradient evaluations for IG's integration steps can be batched,
whereas FRInGe's tracking steps are sequential. The measured latency therefore
reflects both derivative work and the available parallelism; it is a
comparison of the evaluated configurations, not a universal speed ratio.

\begin{table}[H]
\centering
\small
\setlength{\tabcolsep}{5pt}
\caption{Per-image attribution latency under batch-size-one execution. Values
are means $\pm$ 95\% confidence intervals across 64 images. The slowdown is the
ratio of mean latencies.}
\label{tab:runtime_sequential_fringe_vs_ig}
\begin{tabular}{@{}l r@{\,\textpm\,}l r@{\,\textpm\,}l r@{\,\ensuremath{\times}}@{}}
\toprule
Model & \multicolumn{2}{c}{FRInGe (s)} & \multicolumn{2}{c}{IG-300 (s)} & \multicolumn{1}{c}{Slowdown} \\
\midrule
ResNet-18    &  55.12 & 0.80 &  2.90 & 0.01 & 19.0 \\
ResNet-50    &  45.00 & 0.41 &  9.53 & 0.03 &  4.7 \\
ResNet-101   & 228.21 & 1.58 & 11.59 & 0.04 & 19.7 \\
ResNet-152   & 319.21 & 2.89 & 15.28 & 0.01 & 20.9 \\
Inception-v3 & 159.80 & 3.58 & 14.94 & 0.03 & 10.7 \\
VGG-19       &  43.65 & 0.58 &  1.95 & 0.01 & 22.3 \\
\bottomrule
\end{tabular}
\end{table}

The table confirms that the categorical pullback solve is a substantial
practical cost. Appendix~\ref{app:fringeb} examines how the binary
specialization reduces this cost.

\section{Ablation Study Configurations}
\label{app:ablation}

The ablations ask which parts of the input-space update are needed in
addition to a prescribed predictive path. We first compare Euclidean
waypoint tracking, damped Fisher tracking without spatial penalties, and
FRInGe with two-term spatial regularization. We then separate the direction penalty from
the current-input prior and include a non-geometric prior control. The six
trajectories illustrate the consequences on one ResNet-18 input; the
dataset-level comparison in Table~\ref{tab:ablation_study} assesses whether
those visual differences correspond to improved metric values. All
variants retain predictive-waypoint tracking, so this study does not
ablate the presence of that tracking mechanism.

\subsection{Shared Protocol and Configuration Map}

The reconstructed trajectories in
Figures~\ref{fig:abla_max_entropy}--\ref{fig:abla_maxent_gamma_prior}
use a shared ResNet-18 hyperparameters, changing the indicated geometry
and spatial terms. Parameters associated with a disabled operator are
inactive. These illustrations should be distinguished from the
dataset-level ablation: its three core configurations use their
architecture-specific settings in Table~\ref{tab:fringe_hparams}, whereas
the additional prior-only configuration inherits FRInGe's
settings and disables only $\gamma_{\mathrm{step}}$, without retuning.

\begin{center}
\small
\setlength{\tabcolsep}{4pt}
\begin{tabularx}{\linewidth}{@{}Xccc@{}}
\toprule
Configuration & Fisher geometry & Direction penalty & Input prior \\
\midrule
Euclidean tracking & No & No & No \\
Fisher--Rao without spatial regularization & Yes & No & No \\
FRInGe & Yes & Yes & Yes \\
Direction-only FRInGe & Yes & Yes & No \\
Prior-only FRInGe & Yes & No & Yes \\
Euclidean tracking with prior & No & No & Yes \\
\bottomrule
\end{tabularx}
\end{center}

Each figure uses the same visual layout: normalized predictive entropy,
selected intermediate inputs, spatial blur-change maps relative to the
initial input, and the final signed attribution. Grey denotes zero
attribution, blue negative values, and red positive values. Blur-change
maps share a colour scale across the displayed steps within each figure,
but not across ablations. Their spatial pattern can therefore be compared
within a trajectory; colour intensity alone is not a common measure of
change across configurations.

\subsection{Geometry and Joint Spatial Regularization}

\begin{figure}[H]
    \centering
    \includegraphics[width=\textwidth]{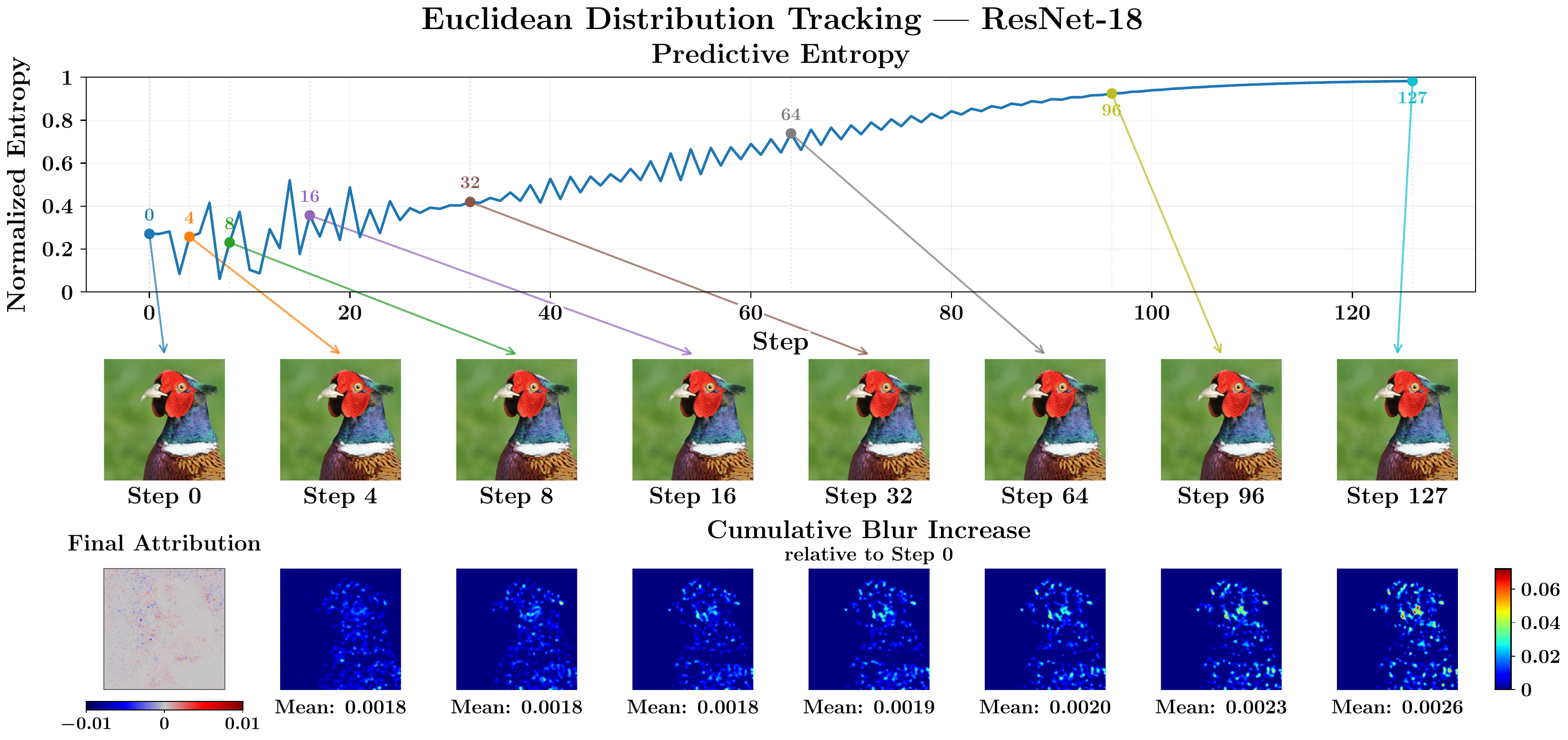}
    \caption{Euclidean predictive-waypoint tracking on one ResNet-18 input. The configuration omits Fisher conditioning and both spatial penalties.}
    \label{fig:abla_max_entropy}
\end{figure}

\paragraph{Euclidean distribution tracking.}
This control follows the spherical-loss gradient directly,
\[
v_k=g_k,
\]
without a Fisher solve, damping, or spatial penalty.
Figure~\ref{fig:abla_max_entropy} shows that predictive entropy can change
substantially while the intermediate inputs remain visually close to the
original. The blur-change maps and final attribution are speckled in this
example. Predictive-waypoint tracking alone therefore need not produce
spatially coherent image transformations; visual similarity of successive
inputs is not evidence that their predictions are unchanged.

\begin{figure}[H]
    \centering
    \includegraphics[width=\textwidth]{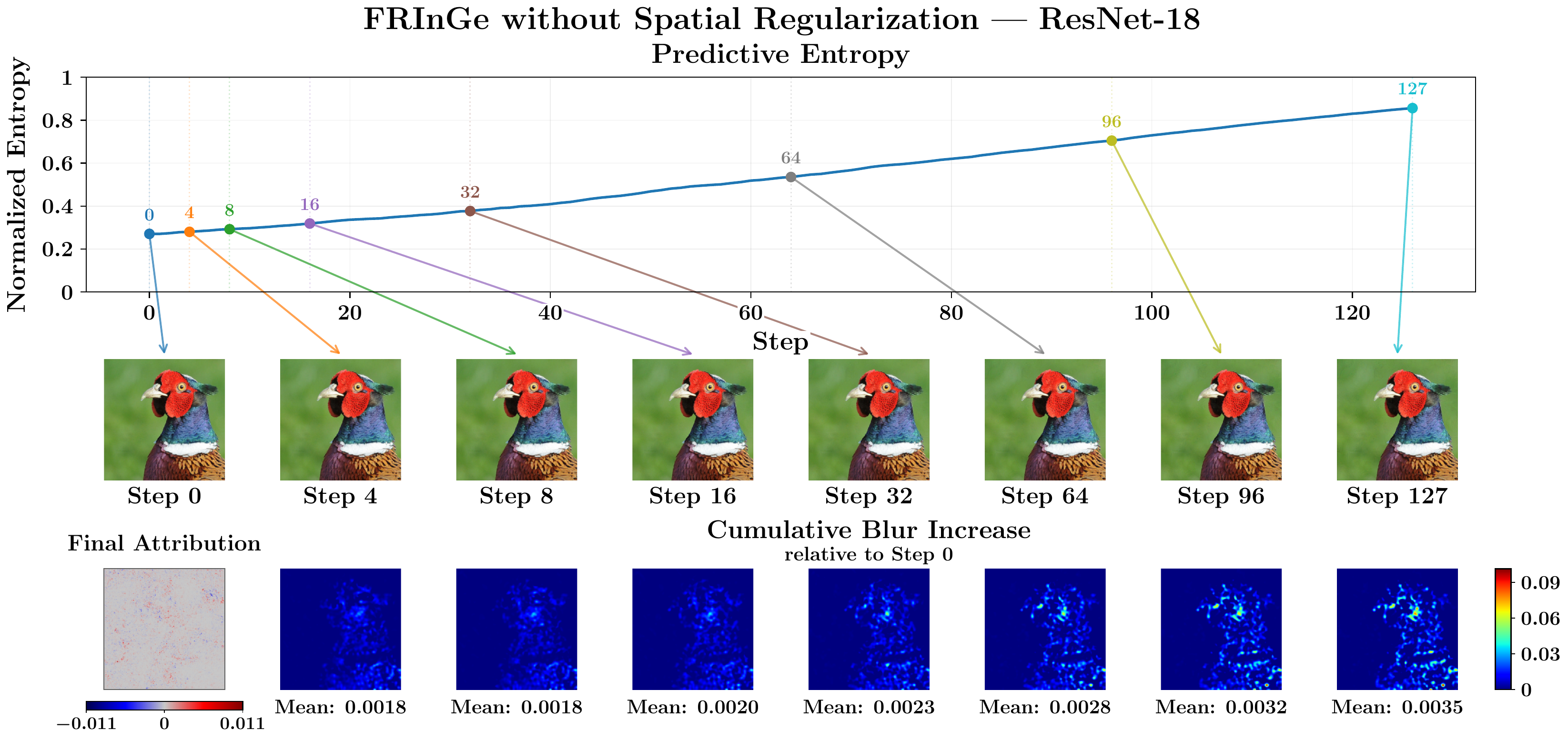}
    \caption{Damped Fisher--Rao tracking without spatial regularization on the same ResNet-18 input.}
    \label{fig:abla_unregularized}
\end{figure}

\paragraph{Fisher--Rao tracking without spatial regularization.}
This variant retains damping and predictive step control but removes both
spatial penalties:
\[
\bigl(G(x_k)+\lambda I\bigr)v_k=g_k.
\]
In Figure~\ref{fig:abla_unregularized}, entropy evolves more smoothly
than in the Euclidean example, but the spatial changes and final map
remain finely speckled. The example separates a smoother predictive
profile from a smoother input-space transformation: the former does not
by itself ensure the latter. It does not establish a general quality
advantage for Fisher conditioning alone.

\paragraph{FRInGe with two-term spatial regularization.}
The direction satisfies
\begin{equation}
\bigl(G(x_k)+\lambda I+\gamma_{\mathrm{step}}L^\top L\bigr)v_k
=
g_k+\gamma_{\mathrm{prior}}L^\top Lx_k,
\end{equation}
with the empirical blur preconditioner described in
Appendix~\ref{app:damping}. In Figure~\ref{fig:abla_full_fringe}, the
blur-change maps are more spatially organized and the final attribution
is less speckled. The changes concentrate on parts of the depicted object
while entropy increases. This is consistent with the spatial prior
shaping the numerical input path, but the appearance of one map does not
establish faithfulness; that interpretation must be assessed using the
quantitative metrics.

\begin{figure}[H]
    \centering
    \includegraphics[width=\textwidth]{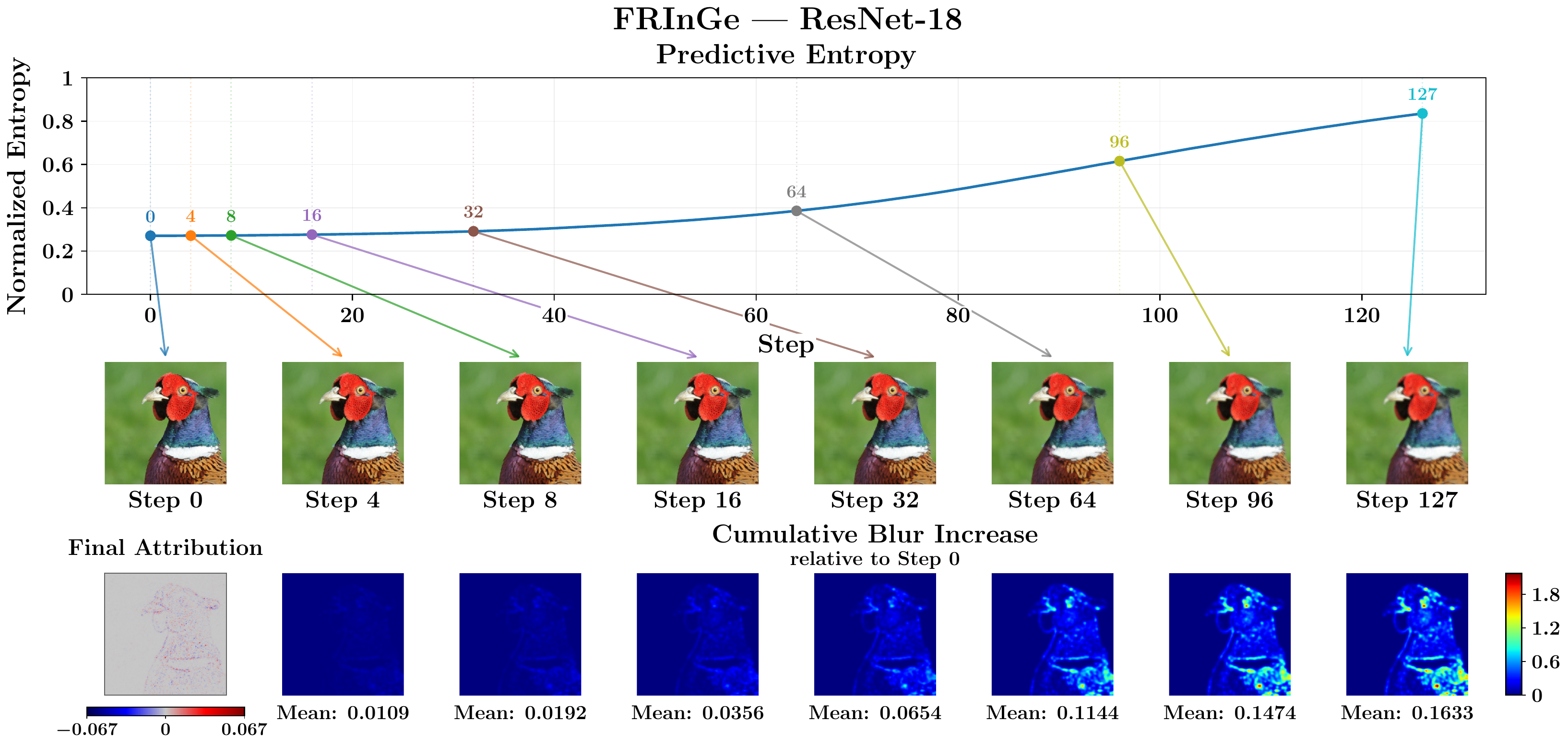}
    \caption{FRInGe on the same ResNet-18 input, combining Fisher geometry with direction regularization and a current-input prior.}
    \label{fig:abla_full_fringe}
\end{figure}

\subsection{Separating the Two Spatial Terms}

\begin{figure}[H]
    \centering
    \includegraphics[width=\textwidth]{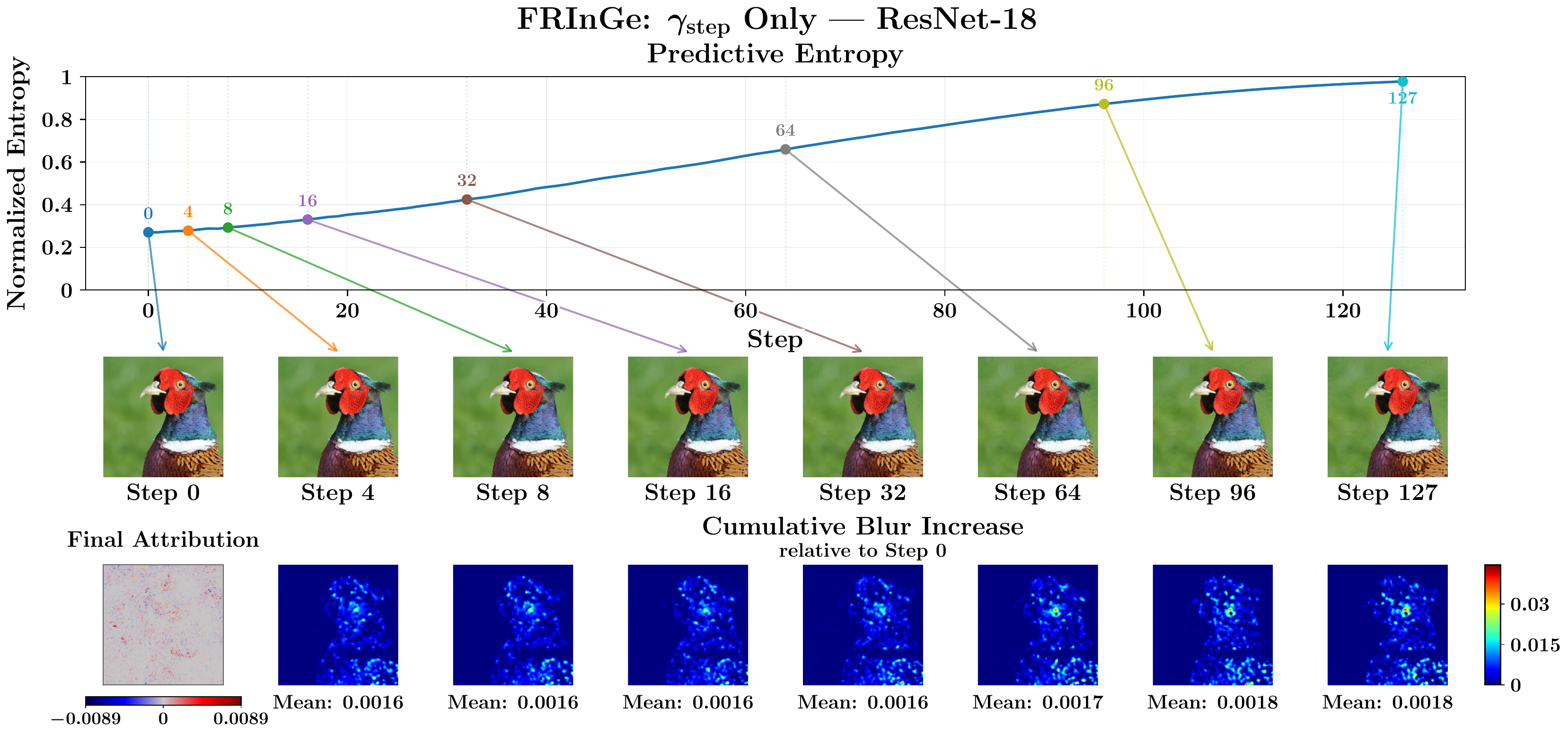}
    \caption{Direction-only FRInGe on the shared ResNet-18 input:
    $\gamma_{\mathrm{step}}>0$ and $\gamma_{\mathrm{prior}}=0$.
    The entropy profile is smooth, while spatial changes remain speckled.}
    \label{fig:abla_fringe_gamma_step}
\end{figure}

The coefficients affect different parts of the update. With
$\gamma_{\mathrm{prior}}=0$, the direction-only configuration solves
\begin{equation}
\bigl(G(x_k)+\lambda I+\gamma_{\mathrm{step}}L^\top L\bigr)v_k=g_k.
\end{equation}
It penalizes spatial variation in the direction rather than directly
encouraging a smoother current input.
Figure~\ref{fig:abla_fringe_gamma_step} shows a smooth entropy profile
but weak, speckled blur changes and a noisy final attribution for this
input. Thus, a regularized direction field need not yield the same
transformation as FRInGe with both spatial terms active.

\begin{figure}[H]
    \centering
    \includegraphics[width=\textwidth]{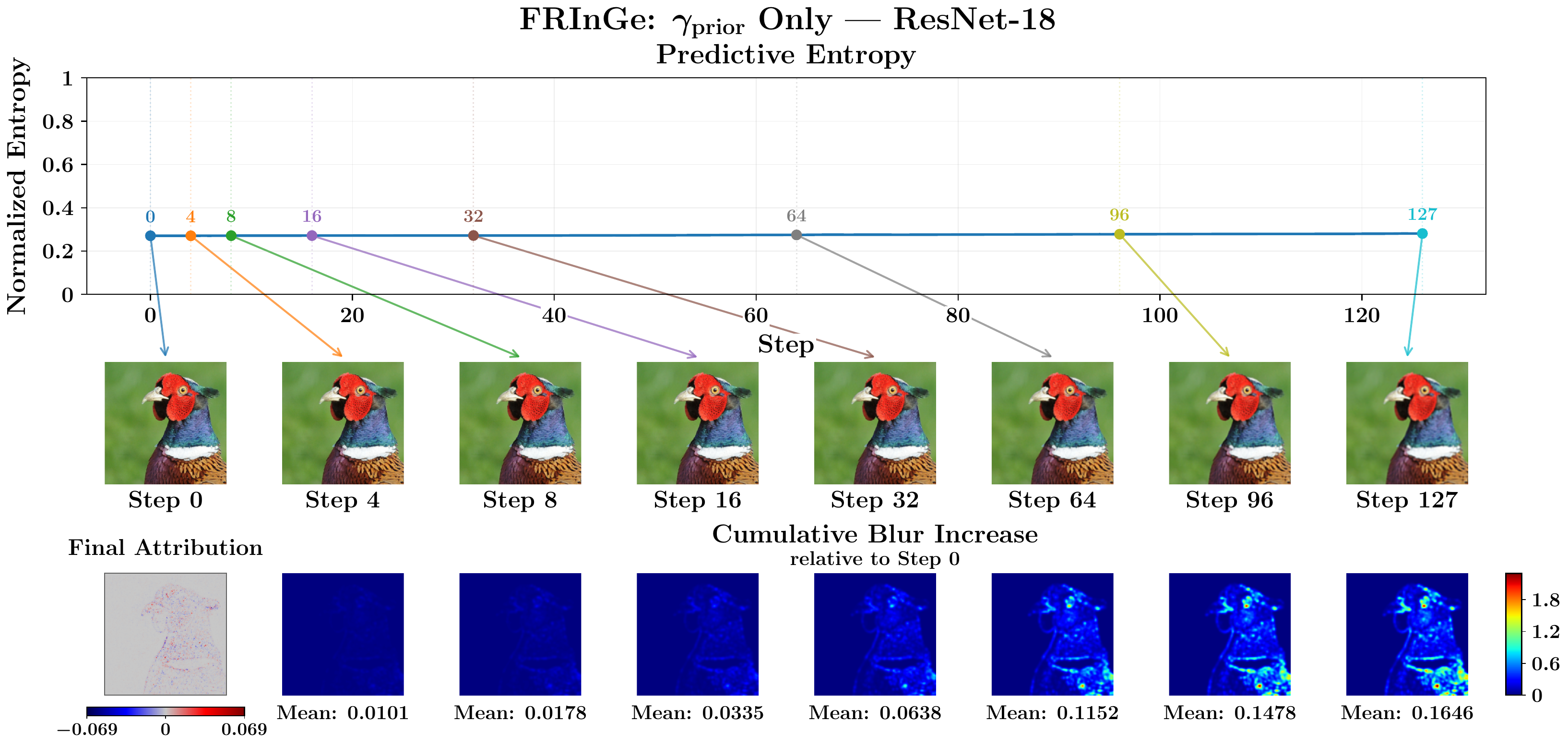}
    \caption{Prior-only FRInGe on the shared ResNet-18 input:
    $\gamma_{\mathrm{step}}=0$ and $\gamma_{\mathrm{prior}}>0$.
    Spatial changes concentrate despite limited entropy progress.}
    \label{fig:abla_fringe_gamma_prior}
\end{figure}

With $\gamma_{\mathrm{step}}=0$, the Fisher-aware prior-only
configuration instead solves
\begin{equation}
(G(x_k)+\lambda I)v_k=g_k+\gamma_{\mathrm{prior}}L^\top Lx_k.
\end{equation}
Under the update $x_{k+1}=x_k-\eta_kv_k$, the additional right-hand-side
term encourages removal of input roughness. In
Figure~\ref{fig:abla_fringe_gamma_prior}, the spatial changes become
concentrated, but normalized entropy increases by only approximately
$0.01$. This contrast illustrates why neither spatial coherence nor
entropy progress alone is sufficient to assess an attribution method.

\begin{figure}[H]
    \centering
    \includegraphics[width=\textwidth]{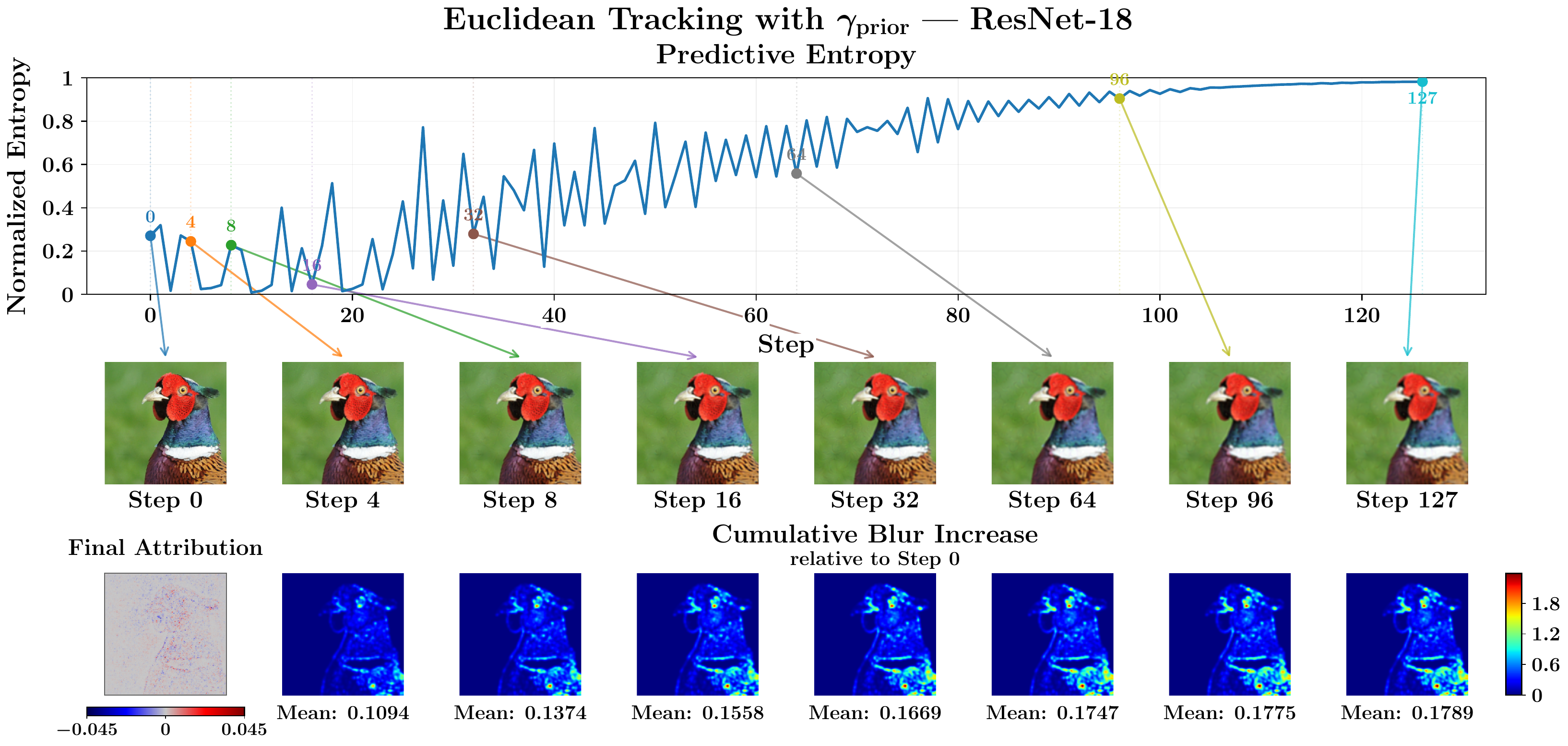}
    \caption{Euclidean waypoint tracking with a current-input prior on
    the shared ResNet-18 input. More organized blur-change maps coexist
    with a noisy final attribution.}
    \label{fig:abla_maxent_gamma_prior}
\end{figure}

Finally, the non-geometric prior control uses
\[
v_k = \nabla_x \mathcal{L}_{k+1}(x)\rvert_{x=x_k} + \gamma_{\mathrm{prior}} L^\top L x_k.
\]
Figure~\ref{fig:abla_maxent_gamma_prior} again shows more organized
blur-change maps than Euclidean tracking without the prior, but a
speckled final attribution. This control illustrates that structured
intermediate transformations need not produce a comparably structured
integrated explanation. Together, the examples motivate examining the
interaction of geometry and both spatial terms, rather than attributing
FRInGe's behavior to smoothing alone.

Table~\ref{tab:ablation_study} reports the corresponding dataset-level
evidence for the core configurations and Fisher-aware prior-only
variant. The single-image controls above explain how the components can
affect a trajectory; they do not isolate a universal contribution of
Fisher geometry under fully matched spatial regularization.
\FloatBarrier
\section{Experimental Protocol and Reproducibility}
\label{app:benchmark_details}

This section specifies the choices needed to interpret and reproduce the
reported comparisons. We describe the image cohorts, pretrained models,
fixed explanation targets, and baseline configurations before defining the metric-specific
responses, attribution reductions, and perturbation references. We then
explain the hyperparameter-selection objective, list the selected
architecture-specific settings, and examine their local sensitivity.
The distinctions matter: FRInGe explains a target logit, but a
perturbation metric may evaluate its target probability, and a magnitude
map used for ranking is not the signed tensor used by every metric.

\subsection{Dataset, Models, and Target Selection}

The original benchmark contains 1{,}001 ImageNet validation images and
six pretrained architectures: ResNet-18, ResNet-50, ResNet-101, ResNet-152,
Inception-v3, and VGG-19, each with its official preprocessing pipeline.
For each image $x$, the explained target is the original
top-1 class, $t=\arg\max_c F^c(x)$, held fixed throughout attribution and
evaluation. Methods compared on that image therefore address the same
target rather than following changing predictions along a perturbation
path.

\subsection{Attribution Baselines}

The baselines vary reference selection, path construction, and gradient
averaging. IG integrates along a straight-line path, whereas EG averages
eight 16-step Gauss--Legendre estimates using references sampled from the
fixed pool. SG averages gradients at noise-perturbed inputs.

GuidedIG adapts feature updates along its path, AGI uses adversarial trajectories,
and IG$^2$ constructs counterfactual-guided paths through representation
matching. BlurIG uses a 100-point Gaussian scale-space path from
$\sigma=10$ to the input.

GGIG adjusts interpolants using gradient ascent,
GeoIG constructs paths under a classifier-induced input geometry, and
MIG uses a learned generative manifold. The tested MIG configuration
uses a frozen pretrained VAE and a 12-point
latent path refined for at most three sweeps. Its endpoints are VAE
reconstructions, and it averages target-probability gradients rather than
computing a segment-wise integral of the target logit.

All results refer to the tested configurations, evaluated using the
metric conventions below.

\subsection{Evaluation Pipeline and Statistical Reporting}

For each image and method, the attribution is a tensor
$A\in\mathbb{R}^{3\times H\times W}$. The following overview distinguishes
the quantities derived from it. Perturbation scores assess the response
to a spatial ranking, MAS additionally assesses magnitude alignment,
infidelity compares predicted and observed local score changes, and
sparseness describes concentration.

\begin{center}
\small
\setlength{\tabcolsep}{3pt}
\begin{tabularx}{\linewidth}{@{}p{0.17\linewidth}p{0.17\linewidth}p{0.24\linewidth}X@{}}
\toprule
Metric & Model response & Attribution representation & Perturbation/reference \\
\midrule
Ins/Del AUC & Target probability & Maximum absolute channel value & Average-blurred image \\
MAS & Normalized target probability & Same ranking and cumulative mass & Gaussian-blurred image \\
Infidelity & Target-logit change & Original signed tensor & Gaussian input noise \\
Sparseness & Not applicable & All absolute tensor entries & No perturbation \\
Max Sensitivity & Not applicable & $\ell_2$-normalized tensor & Uniform input noise \\
\bottomrule
\end{tabularx}
\end{center}

Mean-score tables report the mean plus or minus half
the 95\% bootstrap interval width; percentile intervals need not be
symmetric about the mean. Max Sensitivity instead uses a median and
half-IQR, a dispersion summary rather than a confidence interval.
Higher MAS-Ins and Ins-AUC are preferred; lower MAS-Del, Del-AUC,
Infidelity, and Max Sensitivity are preferred. Higher sparseness means
greater concentration, not necessarily greater faithfulness.

\subsection{Metric Definitions}
\label{app:metric_definitions}
\subsubsection{Spatial Saliency Reduction}

Metrics based on progressive perturbations require a spatial ranking rather than a channel-wise attribution tensor. Given \(A\in\mathbb{R}^{3\times H\times W}\), we define the spatial saliency map
\[
S_{ij}=\max_{c\in\{1,2,3\}} |A_{cij}|,
\]
where \(S_{ij}\) is the saliency assigned to pixel \((i,j)\). Pixels are ranked in descending order of \(S_{ij}\), and this ranking is used in all perturbation-based evaluations.

\subsubsection{Causal Faithfulness: Blur-Based Insertion and Deletion AUC}

We follow the insertion/deletion protocol of Petsiuk et al.~\cite{petsiuk2018rise}.

The reference $x_{\mathrm{blur}}$ is a local average of the input with
an odd window size proportional to image resolution:
\[
k=\max\!\left(3,\; \mathrm{odd}\!\left(\left\lfloor \frac{\min(H,W)}{10}\right\rfloor\right)\right).
\]
Here $\mathrm{odd}$ rounds upward to the next odd integer, including
its argument when that argument is already odd.

At perturbation step \(s\), let \(x_s\) be the perturbed image and
\[
c_s=\softmax(F(x_s))_t
\]
the target-class confidence. Insertion starts from the blurred reference
and restores pixels in saliency order; deletion starts from the original
and replaces them by their reference values in the same order. Their AUCs
measure how quickly confidence is recovered or removed as the fraction of
replaced pixels increases.

\begin{samepage}
The normalized confidence curve is
\[
\widetilde{c}_s
=
\mathrm{clip}\!\left(
\frac{c_s-p_{\mathrm{blur}}}{\max\!\bigl(p_{\mathrm{orig}}-p_{\mathrm{blur}},\varepsilon\bigr)},
0,1
\right),
\]
where
\[
p_{\mathrm{orig}}=\softmax(F(x))_t,
\qquad
p_{\mathrm{blur}}=\softmax(F(x_{\mathrm{blur}}))_t.
\]
\end{samepage}
We integrate this curve using the trapezoidal rule over the
replaced-pixel fraction. The anchors are the target confidences on the original and blurred
inputs. This rescales the curve by their confidence gap and clips it
to $[0,1]$; $\varepsilon>0$ prevents a zero denominator. The blur here
is distinct from the Gaussian reference used by MAS below.

\subsubsection{Magnitude Aligned Scoring (MAS)}

MAS~\cite{walker2024MAS} asks whether attribution magnitudes, not only
their ordering, agree with the classifier's confidence response.
It uses the same spatial saliency $S_{ij}$ but a Gaussian-blurred
reference with a $15\times15$ kernel and standard deviation $\sigma=3.0$.
The target is again the original top-1 class. An image is eligible only
when its original target confidence exceeds the Gaussian-reference
confidence by more than $10^{-6}$. This eligibility condition depends on
the model, image, and baseline, not on the attribution method.

Let $\mathcal I_s$ be the pixels restored or removed after step $s$,
with $s=0,\dots,N$. The density responses represent the fraction of
attribution mass restored by insertion or remaining after deletion:
\[
DR_s^{\mathrm{ins}}
=
\frac{\sum_{(i,j)\in\mathcal I_s}S_{ij}}
{\max\!\left(\sum_{i,j}S_{ij},\varepsilon\right)},
\qquad
DR_s^{\mathrm{del}}=1-DR_s^{\mathrm{ins}}.
\]
The model response first uses the confidence-gap normalization defined
above, with the Gaussian MAS reference in place of the average-blurred
reference. Its insertion curve is replaced by its cumulative maximum,
and its deletion curve by its cumulative minimum. Denote these
monotone normalized responses by $MR_s^{\mathrm{ins}}$ and
$MR_s^{\mathrm{del}}$. Their discrepancy from the density response is
\[
AP_s^{m}=\left|MR_s^{m}-DR_s^{m}\right|,
\qquad m\in\{\mathrm{ins},\mathrm{del}\}.
\]
The pointwise scores are
\[
z_s^{\mathrm{ins}}
=\mathrm{clip}\!\left(MR_s^{\mathrm{ins}}-AP_s^{\mathrm{ins}},0,1\right),
\qquad
z_s^{\mathrm{del}}
=\mathrm{clip}\!\left(MR_s^{\mathrm{del}}+AP_s^{\mathrm{del}},0,1\right).
\]
For either mode, the reported MAS is the trapezoidal area under this
score curve over the perturbation steps:
\[
MAS^{m}
=
\frac{1}{N}\left(
\frac{z_0^{m}+z_N^{m}}{2}
+\sum_{s=1}^{N-1}z_s^{m}
\right).
\]
Higher MAS-Insertion rewards rapid confidence recovery with aligned
attribution mass; lower MAS-Deletion rewards rapid removal with aligned
remaining mass. These are attribution magnitude-alignment scores, not
measures of classification-probability calibration. Summaries exclude
ineligible images, so their sample sizes can differ from those of the
other metrics even within the same cohort.

\subsubsection{Local Robustness: Infidelity}

Infidelity~\cite{yeh2019fidelity} measures how well an attribution predicts the change in target score under small random perturbations. For attribution \(\Phi(x)\) and perturbation \(\delta\), it is defined as
\[
\mathbb{E}_{\delta}\!\left[
\left(
\delta^\top \Phi(x) - \bigl(F^t(x)-F^t(x-\delta)\bigr)
\right)^2
\right].
\]
The first term, \(\delta^\top \Phi(x)\), is the attribution-predicted score variation under perturbation \(\delta\), while the second term is the actual change in model output. Lower infidelity indicates better local faithfulness. We estimate this expectation by Monte Carlo sampling with 50 Gaussian perturbations per image, each drawn with standard deviation \(\sigma=0.02\).

\subsubsection{Explanation Focus: Sparseness}

We quantify explanation focus using the Gini index of all absolute
attribution entries~\cite{hurley2009comparing}. For
$A\in\mathbb{R}^{3\times H\times W}$, write
\[
a=\bigl(|A_1|,\dots,|A_n|\bigr)\in\mathbb{R}_{\ge 0}^n,
\qquad n=3\cdot H\cdot W.
\]
Let \(a_{(1)}\le a_{(2)}\le \dots \le a_{(n)}\) denote these values sorted in ascending order. The sparseness score is then computed as
\[
\mathrm{Sparseness}(A)
=
\frac{2\sum_{i=1}^{n} i\,a_{(i)}}{n\sum_{i=1}^{n} a_{(i)}+\varepsilon}
-\frac{n+1}{n},
\]
with the reported score clipped to $[0,1]$.

This score is the Gini index of the attribution magnitude distribution. Values near \(0\) indicate a diffuse or nearly uniform attribution map, whereas values near \(1\) indicate that attribution mass is concentrated on a small number of entries. This metric is reported only as a complementary descriptive statistic and is not used directly in the causal-faithfulness or calibration evaluations above.
\subsubsection{Local Robustness: Max Sensitivity}

We additionally report Max Sensitivity~\cite{yeh2019fidelity} as a robustness diagnostic under small
input perturbations. For a given input \(x\), fixed target class \(t\), and
attribution method \(\Phi\), we compute the attribution map and normalize it by
its \(\ell_2\) norm so that the score reflects changes in explanation structure
rather than overall scale:
\[
\widehat{\Phi}(x)=\frac{\Phi(x)}{\|\Phi(x)\|_2+\varepsilon}.
\]

We then sample perturbations \(\delta^{(1)},\dots,\delta^{(M)}\) independently,
with entries drawn uniformly from \([-r,r]\), and recompute the normalized
attribution for each perturbed input. In our experiments, \(r=0.02\) and
\(M=10\). The reported score is
\[
\mathrm{MaxSens}(x)
=
\max_{m=1,\dots,M}
\frac{\left\|\widehat{\Phi}(x)-\widehat{\Phi}(x+\delta^{(m)})\right\|_2}
{\|\delta^{(m)}\|_2+\varepsilon}.
\]

Lower values indicate greater local robustness. Since the maximum is taken over
a finite Monte Carlo sample, this is a sampled approximation of local sensitivity
rather than an exact worst-case bound.

\subsection{Hyperparameter Selection Protocol}

Architecture-specific settings for the three core configurations were
selected using Optuna~\cite{akiba2019optuna}. The search includes the
trust-region parameters $\tau$, $\eta_{\max}$, and
$\delta_{\mathrm{euc}}$, damping $\lambda$, and, where applicable, the
two spatial coefficients. The added prior-only ablation is not a
separately optimized method: it inherits FRInGe's tuned hyperparameters
and sets $\gamma_{\mathrm{step}}=0$.

The optimization objective was a weighted harmonic mean of three quantities: blur-based insertion AUC, inverted blur-based deletion AUC, and an infidelity-derived fidelity score. Let \(\mathrm{InsAUC}\) and \(\mathrm{DelAUC}\) denote the insertion and deletion AUC values on the tuning subset, and let \(\mathrm{Inf}\) denote the infidelity score. We define $\mathrm{Fid}=\exp(-k\,\mathrm{Inf}), k=2$
and optimize
\begin{equation}
\mathrm{Score}
=
\frac{w_{\mathrm{ins}}+w_{\mathrm{del}}+w_{\mathrm{fid}}}
{\frac{w_{\mathrm{ins}}}{\mathrm{InsAUC}+\varepsilon}
+
\frac{w_{\mathrm{del}}}{(1-\mathrm{DelAUC})+\varepsilon}
+
\frac{w_{\mathrm{fid}}}{\mathrm{Fid}+\varepsilon}},
\end{equation}
with weights $w_{\mathrm{ins}}=1,
w_{\mathrm{del}}=1,
w_{\mathrm{fid}}=0.5$,
and $\varepsilon=10^{-6}$. The harmonic mean discourages a high score on
one component from compensating for poor performance on another. MAS and
sparseness are not components of this objective, so the selected
configuration is not directly optimized for every reported metric.
Because hyperparameters were tuned on a subset of the validation cohort rather than a disjoint split, the reported results should be interpreted as benchmark comparisons rather than a held-out evaluation of hyperparameter generalization.

\subsection{Architecture-Specific Hyperparameters}

Table~\ref{tab:fringe_hparams} lists the selected hyperparameters for
the three core configurations. Euclidean tracking bypasses the linear
system, so damping, Fisher-solve, and spatial-penalty parameters are
inapplicable. Fisher--Rao tracking without spatial regularization
(\emph{Unreg.\ FR}) retains the damped Fisher system but disables both
spatial terms; FRInGe uses both terms
in the system in Appendix~\ref{app:damping}. The prior-only variant
inherits the FRInGe row and sets only $\gamma_{\mathrm{step}}=0$,
so those otherwise duplicated rows are omitted. In the main benchmark runs for FRInGe, each conjugate-gradient solve allows at most
25 iterations per tracking step. Metric settings are held fixed within each
comparison and are distinct from these optimization parameters.

\begin{table}[H]
\centering
\small
\setlength{\tabcolsep}{3pt}
\renewcommand{\arraystretch}{1.12}
\caption{Architecture-specific ablation hyperparameters. Dashes indicate inapplicable parameters.}
\label{tab:fringe_hparams}
\begin{tabular*}{\textwidth}{@{\extracolsep{\fill}}llcccccc@{}}
\toprule
\textbf{Model} & \textbf{Variant} & $\tau$ & $\eta_{\max}$ & $\delta_{\mathrm{euc}}$ & $\lambda$ & $\gamma_{\mathrm{step}}$ & $\gamma_{\mathrm{prior}}$ \\
\midrule
\multirow{3}{*}{ResNet-18}
& Euclidean   & --       & 29.58 & --    & --        & --       & -- \\
& Unreg.\ FR  & 3.03e-4  & 1.98  & 34.52 & 2.77e-11  & --       & -- \\
& FRInGe & 3.03e-4  & 13.56 & 0.61  & 4.77e-11  & 9.97e-3  & 9.75e-4 \\
\midrule
\multirow{3}{*}{ResNet-50}
& Euclidean   & --       & 29.70 & --    & --        & --       & -- \\
& Unreg.\ FR  & 3.21e-4  & 1.13  & 18.32 & 1.28e-10  & --       & -- \\
& FRInGe & 2.70e-3  & 1.18  & 5.01  & 2.58e-4   & 9.94e-3  & 9.49e-4 \\
\midrule
\multirow{3}{*}{ResNet-101}
& Euclidean   & --       & 29.13 & --    & --        & --       & -- \\
& Unreg.\ FR  & 6.43e-4  & 1.02  & 14.48 & 1.12e-12  & --       & -- \\
& FRInGe & 3.69e-4  & 1.53  & 3.85  & 8.02e-8   & 9.98e-3  & 6.08e-4 \\
\midrule
\multirow{3}{*}{ResNet-152}
& Euclidean   & --       & 0.19  & --    & --        & --       & -- \\
& Unreg.\ FR  & 3.29e-4  & 1.06  & 26.35 & 3.15e-10  & --       & -- \\
& FRInGe & 4.13e-4  & 1.75  & 3.32  & 7.55e-5   & 6.06e-3  & 9.16e-4 \\
\midrule
\multirow{3}{*}{Inception-v3}
& Euclidean   & --       & 6.28  & --    & --        & --       & -- \\
& Unreg.\ FR  & 1.97e-3  & 2.81  & 3.28  & 9.79e-3   & --       & -- \\
& FRInGe & 5.34e-4  & 16.92 & 24.18 & 7.17e-6   & 9.73e-3  & 3.43e-4 \\
\midrule
\multirow{3}{*}{VGG-19}
& Euclidean   & --       & 0.02  & --    & --        & --       & -- \\
& Unreg.\ FR  & 3.22e-4  & 1.11  & 9.39  & 2.17e-12  & --       & -- \\
& FRInGe & 3.04e-4  & 2.75  & 0.70  & 3.03e-11  & 5.13e-3  & 9.62e-4 \\
\bottomrule
\end{tabular*}
\end{table}

\subsection{Hyperparameter Sensitivity Across Architectures}
\label{app:hyper_sens_anal}

\begin{figure}[H]
    \centering
    \includegraphics[width=\textwidth]{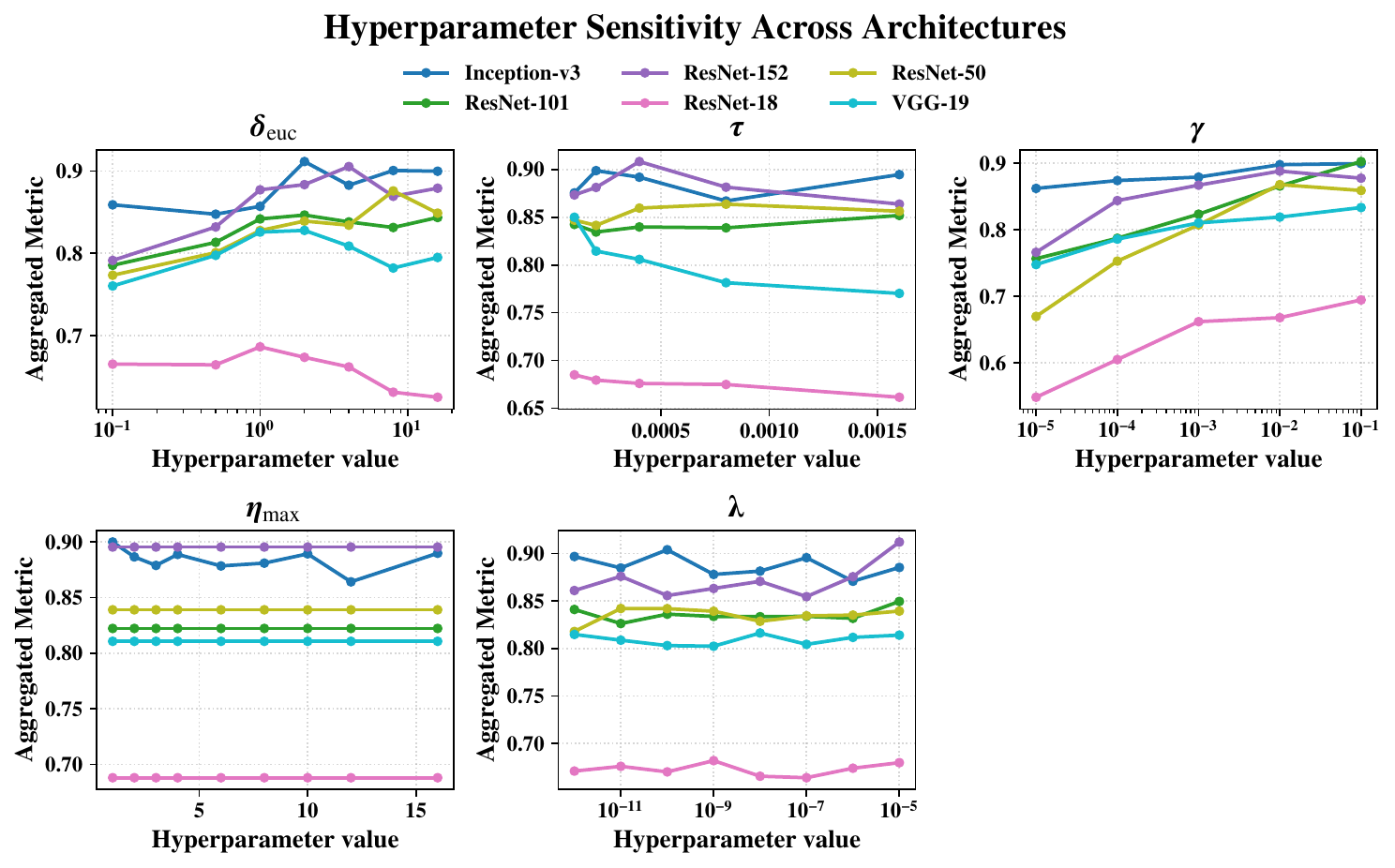}
    \caption{Sensitivity analysis of FRInGe hyperparameters across all evaluated architectures. Each panel varies one hyperparameter while keeping the others fixed and reports the aggregated tuning objective. For the smoothing sweep, we vary the regularization strength jointly, setting \(\gamma_{\mathrm{step}}=\gamma_{\mathrm{prior}}=\gamma\).}
    \label{fig:hyperparam_sensitivity}
\end{figure}

The selected hyperparameters do not show how sensitive the method is
to nearby choices. Figure~\ref{fig:hyperparam_sensitivity} therefore
varies one parameter at a time while holding the others fixed and reports
the same weighted harmonic-mean objective used for selection. The spatial
sweep is an explicit exception: it ties
$\gamma_{\mathrm{step}}=\gamma_{\mathrm{prior}}=\gamma$ and varies both
coefficients jointly. It does not measure their separate effects.

Over the displayed ranges, the objective is relatively insensitive to
$\eta_{\max}$ and more architecture-dependent for $\tau$ and $\lambda$.
A permissive maximum step size can be inactive when another cap is
tighter, whereas $\tau$ also changes the nominal waypoint count.
Damping affects the computed direction and the numerical solve; its
role should not be inferred from a single architecture's curve.

The more pronounced changes occur for the Euclidean radius and tied
spatial strength. These parameters directly alter input-segment lengths
and the spatial character of the path. Small radii can restrict progress
within the available updates, while large radii can leave gradient
variation unresolved by quadrature. The smoothing sweep illustrates the
benefit and eventual saturation of the tested low-frequency prior,
without establishing that larger regularization is universally better.

These are conditional, one-factor sweeps, not a complete analysis of
parameter interactions or a new selection rule. In particular, the tied
spatial sweep cannot replace the separate-component controls in
Appendix~\ref{app:ablation}.

\section{Additional Quantitative Results}
\label{app:tables}

This section complements the main results in three stages. The component
ablation compares the roles of geometry and spatial regularization;
the architecture-specific tables extend the attribution-method comparison;
and a separate perturbation study assesses local explanation stability.
These answer different questions, so concentration, magnitude alignment,
perturbation faithfulness, and robustness are interpreted separately.

Unless a table states otherwise, entries are means plus or minus half the
95\% image-bootstrap interval width. The best displayed mean within each
comparison is bold and underlined; bold overlap highlighting compares the
displayed ranges and is descriptive, not a test of equivalence or a paired
significance test. A symmetric displayed range need not reproduce the
endpoints of an asymmetric percentile interval. Max Sensitivity uses
median plus or minus half-IQR instead. The metric definitions and
normalization conventions are given in Appendix~\ref{app:metric_definitions}.

\subsection{Ablation Study}

Table~\ref{tab:ablation_study} evaluates the three core configurations and
the added Fisher-aware prior-only variant across all six architectures.
The latter retains FRInGe's damping, step controls, prior
coefficient, and preconditioner, changing only
$\gamma_{\mathrm{step}}=0$; it is not separately retuned. The core
Euclidean tracking, Fisher tracking without spatial regularization, and FRInGe use their respective selected hyperparameters. Their comparison therefore
evaluates complete configurations, not geometry alone under matched
regularization.

FRInGe has the highest displayed MAS-Insertion and lowest
MAS-Deletion mean on each architecture. The prior-only variant improves Ins-AUC and Del-AUC over the unregularized configurations and has the lower Del-AUC mean on ResNet-50 and ResNet-152, while FRInGe
has the higher insertion-AUC mean throughout. Fisher tracking without spatial
regularization is more sparse without having better MAS. This contrast shows
why concentration cannot substitute for attribution-quality evaluation.
The illustrative trajectories and component definitions appear in
Appendix~\ref{app:ablation}.

\begin{table*}[hp]
\centering
\footnotesize
\setlength{\tabcolsep}{0pt}
\renewcommand{\arraystretch}{1.08}
\caption{Ablation study of FRInGe components across evaluated architectures. \textbf{Euclid.} ignores manifold curvature, \textbf{Unreg.\ FR} retains damping and pullback geometry but disables both spatial terms, and \textbf{Prior-only} fixes $\gamma_{\mathrm{step}}=0$ using FRInGe's tuned hyperparameters without separate retuning. \textbf{FRInGe} uses both spatial terms. Best results are in \best{bold underline}; methods with overlapping confidence intervals are in \tie{bold}.}
\label{tab:ablation_study}
\begin{tabular*}{\textwidth}{@{\extracolsep{\fill}}l cccc @{\hspace{10pt}} cccc@{}}
\toprule
& \multicolumn{4}{c}{\textbf{ResNet-18}} & \multicolumn{4}{c}{\textbf{ResNet-50}} \\
\cmidrule{2-5} \cmidrule{6-9}
\textbf{Metric}
& \textbf{Euclid.} & \textbf{Unreg.\ FR} & \textbf{Prior-only} & \textbf{FRInGe}
& \textbf{Euclid.} & \textbf{Unreg.\ FR} & \textbf{Prior-only} & \textbf{FRInGe} \\
\midrule
MAS-Ins $\uparrow$
& \statentry{0.414}{.017} & \statentry{0.399}{.017} & \statentry{0.461}{.016} & \best{\statentry{0.597}{.015}}
& \statentry{0.491}{.018} & \statentry{0.477}{.018} & \statentry{0.581}{.014} & \best{\statentry{0.648}{.014}} \\
MAS-Del $\downarrow$
& \statentry{0.319}{.012} & \statentry{0.330}{.012} & \statentry{0.324}{.010} & \best{\statentry{0.264}{.010}}
& \statentry{0.361}{.014} & \statentry{0.372}{.014} & \statentry{0.326}{.010} & \best{\statentry{0.302}{.012}} \\
Ins-AUC $\uparrow$
& \statentry{0.358}{.014} & \statentry{0.347}{.013} & \statentry{0.409}{.014} & \best{\statentry{0.492}{.016}}
& \statentry{0.415}{.014} & \statentry{0.403}{.014} & \tie{\statentry{0.560}{.015}} & \best{\statentry{0.581}{.016}} \\
Del-AUC $\downarrow$
& \statentry{0.195}{.010} & \statentry{0.207}{.010} & \statentry{0.165}{.009} & \best{\statentry{0.130}{.008}}
& \statentry{0.246}{.012} & \statentry{0.261}{.012} & \best{\statentry{0.152}{.009}} & \tie{\statentry{0.161}{.009}} \\
Sparseness $\uparrow$
& \statentry{0.764}{.003} & \best{\statentry{0.785}{.003}} & \statentry{0.661}{.002} & \statentry{0.706}{.002}
& \statentry{0.763}{.003} & \best{\statentry{0.791}{.004}} & \statentry{0.605}{.003} & \statentry{0.702}{.002} \\
Infidelity $\downarrow$
& \best{\statentry{0.058}{.008}} & \best{\statentry{0.058}{.008}} & \best{\statentry{0.058}{.008}} & \best{\statentry{0.058}{.008}}
& \best{\statentry{0.149}{.050}} & \best{\statentry{0.149}{.050}} & \tie{\statentry{0.152}{.050}} & \tie{\statentry{0.150}{.050}} \\
\midrule

& \multicolumn{4}{c}{\textbf{ResNet-101}} & \multicolumn{4}{c}{\textbf{ResNet-152}} \\
\cmidrule{2-5} \cmidrule{6-9}
MAS-Ins $\uparrow$
& \statentry{0.568}{.017} & \statentry{0.554}{.018} & \statentry{0.598}{.014} & \best{\statentry{0.679}{.014}}
& \statentry{0.591}{.017} & \statentry{0.581}{.018} & \statentry{0.624}{.013} & \best{\statentry{0.691}{.013}} \\
MAS-Del $\downarrow$
& \statentry{0.421}{.017} & \statentry{0.432}{.017} & \statentry{0.389}{.014} & \best{\statentry{0.351}{.015}}
& \statentry{0.451}{.018} & \statentry{0.464}{.018} & \tie{\statentry{0.403}{.015}} & \best{\statentry{0.392}{.017}} \\
Ins-AUC $\uparrow$
& \statentry{0.499}{.015} & \statentry{0.488}{.015} & \statentry{0.595}{.015} & \best{\statentry{0.641}{.017}}
& \statentry{0.531}{.015} & \statentry{0.519}{.015} & \tie{\statentry{0.666}{.015}} & \best{\statentry{0.681}{.016}} \\
Del-AUC $\downarrow$
& \statentry{0.309}{.014} & \statentry{0.321}{.014} & \best{\statentry{0.228}{.012}} & \best{\statentry{0.228}{.012}}
& \statentry{0.351}{.014} & \statentry{0.362}{.014} & \best{\statentry{0.240}{.012}} & \tie{\statentry{0.265}{.013}} \\
Sparseness $\uparrow$
& \statentry{0.772}{.003} & \best{\statentry{0.795}{.004}} & \statentry{0.634}{.003} & \statentry{0.706}{.002}
& \statentry{0.779}{.003} & \best{\statentry{0.799}{.004}} & \statentry{0.589}{.003} & \statentry{0.703}{.002} \\
Infidelity $\downarrow$
& \best{\statentry{0.121}{.022}} & \best{\statentry{0.121}{.022}} & \tie{\statentry{0.125}{.022}} & \best{\statentry{0.121}{.022}}
& \best{\statentry{0.111}{.020}} & \best{\statentry{0.111}{.020}} & \tie{\statentry{0.113}{.020}} & \best{\statentry{0.111}{.020}} \\
\midrule

& \multicolumn{4}{c}{\textbf{Inception-v3}} & \multicolumn{4}{c}{\textbf{VGG-19}} \\
\cmidrule{2-5} \cmidrule{6-9}
MAS-Ins $\uparrow$
& \statentry{0.688}{.015} & \statentry{0.681}{.015} & \statentry{0.676}{.010} & \best{\statentry{0.761}{.009}}
& \statentry{0.427}{.017} & \statentry{0.413}{.017} & \statentry{0.478}{.016} & \best{\statentry{0.588}{.015}} \\
MAS-Del $\downarrow$
& \tie{\statentry{0.467}{.019}} & \tie{\statentry{0.483}{.019}} & \tie{\statentry{0.464}{.017}} & \best{\statentry{0.444}{.020}}
& \statentry{0.319}{.012} & \statentry{0.329}{.012} & \statentry{0.309}{.010} & \best{\statentry{0.255}{.008}} \\
Ins-AUC $\uparrow$
& \statentry{0.580}{.013} & \statentry{0.557}{.013} & \statentry{0.672}{.013} & \best{\statentry{0.709}{.014}}
& \statentry{0.389}{.015} & \statentry{0.382}{.015} & \statentry{0.444}{.015} & \best{\statentry{0.503}{.016}} \\
Del-AUC $\downarrow$
& \statentry{0.300}{.013} & \statentry{0.315}{.013} & \tie{\statentry{0.241}{.012}} & \best{\statentry{0.240}{.012}}
& \statentry{0.193}{.010} & \statentry{0.204}{.010} & \statentry{0.157}{.008} & \best{\statentry{0.119}{.007}} \\
Sparseness $\uparrow$
& \statentry{0.763}{.003} & \best{\statentry{0.801}{.003}} & \statentry{0.617}{.003} & \statentry{0.723}{.003}
& \tie{\statentry{0.807}{.003}} & \best{\statentry{0.814}{.006}} & \statentry{0.691}{.003} & \statentry{0.707}{.003} \\
Infidelity $\downarrow$
& \best{\statentry{0.020}{.005}} & \best{\statentry{0.020}{.005}} & \tie{\statentry{0.026}{.005}} & \best{\statentry{0.020}{.005}}
& \best{\statentry{0.343}{.064}} & \best{\statentry{0.343}{.064}} & \tie{\statentry{0.344}{.064}} & \best{\statentry{0.343}{.064}} \\
\bottomrule
\end{tabular*}
\end{table*}

\subsection{Comparison across attribution methods}
\label{app:method_comparison}
\label{app:eg_blurig}
\label{app:mig_comparison}

Tables~\ref{tab:full_inception_v3}--\ref{tab:full_vgg19} compare FRInGe
with all ten attribution baselines across six architectures. Method
settings, cohort differences, and metric conventions are specified in
Appendix~\ref{app:benchmark_details}. AGI denotes Adversarial IG and SG
denotes SmoothGrad.

The comparison reveals distinct strengths rather than a single uniformly
best method. FRInGe has the highest mean MAS-Insertion on most
architectures, while MAS-Deletion favors different methods depending on
the architecture. Among the methods in these tables, BlurIG leads mean Ins-AUC and Del-AUC, whereas IG$^2$ produces the most concentrated maps. Infidelity
intervals overlap for many methods, so their displayed rankings should
not be interpreted as clear statistical separation. Together, these
results distinguish attribution magnitude alignment from feature ordering
and concentration; an advantage in one does not establish superiority
under the others.

FRInGe's spatial regularization and BlurIG's Gaussian scale-space path
share a low-frequency inductive bias, although their trajectories differ.
This may help explain their competitive results on these image models.
FRInGe additionally conditions the trajectory on the classifier's
pullback Fisher geometry and predictive waypoints, consistent with its
strong MAS-Insertion results. BlurIG's blur-based AUC advantage may also
partly reflect alignment between its path and the evaluation
perturbations. These are plausible interpretations, not established
causal explanations; testing the latter would require other perturbation
references.

\begin{table}[H]
    \centering
    \small
    \setlength{\tabcolsep}{1pt}
    \renewcommand{\arraystretch}{1.15}
    \caption{Quantitative evaluation on Inception-v3. Means $\pm$ half-widths of 95\% image-bootstrap intervals; descriptive highlighting.}
    \label{tab:full_inception_v3}
    \begin{tabular*}{\textwidth}{@{\extracolsep{\fill}}lcccccc@{}}
        \toprule
        & \multicolumn{2}{c}{\textit{Attribution calibration}}
        & \multicolumn{2}{c}{\textit{Causal faithfulness}}
        & \multicolumn{2}{c}{\textit{Focus and robustness}} \\
        \cmidrule(lr){2-3}\cmidrule(lr){4-5}\cmidrule(lr){6-7}
        \textbf{Method} & \textbf{MAS-Ins} $\uparrow$ & \textbf{MAS-Del} $\downarrow$ & \textbf{Ins-AUC} $\uparrow$ & \textbf{Del-AUC} $\downarrow$ & \textbf{Sparseness} $\uparrow$ & \textbf{Infidelity} $\downarrow$ \\
        \midrule
        \textbf{FRInGe} & \best{\statentry{0.761}{0.009}} & \statentry{0.444}{0.020} & \statentry{0.709}{0.014} & \statentry{0.240}{0.012} & \statentry{0.723}{0.003} & \best{\statentry{0.020}{0.005}} \\
        AGI & \statentry{0.605}{0.011} & \statentry{0.520}{0.017} & \statentry{0.573}{0.014} & \statentry{0.321}{0.013} & \statentry{0.514}{0.006} & \tie{\statentry{0.021}{0.005}} \\
        GGIG & \statentry{0.631}{0.011} & \statentry{0.473}{0.015} & \statentry{0.658}{0.015} & \statentry{0.215}{0.011} & \statentry{0.471}{0.004} & \statentry{0.411}{0.018} \\
        GeoIG & \statentry{0.647}{0.009} & \statentry{0.469}{0.016} & \statentry{0.668}{0.014} & \statentry{0.216}{0.011} & \statentry{0.529}{0.002} & \statentry{4{,}760}{209} \\
        GuidedIG & \statentry{0.701}{0.010} & \statentry{0.452}{0.019} & \statentry{0.675}{0.013} & \statentry{0.242}{0.013} & \statentry{0.665}{0.003} & \tie{\statentry{0.027}{0.005}} \\
        IG & \statentry{0.673}{0.010} & \statentry{0.485}{0.019} & \statentry{0.647}{0.014} & \statentry{0.271}{0.013} & \statentry{0.624}{0.003} & \statentry{0.032}{0.006} \\
        IG$^2$ & \statentry{0.691}{0.013} & \statentry{0.455}{0.019} & \statentry{0.596}{0.013} & \statentry{0.289}{0.013} & \best{\statentry{0.739}{0.003}} & \best{\statentry{0.020}{0.005}} \\
        SG & \statentry{0.660}{0.009} & \best{\statentry{0.373}{0.008}} & \statentry{0.690}{0.013} & \statentry{0.206}{0.011} & \statentry{0.540}{0.003} & \tie{\statentry{0.021}{0.005}} \\
        BlurIG & \statentry{0.735}{0.009} & \tie{\statentry{0.386}{0.017}} & \best{\statentry{0.758}{0.013}} & \best{\statentry{0.174}{0.011}} & \statentry{0.655}{0.004} & \tie{\statentry{0.022}{0.006}} \\
        EG & \statentry{0.648}{0.011} & \statentry{0.481}{0.018} & \statentry{0.634}{0.015} & \statentry{0.267}{0.014} & \statentry{0.557}{0.002} & \tie{\statentry{0.022}{0.006}} \\
        MIG & \statentry{0.671}{0.012} & \statentry{0.467}{0.018} & \statentry{0.644}{0.016} & \statentry{0.281}{0.014} & \statentry{0.622}{0.004} & \tie{\statentry{0.021}{0.006}} \\
        \bottomrule
    \end{tabular*}
\end{table}

\begin{table}[H]
    \centering
    \small
    \setlength{\tabcolsep}{1pt}
    \renewcommand{\arraystretch}{1.15}
    \caption{Quantitative evaluation on ResNet-18. Means $\pm$ half-widths of 95\% image-bootstrap intervals; descriptive highlighting.}
    \label{tab:full_resnet18}
    \begin{tabular*}{\textwidth}{@{\extracolsep{\fill}}lcccccc@{}}
        \toprule
        & \multicolumn{2}{c}{\textit{Attribution calibration}}
        & \multicolumn{2}{c}{\textit{Causal faithfulness}}
        & \multicolumn{2}{c}{\textit{Focus and robustness}} \\
        \cmidrule(lr){2-3}\cmidrule(lr){4-5}\cmidrule(lr){6-7}
        \textbf{Method} & \textbf{MAS-Ins} $\uparrow$ & \textbf{MAS-Del} $\downarrow$ & \textbf{Ins-AUC} $\uparrow$ & \textbf{Del-AUC} $\downarrow$ & \textbf{Sparseness} $\uparrow$ & \textbf{Infidelity} $\downarrow$ \\
        \midrule
        \textbf{FRInGe} & \best{\statentry{0.597}{0.015}} & \best{\statentry{0.264}{0.010}} & \statentry{0.492}{0.016} & \statentry{0.130}{0.008} & \statentry{0.706}{0.002} & \tie{\statentry{0.058}{0.008}} \\
        AGI & \statentry{0.387}{0.014} & \statentry{0.395}{0.010} & \statentry{0.347}{0.013} & \statentry{0.212}{0.010} & \statentry{0.504}{0.007} & \best{\statentry{0.057}{0.008}} \\
        GGIG & \statentry{0.426}{0.016} & \statentry{0.373}{0.008} & \statentry{0.404}{0.016} & \statentry{0.164}{0.009} & \statentry{0.450}{0.002} & \statentry{0.432}{0.017} \\
        GeoIG & \statentry{0.453}{0.014} & \statentry{0.359}{0.007} & \statentry{0.430}{0.014} & \statentry{0.135}{0.008} & \statentry{0.497}{0.002} & \statentry{3{,}810}{132} \\
        GuidedIG & \statentry{0.477}{0.016} & \statentry{0.320}{0.010} & \statentry{0.448}{0.014} & \statentry{0.147}{0.008} & \statentry{0.649}{0.002} & \statentry{0.077}{0.008} \\
        IG & \statentry{0.443}{0.015} & \statentry{0.359}{0.011} & \statentry{0.415}{0.014} & \statentry{0.178}{0.010} & \statentry{0.608}{0.003} & \statentry{0.104}{0.009} \\
        IG$^2$ & \statentry{0.421}{0.016} & \statentry{0.314}{0.011} & \statentry{0.369}{0.013} & \statentry{0.179}{0.010} & \best{\statentry{0.714}{0.002}} & \best{\statentry{0.057}{0.008}} \\
        SG & \statentry{0.488}{0.014} & \statentry{0.325}{0.004} & \statentry{0.477}{0.015} & \statentry{0.122}{0.007} & \statentry{0.513}{0.002} & \best{\statentry{0.057}{0.008}} \\
        BlurIG & \tie{\statentry{0.577}{0.014}} & \tie{\statentry{0.276}{0.008}} & \best{\statentry{0.563}{0.015}} & \best{\statentry{0.096}{0.007}} & \statentry{0.634}{0.004} & \tie{\statentry{0.061}{0.009}} \\
        EG & \statentry{0.423}{0.016} & \statentry{0.368}{0.009} & \statentry{0.398}{0.015} & \statentry{0.165}{0.010} & \statentry{0.530}{0.002} & \tie{\statentry{0.063}{0.009}} \\
        MIG & \statentry{0.440}{0.016} & \statentry{0.358}{0.011} & \statentry{0.405}{0.016} & \statentry{0.197}{0.011} & \statentry{0.590}{0.004} & \tie{\statentry{0.058}{0.009}} \\
        \bottomrule
    \end{tabular*}
\end{table}

\begin{table}[H]
    \centering
    \small
    \setlength{\tabcolsep}{1pt}
    \renewcommand{\arraystretch}{1.15}
    \caption{Quantitative evaluation on ResNet-50. Means $\pm$ half-widths of 95\% image-bootstrap intervals; descriptive highlighting.}
    \label{tab:full_resnet50}
    \begin{tabular*}{\textwidth}{@{\extracolsep{\fill}}lcccccc@{}}
        \toprule
        & \multicolumn{2}{c}{\textit{Attribution calibration}}
        & \multicolumn{2}{c}{\textit{Causal faithfulness}}
        & \multicolumn{2}{c}{\textit{Focus and robustness}} \\
        \cmidrule(lr){2-3}\cmidrule(lr){4-5}\cmidrule(lr){6-7}
        \textbf{Method} & \textbf{MAS-Ins} $\uparrow$ & \textbf{MAS-Del} $\downarrow$ & \textbf{Ins-AUC} $\uparrow$ & \textbf{Del-AUC} $\downarrow$ & \textbf{Sparseness} $\uparrow$ & \textbf{Infidelity} $\downarrow$ \\
        \midrule
        \textbf{FRInGe} & \best{\statentry{0.648}{0.014}} & \best{\statentry{0.302}{0.012}} & \statentry{0.581}{0.016} & \statentry{0.161}{0.009} & \statentry{0.702}{0.002} & \tie{\statentry{0.150}{0.050}} \\
        AGI & \statentry{0.463}{0.015} & \statentry{0.424}{0.011} & \statentry{0.423}{0.014} & \statentry{0.257}{0.012} & \statentry{0.490}{0.006} & \best{\statentry{0.147}{0.048}} \\
        GGIG & \statentry{0.491}{0.015} & \statentry{0.410}{0.011} & \statentry{0.477}{0.016} & \statentry{0.213}{0.010} & \statentry{0.463}{0.003} & \statentry{2.004}{0.101} \\
        GeoIG & \statentry{0.523}{0.014} & \statentry{0.379}{0.009} & \statentry{0.527}{0.016} & \statentry{0.159}{0.009} & \statentry{0.507}{0.002} & \statentry{5{,}430}{198} \\
        GuidedIG & \statentry{0.551}{0.016} & \statentry{0.347}{0.012} & \statentry{0.527}{0.015} & \statentry{0.183}{0.010} & \statentry{0.656}{0.003} & \tie{\statentry{0.175}{0.047}} \\
        IG & \statentry{0.516}{0.015} & \statentry{0.388}{0.013} & \statentry{0.481}{0.014} & \statentry{0.226}{0.012} & \statentry{0.621}{0.003} & \tie{\statentry{0.232}{0.054}} \\
        IG$^2$ & \statentry{0.492}{0.017} & \statentry{0.355}{0.013} & \statentry{0.424}{0.014} & \statentry{0.233}{0.012} & \best{\statentry{0.716}{0.002}} & \tie{\statentry{0.148}{0.050}} \\
        SG & \statentry{0.562}{0.013} & \statentry{0.323}{0.005} & \statentry{0.582}{0.016} & \tie{\statentry{0.144}{0.008}} & \statentry{0.539}{0.003} & \best{\statentry{0.147}{0.048}} \\
        BlurIG & \statentry{0.619}{0.014} & \tie{\statentry{0.303}{0.010}} & \best{\statentry{0.621}{0.015}} & \best{\statentry{0.128}{0.008}} & \statentry{0.639}{0.004} & \tie{\statentry{0.163}{0.055}} \\
        EG & \statentry{0.507}{0.015} & \statentry{0.390}{0.012} & \statentry{0.482}{0.016} & \statentry{0.205}{0.012} & \statentry{0.542}{0.002} & \tie{\statentry{0.165}{0.055}} \\
        MIG & \statentry{0.527}{0.017} & \statentry{0.380}{0.013} & \statentry{0.491}{0.016} & \statentry{0.238}{0.012} & \statentry{0.609}{0.004} & \tie{\statentry{0.159}{0.056}} \\
        \bottomrule
    \end{tabular*}
\end{table}

\begin{table}[H]
    \centering
    \small
    \setlength{\tabcolsep}{1pt}
    \renewcommand{\arraystretch}{1.15}
    \caption{Quantitative evaluation on ResNet-101. Means $\pm$ half-widths of 95\% image-bootstrap intervals; descriptive highlighting.}
    \label{tab:full_resnet101}
    \begin{tabular*}{\textwidth}{@{\extracolsep{\fill}}lcccccc@{}}
        \toprule
        & \multicolumn{2}{c}{\textit{Attribution calibration}}
        & \multicolumn{2}{c}{\textit{Causal faithfulness}}
        & \multicolumn{2}{c}{\textit{Focus and robustness}} \\
        \cmidrule(lr){2-3}\cmidrule(lr){4-5}\cmidrule(lr){6-7}
        \textbf{Method} & \textbf{MAS-Ins} $\uparrow$ & \textbf{MAS-Del} $\downarrow$ & \textbf{Ins-AUC} $\uparrow$ & \textbf{Del-AUC} $\downarrow$ & \textbf{Sparseness} $\uparrow$ & \textbf{Infidelity} $\downarrow$ \\
        \midrule
        \textbf{FRInGe} & \best{\statentry{0.679}{0.014}} & \tie{\statentry{0.351}{0.015}} & \statentry{0.641}{0.017} & \statentry{0.228}{0.012} & \statentry{0.706}{0.002} & \tie{\statentry{0.121}{0.022}} \\
        AGI & \statentry{0.520}{0.014} & \statentry{0.475}{0.014} & \statentry{0.511}{0.015} & \statentry{0.315}{0.014} & \statentry{0.492}{0.006} & \best{\statentry{0.119}{0.021}} \\
        GGIG & \statentry{0.555}{0.014} & \statentry{0.437}{0.013} & \statentry{0.572}{0.017} & \statentry{0.241}{0.011} & \statentry{0.482}{0.003} & \statentry{1.681}{0.084} \\
        GeoIG & \statentry{0.567}{0.013} & \statentry{0.414}{0.012} & \statentry{0.607}{0.016} & \statentry{0.199}{0.010} & \statentry{0.506}{0.002} & \statentry{6{,}030}{207} \\
        GuidedIG & \statentry{0.605}{0.014} & \statentry{0.397}{0.015} & \statentry{0.602}{0.016} & \statentry{0.243}{0.013} & \statentry{0.660}{0.003} & \tie{\statentry{0.149}{0.021}} \\
        IG & \statentry{0.577}{0.014} & \statentry{0.433}{0.015} & \statentry{0.570}{0.015} & \statentry{0.278}{0.013} & \statentry{0.620}{0.003} & \statentry{0.189}{0.022} \\
        IG$^2$ & \statentry{0.569}{0.016} & \statentry{0.416}{0.016} & \statentry{0.505}{0.015} & \statentry{0.300}{0.014} & \best{\statentry{0.722}{0.002}} & \tie{\statentry{0.120}{0.022}} \\
        SG & \statentry{0.605}{0.012} & \tie{\statentry{0.341}{0.006}} & \statentry{0.646}{0.016} & \statentry{0.187}{0.010} & \statentry{0.537}{0.003} & \best{\statentry{0.119}{0.021}} \\
        BlurIG & \tie{\statentry{0.662}{0.013}} & \best{\statentry{0.338}{0.014}} & \best{\statentry{0.694}{0.016}} & \best{\statentry{0.166}{0.010}} & \statentry{0.635}{0.004} & \tie{\statentry{0.126}{0.024}} \\
        EG & \statentry{0.562}{0.015} & \statentry{0.439}{0.015} & \statentry{0.572}{0.018} & \statentry{0.261}{0.014} & \statentry{0.545}{0.002} & \tie{\statentry{0.129}{0.024}} \\
        MIG & \statentry{0.591}{0.016} & \statentry{0.425}{0.015} & \statentry{0.578}{0.018} & \statentry{0.288}{0.014} & \statentry{0.613}{0.004} & \tie{\statentry{0.124}{0.024}} \\
        \bottomrule
    \end{tabular*}
\end{table}

\begin{table}[H]
    \centering
    \small
    \setlength{\tabcolsep}{1pt}
    \renewcommand{\arraystretch}{1.15}
    \caption{Quantitative evaluation on ResNet-152. Means $\pm$ half-widths of 95\% image-bootstrap intervals; descriptive highlighting.}
    \label{tab:full_resnet152}
    \begin{tabular*}{\textwidth}{@{\extracolsep{\fill}}lcccccc@{}}
        \toprule
        & \multicolumn{2}{c}{\textit{Attribution calibration}}
        & \multicolumn{2}{c}{\textit{Causal faithfulness}}
        & \multicolumn{2}{c}{\textit{Focus and robustness}} \\
        \cmidrule(lr){2-3}\cmidrule(lr){4-5}\cmidrule(lr){6-7}
        \textbf{Method} & \textbf{MAS-Ins} $\uparrow$ & \textbf{MAS-Del} $\downarrow$ & \textbf{Ins-AUC} $\uparrow$ & \textbf{Del-AUC} $\downarrow$ & \textbf{Sparseness} $\uparrow$ & \textbf{Infidelity} $\downarrow$ \\
        \midrule
        \textbf{FRInGe} & \best{\statentry{0.691}{0.013}} & \statentry{0.392}{0.017} & \tie{\statentry{0.681}{0.016}} & \statentry{0.265}{0.013} & \statentry{0.703}{0.002} & \tie{\statentry{0.111}{0.020}} \\
        AGI & \statentry{0.537}{0.014} & \statentry{0.500}{0.015} & \statentry{0.538}{0.016} & \statentry{0.358}{0.015} & \statentry{0.496}{0.006} & \tie{\statentry{0.111}{0.019}} \\
        GGIG & \statentry{0.563}{0.015} & \statentry{0.469}{0.015} & \statentry{0.592}{0.017} & \statentry{0.301}{0.013} & \statentry{0.490}{0.003} & \statentry{2.539}{0.134} \\
        GeoIG & \statentry{0.588}{0.012} & \statentry{0.432}{0.014} & \statentry{0.651}{0.016} & \tie{\statentry{0.229}{0.012}} & \statentry{0.510}{0.002} & \statentry{6{,}290}{225} \\
        GuidedIG & \statentry{0.619}{0.015} & \statentry{0.419}{0.016} & \statentry{0.631}{0.015} & \statentry{0.275}{0.013} & \statentry{0.661}{0.003} & \tie{\statentry{0.142}{0.020}} \\
        IG & \statentry{0.588}{0.015} & \statentry{0.463}{0.016} & \statentry{0.587}{0.015} & \statentry{0.324}{0.015} & \statentry{0.624}{0.003} & \statentry{0.192}{0.021} \\
        IG$^2$ & \statentry{0.590}{0.016} & \statentry{0.442}{0.018} & \statentry{0.539}{0.015} & \statentry{0.343}{0.014} & \best{\statentry{0.726}{0.002}} & \best{\statentry{0.110}{0.019}} \\
        SG & \statentry{0.620}{0.012} & \best{\statentry{0.349}{0.007}} & \tie{\statentry{0.682}{0.015}} & \tie{\statentry{0.215}{0.011}} & \statentry{0.546}{0.003} & \tie{\statentry{0.111}{0.019}} \\
        BlurIG & \tie{\statentry{0.666}{0.012}} & \statentry{0.379}{0.016} & \best{\statentry{0.704}{0.015}} & \best{\statentry{0.206}{0.012}} & \statentry{0.639}{0.004} & \tie{\statentry{0.120}{0.023}} \\
        EG & \statentry{0.575}{0.014} & \statentry{0.458}{0.016} & \statentry{0.596}{0.018} & \statentry{0.296}{0.015} & \statentry{0.546}{0.002} & \tie{\statentry{0.121}{0.023}} \\
        MIG & \statentry{0.608}{0.015} & \statentry{0.456}{0.016} & \statentry{0.609}{0.018} & \statentry{0.333}{0.015} & \statentry{0.614}{0.004} & \tie{\statentry{0.116}{0.023}} \\
        \bottomrule
    \end{tabular*}
\end{table}

\begin{table}[H]
    \centering
    \small
    \setlength{\tabcolsep}{1pt}
    \renewcommand{\arraystretch}{1.15}
    \caption{Quantitative evaluation on VGG-19. Means $\pm$ half-widths of 95\% image-bootstrap intervals; descriptive highlighting.}
    \label{tab:full_vgg19}
    \begin{tabular*}{\textwidth}{@{\extracolsep{\fill}}lcccccc@{}}
        \toprule
        & \multicolumn{2}{c}{\textit{Attribution calibration}}
        & \multicolumn{2}{c}{\textit{Causal faithfulness}}
        & \multicolumn{2}{c}{\textit{Focus and robustness}} \\
        \cmidrule(lr){2-3}\cmidrule(lr){4-5}\cmidrule(lr){6-7}
        \textbf{Method} & \textbf{MAS-Ins} $\uparrow$ & \textbf{MAS-Del} $\downarrow$ & \textbf{Ins-AUC} $\uparrow$ & \textbf{Del-AUC} $\downarrow$ & \textbf{Sparseness} $\uparrow$ & \textbf{Infidelity} $\downarrow$ \\
        \midrule
        \textbf{FRInGe} & \tie{\statentry{0.588}{0.015}} & \best{\statentry{0.255}{0.008}} & \statentry{0.503}{0.016} & \statentry{0.119}{0.007} & \statentry{0.707}{0.003} & \tie{\statentry{0.343}{0.064}} \\
        AGI & \statentry{0.440}{0.014} & \statentry{0.405}{0.009} & \statentry{0.448}{0.016} & \statentry{0.192}{0.010} & \statentry{0.433}{0.006} & \tie{\statentry{0.346}{0.064}} \\
        GGIG & \statentry{0.460}{0.015} & \statentry{0.357}{0.008} & \statentry{0.441}{0.016} & \statentry{0.159}{0.008} & \statentry{0.497}{0.003} & \statentry{0.862}{0.065} \\
        GeoIG & \statentry{0.487}{0.014} & \statentry{0.335}{0.006} & \statentry{0.490}{0.015} & \statentry{0.118}{0.007} & \statentry{0.529}{0.002} & \statentry{4{,}270}{186} \\
        GuidedIG & \statentry{0.496}{0.015} & \statentry{0.305}{0.009} & \statentry{0.468}{0.015} & \statentry{0.148}{0.008} & \statentry{0.674}{0.003} & \tie{\statentry{0.387}{0.063}} \\
        IG & \statentry{0.479}{0.015} & \statentry{0.337}{0.010} & \statentry{0.450}{0.014} & \statentry{0.170}{0.009} & \statentry{0.637}{0.003} & \tie{\statentry{0.419}{0.066}} \\
        IG$^2$ & \statentry{0.448}{0.017} & \statentry{0.290}{0.010} & \statentry{0.419}{0.015} & \statentry{0.159}{0.009} & \best{\statentry{0.765}{0.003}} & \best{\statentry{0.342}{0.064}} \\
        SG & \statentry{0.523}{0.014} & \statentry{0.304}{0.004} & \statentry{0.522}{0.016} & \statentry{0.113}{0.007} & \statentry{0.547}{0.003} & \best{\statentry{0.342}{0.064}} \\
        BlurIG & \best{\statentry{0.596}{0.014}} & \tie{\statentry{0.266}{0.008}} & \best{\statentry{0.567}{0.015}} & \best{\statentry{0.098}{0.007}} & \statentry{0.649}{0.004} & \tie{\statentry{0.361}{0.073}} \\
        EG & \statentry{0.439}{0.016} & \statentry{0.357}{0.009} & \statentry{0.423}{0.016} & \statentry{0.165}{0.010} & \statentry{0.560}{0.003} & \tie{\statentry{0.367}{0.072}} \\
        MIG & \statentry{0.475}{0.016} & \statentry{0.345}{0.010} & \statentry{0.433}{0.016} & \statentry{0.190}{0.010} & \statentry{0.620}{0.004} & \tie{\statentry{0.356}{0.072}} \\
        \bottomrule
    \end{tabular*}
\end{table}

\subsection{Robustness under small input perturbations}
\label{app:max_sens_appendix}
We additionally report Max Sensitivity as a direct robustness diagnostic under small input perturbations, evaluated on a sample of 64 images for each architecture. Because this metric is defined through a maximum over sampled perturbations and can therefore be skewed by a small number of difficult examples, we summarize it here using the median together with the interquartile half-width, i.e., \((Q_3-Q_1)/2\). This yields a robust summary of central tendency and dispersion that is less sensitive to outliers than the mean.

Table~\ref{tab:max_sensitivity_iqr} shows lower median sensitivity for
FRInGe on ResNet-18 and VGG-19, and for IG on the other four architectures.
FRInGe's half-IQR is smaller on five architectures, but a narrower spread
does not offset a higher median or establish greater robustness.
The results therefore do not support a general local-stability advantage
for FRInGe. This diagnostic is complementary to MAS and AUC: an
attribution can align well with a model's perturbation response without
being less sensitive to small input changes.

\begin{table*}[h]
    \centering
    \small
    \setlength{\tabcolsep}{0pt}
    \renewcommand{\arraystretch}{1.15}
    \caption{Max Sensitivity ($\downarrow$) across architectures for FRInGe and standard IG, evaluated on 64 images per model. Values are reported as $\mathrm{median} \pm \mathrm{half\text{-}IQR}$, where $\mathrm{half\text{-}IQR}=(Q_3-Q_1)/2$. The best median in each column is shown in \best{bold underline}; methods whose $\mathrm{median} \pm \mathrm{half\text{-}IQR}$ intervals overlap with the best are shown in \tie{bold} as a descriptive near-tie heuristic.}
    \label{tab:max_sensitivity_iqr}
    \begin{tabular*}{\textwidth}{@{\extracolsep{\fill}}lcccccc@{}}
        \toprule
        \textbf{Method} & \textbf{ResNet-18} & \textbf{ResNet-50} & \textbf{ResNet-101} & \textbf{ResNet-152} & \textbf{Inception-v3} & \textbf{VGG-19} \\
        \midrule
        FRInGe (Ours) & \best{\statentry{0.0758}{.0122}} & \tie{\statentry{0.1139}{.0141}} & \statentry{0.1225}{.0138} & \statentry{0.1383}{.0170} & \statentry{0.1081}{.0131} & \best{\statentry{0.0665}{.0094}} \\
        Standard IG   & \tie{\statentry{0.0833}{.0171}} & \best{\statentry{0.0917}{.0313}} & \best{\statentry{0.0829}{.0209}} & \best{\statentry{0.0900}{.0313}} & \best{\statentry{0.0381}{.0093}} & \tie{\statentry{0.0678}{.0126}} \\
        \bottomrule
    \end{tabular*}
\end{table*}

\FloatBarrier
\section{Additional Qualitative Results}
\label{app:qualitative}

For each architecture, images are ranked by FRInGe's average signed
improvement over the seven displayed baselines, with weights
$(1,1,0.1,0.1)$ for MAS-Insertion, MAS-Deletion, insertion AUC, and
deletion AUC. Differences are oriented so that positive values favor
FRInGe. We show two highest-ranked images, two near the median, and
two lowest-ranked images; these are stratified examples, not a random
sample.

Maps show absolute attributions summed over channels, clipped at the
99th percentile, and independently normalized to $[0,1]$.
MI/MD denote MAS-Insertion/Deletion and I/D normalized insertion/deletion
AUC; \emph{Ours} denotes FRInGe and GIG denotes GeoIG.

\begin{figure*}[h]
    \centering
    \includegraphics[width=\textwidth]{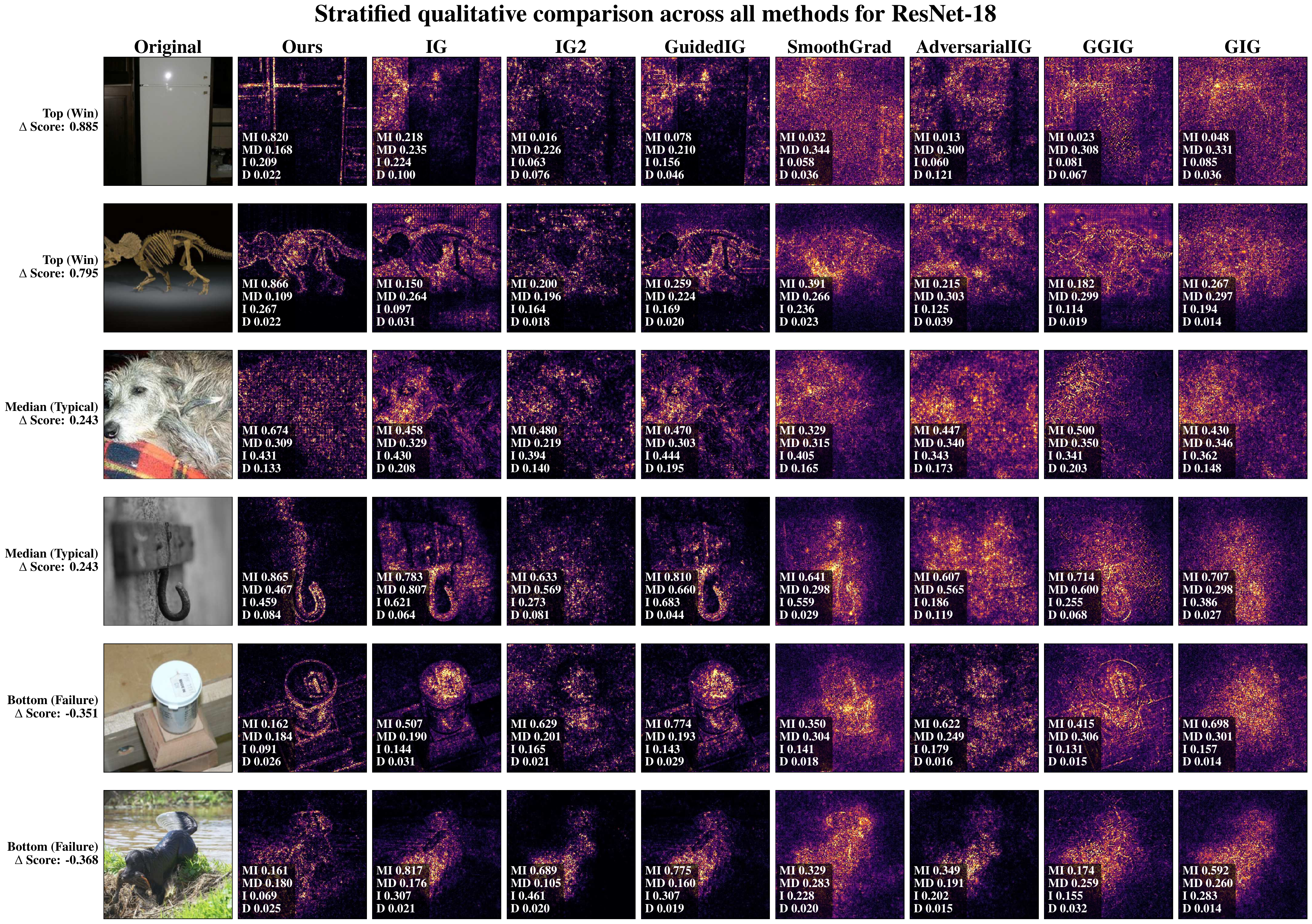}
    \caption{ResNet-18 magnitude maps for top-, median-, and bottom-ranked examples under the selection rule in Appendix~\ref{app:qualitative}.}
    \label{fig:qualitative_resnet18}
\end{figure*}

\begin{figure*}[h]
    \centering
    \includegraphics[width=\textwidth]{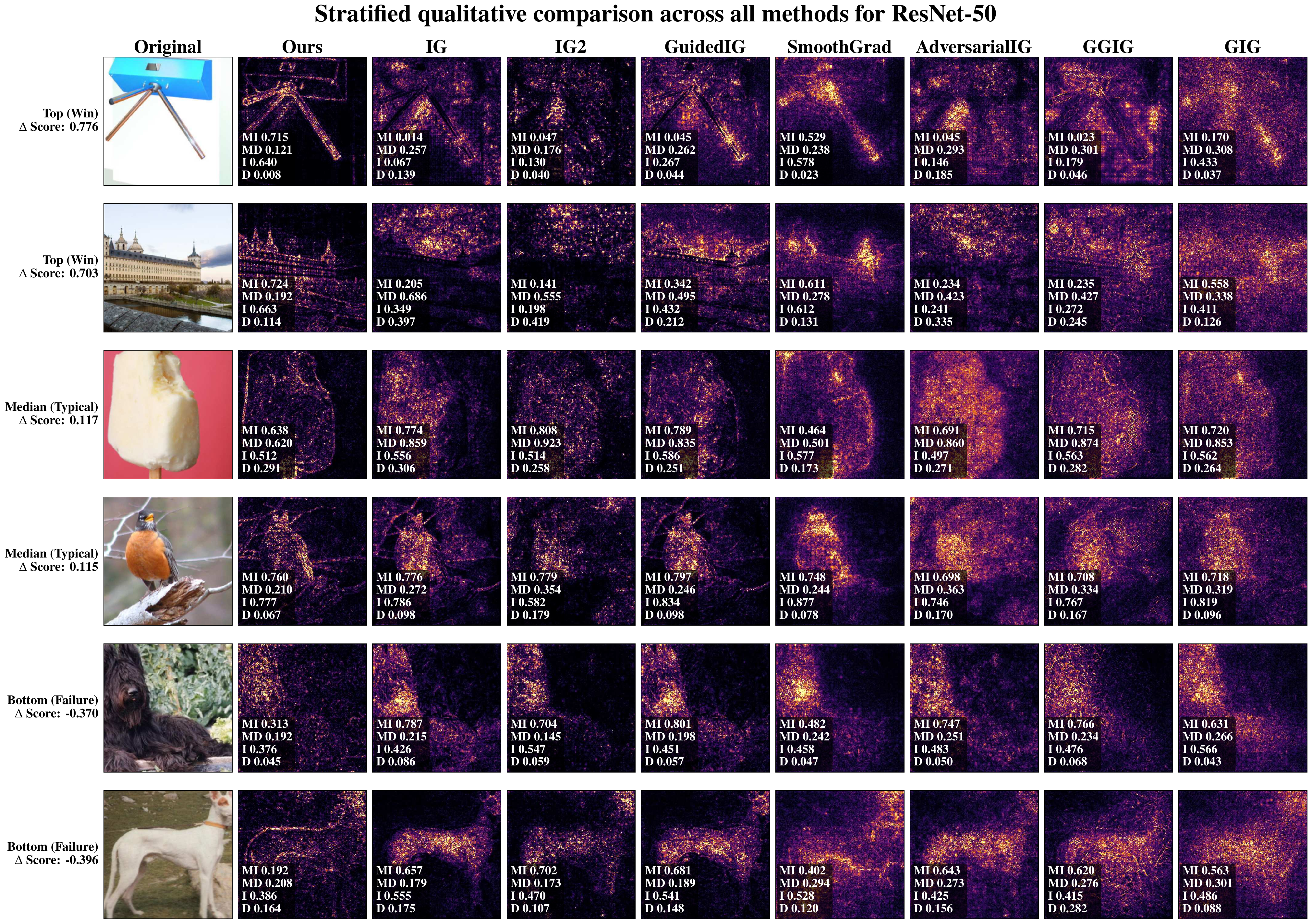}
    \caption{ResNet-50 magnitude maps under the same stratified selection and display conventions.}
    \label{fig:qualitative_resnet50}
\end{figure*}

\begin{figure*}[h]
    \centering
    \includegraphics[width=\textwidth]{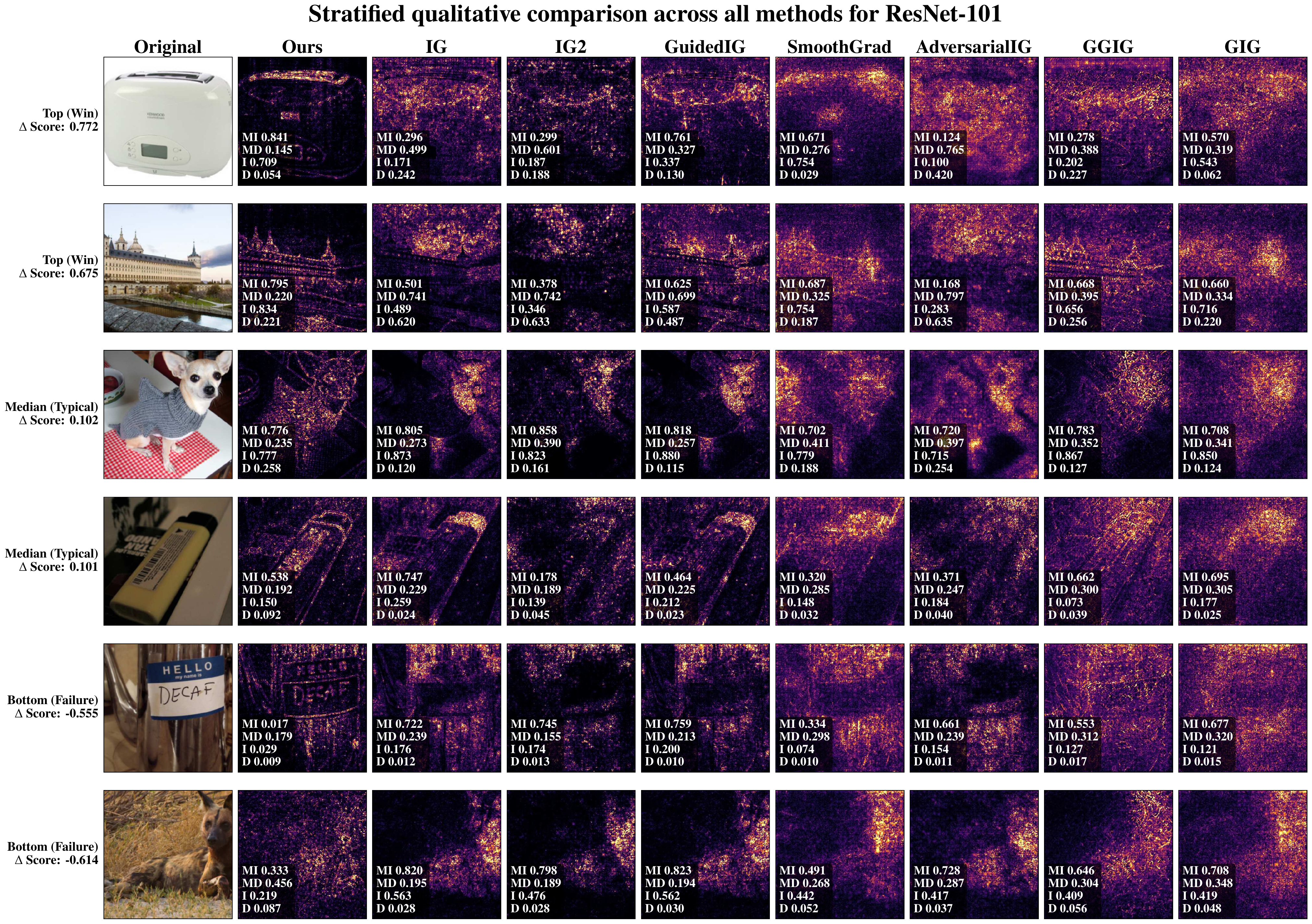}
    \caption{ResNet-101 magnitude maps under the same stratified selection and display conventions.}
    \label{fig:qualitative_resnet101}
\end{figure*}

\begin{figure*}[h]
    \centering
    \includegraphics[width=\textwidth]{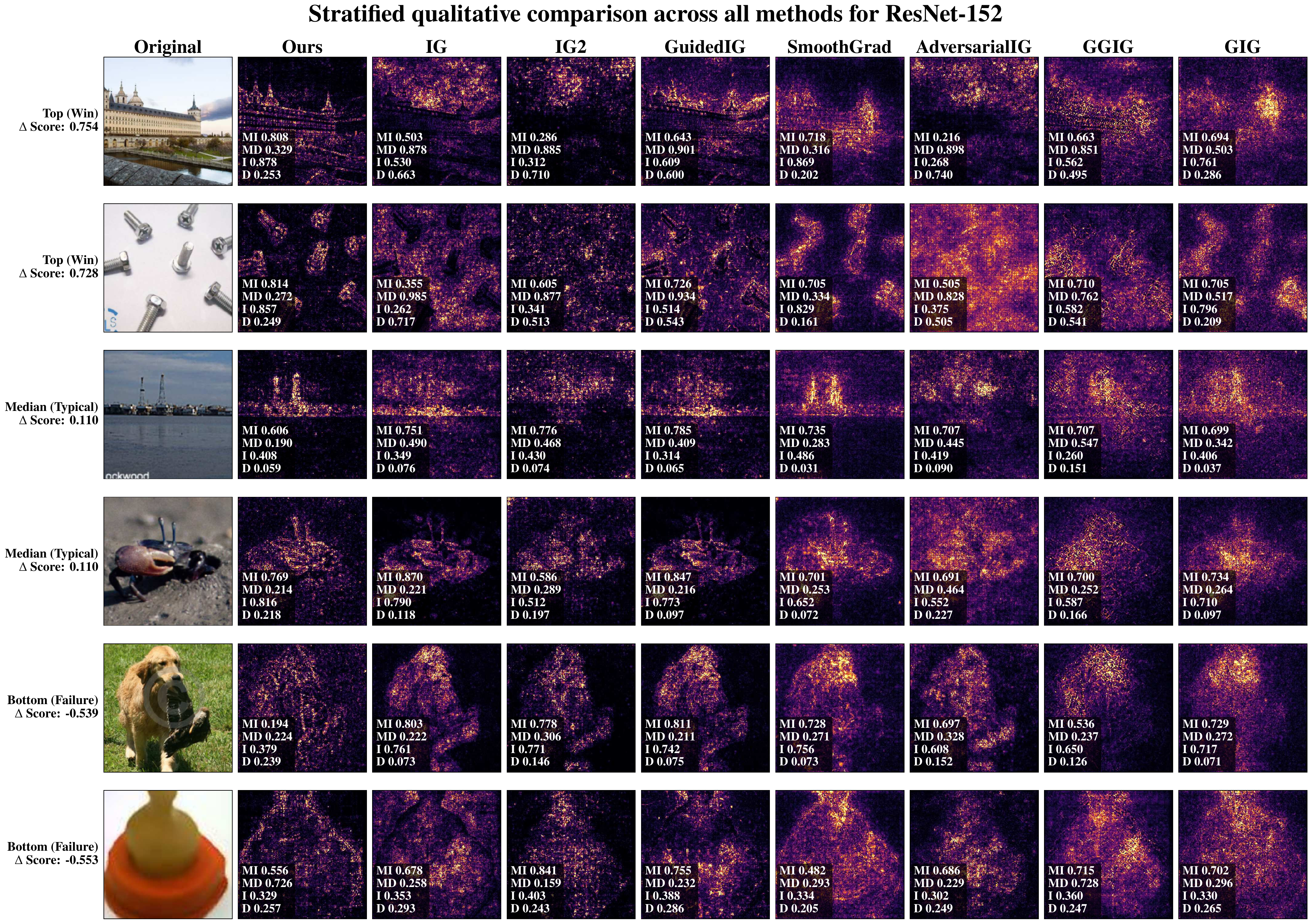}
    \caption{ResNet-152 magnitude maps under the same stratified selection and display conventions.}
    \label{fig:qualitative_resnet152}
\end{figure*}

\begin{figure*}[h]
    \centering
    \includegraphics[width=\textwidth]{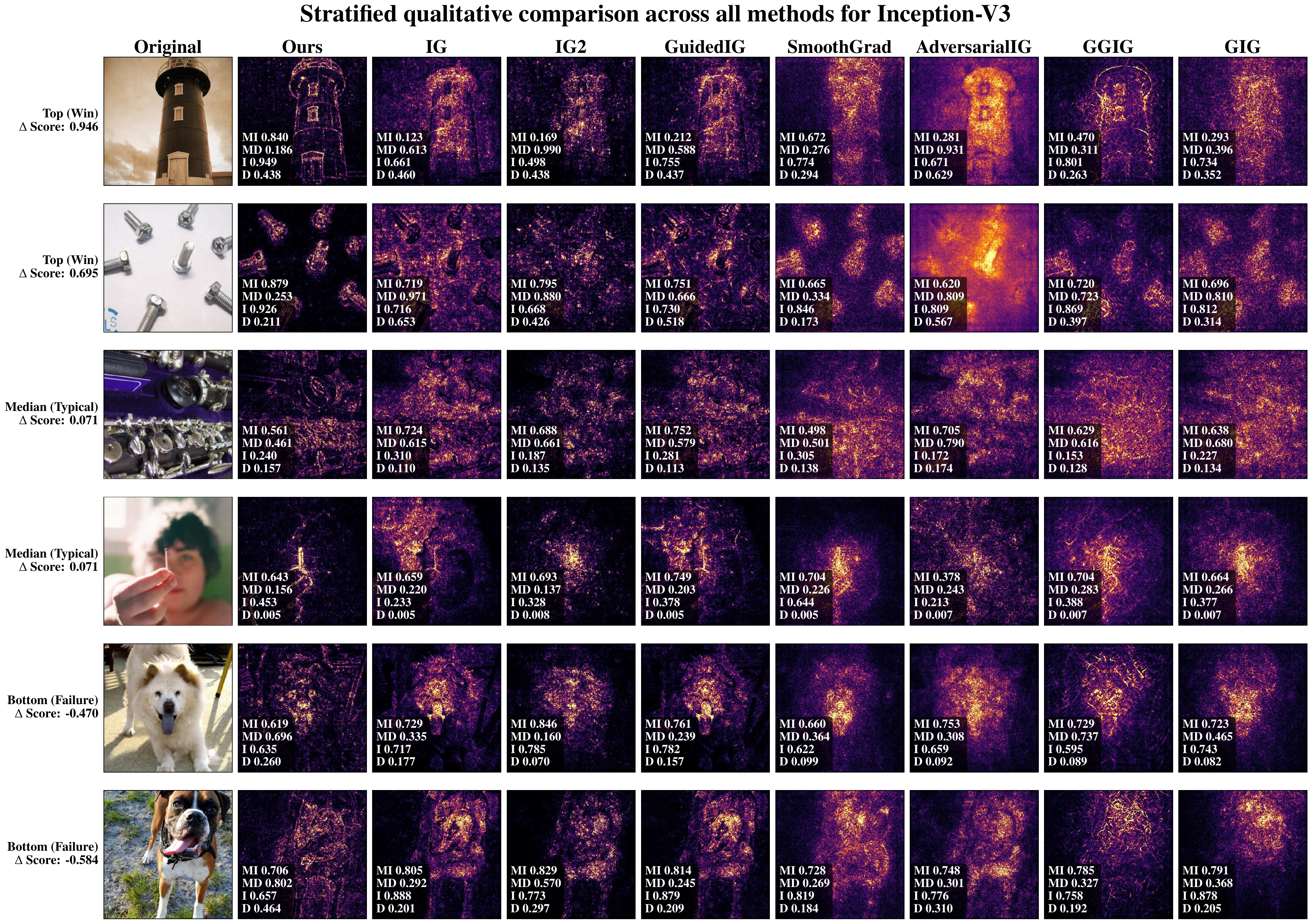}
    \caption{Qualitative comparison across the displayed attribution methods on Inception-v3 under the same stratified selection and display conventions.}
    \label{fig:qualitative_inception_v3}
\end{figure*}
\begin{figure*}[h]
    \centering
    \includegraphics[width=\textwidth]{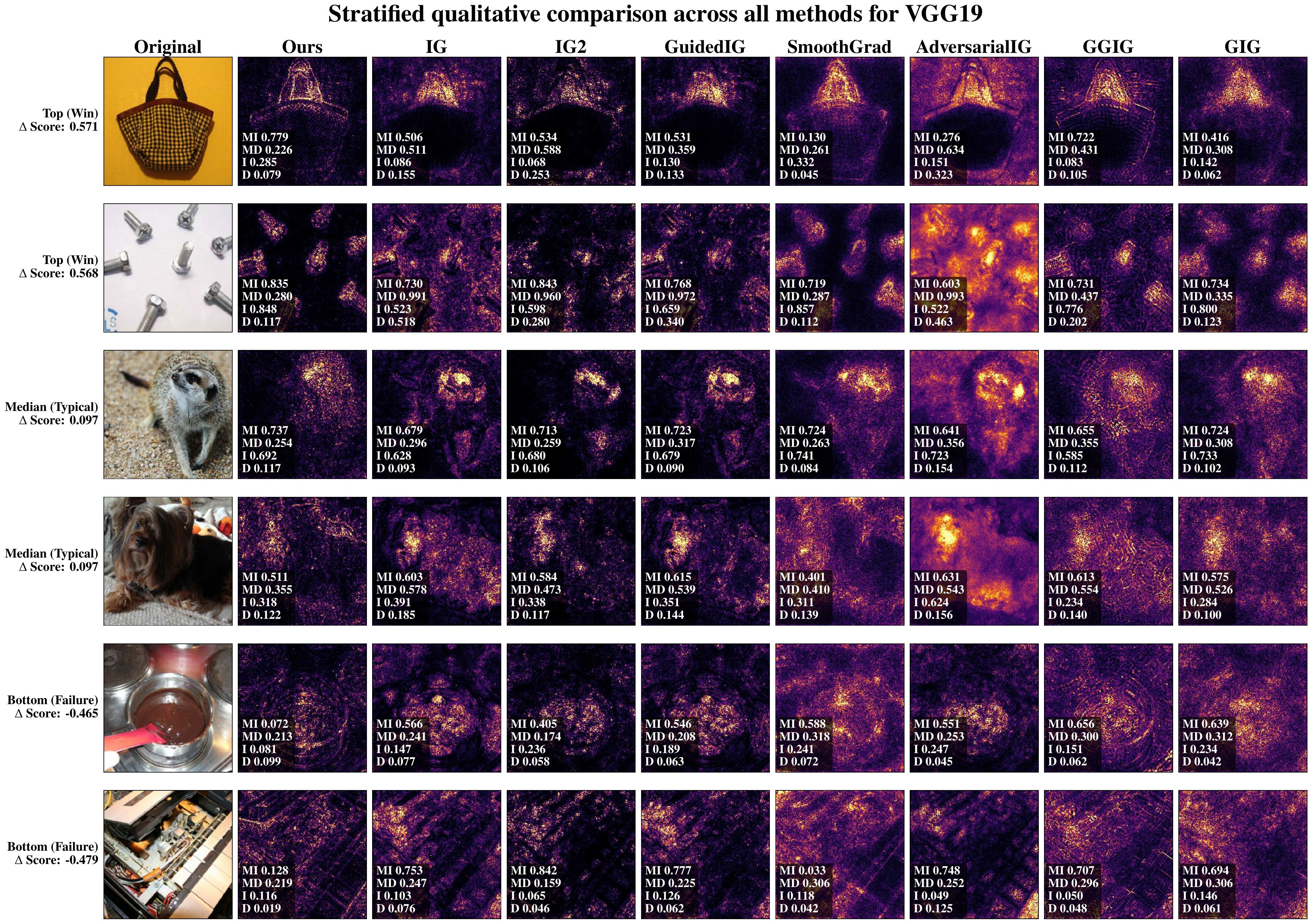}
    \caption{VGG-19 magnitude maps under the same stratified selection and display conventions.}
    \label{fig:qualitative_vgg19}
\end{figure*}

\FloatBarrier
\section{Target-vs-Rest Fisher--Rao Specialization (FRInGe-B)}
\label{app:fringeb}

FRInGe-B changes the predictive object before solving the geometric update. It
coarse-grains the categorical prediction into the probability of the explained
class versus the probability of all remaining classes. Within this Bernoulli
model, its Fisher--Rao geometry and rank-one pullback are exact. Relative to
categorical FRInGe, the deliberate modeling simplification is the removal of variation
\emph{within} the non-target group. FRInGe-B is therefore a
lower-cost specialization, not a numerical approximation that converges to the
categorical formulation as its solver tolerance is tightened.

We first define the chance-preserving binary reference and its
Fisher--Rao waypoints, then derive the rank-one direction and the spatially
regularized correction. A metric decomposition identifies exactly which
categorical information is discarded, and the algorithm combines these
ingredients with the same attribution quadrature as categorical FRInGe.
The empirical part separates a matched 64-image diagnostic pilot,
independently tuned full-cohort comparisons, and shared-setting
qualitative examples. We conclude with the endpoint and modality
limitations needed to interpret the specialization. Exactness below
concerns the binary predictive model and algebraic identities; finite
linear solves, input-space tracking, and quadrature remain approximate.

\begin{table}[H]
\centering
\small
\setlength{\tabcolsep}{5pt}
\caption{FRInGe-B at a glance. \emph{Exact} refers to the coarse-grained
Bernoulli model, not to the original $C$-class geometry.}
\label{tab:fringeb_summary}
\begin{tabularx}{\linewidth}{@{}p{0.27\linewidth}X@{}}
\toprule
Predictive state & Target probability $p_t(x)$ versus rest probability $1-p_t(x)$ \\
Reference & $(1/C,\,1-1/C)$, the image of the categorical uniform distribution \\
Pullback geometry & Exact Bernoulli Fisher metric; rank at most one \\
Classifier-dependent solve & One target-vs-rest log-odds gradient; no Fisher JVP/VJP products \\
Information retained & Changes that move probability mass between class $t$ and its complement \\
Information discarded & Redistribution of mass among the $C-1$ non-target classes \\
\bottomrule
\end{tabularx}
\end{table}

\subsection{Exact Binary Predictive Geometry}

FRInGe-B applies the same predictive-space construction as FRInGe
after grouping the classes into the explained class $t$ and all remaining
classes. With $p_t(x)=\softmax(F(x))_t$, the resulting binary prediction is
\begin{equation}
\widetilde p_t(x)
=
\begin{bmatrix}
p_t(x)\\[1mm]
1-p_t(x)
\end{bmatrix},
\label{eq:fringeb_collapse}
\end{equation}
where the second entry is the total probability assigned to non-target
classes. Applying this grouping to the uniform categorical reference gives
\begin{equation}
\widetilde q_t^\star
=
\begin{bmatrix}
1/C\\[1mm]
1-1/C
\end{bmatrix}.
\label{eq:fringeb_reference}
\end{equation}
This reference preserves the original chance probability $1/C$ for
class $t$. For $C>2$, it therefore differs from the unconstrained binary
maximum-entropy distribution $(1/2,1/2)$. When $C=2$, the grouping preserves
the full predictive distribution.

Starting from $x_0=x$, we construct the Fisher--Rao path from
$\widetilde p_t(x_0)$ to $\widetilde q_t^\star$. Let
$\widetilde s(\alpha)$ denote its square-root representation, obtained
using the spherical interpolation in Eq.~\eqref{eq:slerp_simple}.
As in FRInGe, we discretize this path into waypoints:
\[
\widetilde s_k
=
\widetilde s(\alpha_k)
=
\begin{bmatrix}
\widetilde s_{k,t}\\[1mm]
\widetilde s_{k,\mathrm{rest}}
\end{bmatrix},
\qquad
\alpha_k=\frac{k}{T_{\mathrm B}},
\qquad k=0,\dots,T_{\mathrm B},
\]
where $T_{\mathrm B}$ is the number of binary tracking steps. The index $k$
identifies the waypoint, while $t$ and $\mathrm{rest}$ identify its two
coordinates. Specifically, $\widetilde s_{k,t}^{\,2}$ is the prescribed
target-class probability and $\widetilde s_{k,\mathrm{rest}}^{\,2}$ is
the prescribed probability of all other classes combined, with
$\widetilde s_{k,t}^{\,2}+\widetilde s_{k,\mathrm{rest}}^{\,2}=1$.
These waypoint probabilities need not equal those attained at the
realized input $x_k$. At state $x_k$, we target the next waypoint
$\widetilde s_{k+1}$, following the same indexing convention as FRInGe.

To derive the binary pullback metric, define the target-vs-rest log-odds
\begin{equation}
\delta_t(x)
=
F^t(x)-\log\!\left(\sum_{j\ne t}\exp F^j(x)\right),
\qquad
p_t(x)=\sigma(\delta_t(x)),
\label{eq:fringeb_logodds}
\end{equation}
where $\sigma$ is the logistic sigmoid, and let
$u_t(x)=\nabla_x\delta_t(x)\in\mathbb{R}^n$.
The Bernoulli Fisher information in log-odds coordinates is
$p_t(x)(1-p_t(x))$. Its exact pullback to input space is therefore
\begin{equation}
G_{\mathrm B}(x)
=
p_t(x)(1-p_t(x))\,u_t(x)u_t(x)^\top
\in\mathbb{R}^{n\times n}.
\label{eq:fringeb_metric}
\end{equation}
This metric has rank at most one, independently of the number of
original classes.

The spherical alignment loss toward the next waypoint is
\[
\begin{aligned}
\widetilde{\mathcal L}_{k+1}(x)
&=1-\bigl\langle
\sqrt{\widetilde p_t(x)},\widetilde s_{k+1}
\bigr\rangle\\
&=1-\sqrt{p_t(x)}\,\widetilde s_{k+1,t}
-\sqrt{1-p_t(x)}\,\widetilde s_{k+1,\mathrm{rest}}.
\end{aligned}
\]
The waypoint is held fixed when differentiating this loss. Since its
dependence on $x$ is entirely through $\delta_t(x)$, the gradient is
collinear with $u_t(x)$. Defining
\[
p_{t,k}=p_t(x_k),
\qquad
u_k=u_t(x_k),
\]
we obtain the same per-step gradient structure used by FRInGe:
\begin{equation}
\begin{aligned}
g_k
&=\nabla_x\widetilde{\mathcal L}_{k+1}(x)\rvert_{x=x_k}
=\beta_k u_k,\\
\beta_k
&=\frac12\bigl[
p_{t,k}\sqrt{1-p_{t,k}}\,
\widetilde s_{k+1,\mathrm{rest}}
-(1-p_{t,k})\sqrt{p_{t,k}}\,
\widetilde s_{k+1,t}
\bigr].
\end{aligned}
\label{eq:fringeb_beta}
\end{equation}
Thus, $\beta_k$ combines the current target probability $p_{t,k}$ with
the prescribed coordinates of the next waypoint $\widetilde s_{k+1}$.

\subsection{Rank-One Direction and Spatial Regularization}

Without spatial smoothing, the damped FRInGe-B system is
\[
\bigl(\lambda I+p_{t,k}(1-p_{t,k})u_ku_k^\top\bigr)v_k
=g_k=\beta_k u_k.
\]
Because the right-hand side lies in the span of $u_k$, its exact solution is
\begin{equation}
v_k
=\frac{\beta_k}{\lambda+p_{t,k}(1-p_{t,k})\|u_k\|_2^2}\,u_k.
\label{eq:fringeb_closed_form}
\end{equation}
The damping $\lambda>0$ keeps the denominator positive when the target
probability is saturated or $u_k$ is small. Equation~\eqref{eq:fringeb_closed_form}
requires one reverse-mode derivative of the target-vs-rest log-odds and no
pullback-Fisher products or Fisher conjugate-gradient solve. This is the main
source of the computational reduction.

The benchmarked image-model variant retains both spatial terms of
categorical FRInGe. Define
\[
H=\lambda I+\gamma_{\mathrm{step}}L^\top L,
\qquad
b_k=g_k+\gamma_{\mathrm{prior}}L^\top Lx_k.
\]
Here $H$ is fixed across waypoints, whereas $b_k$ depends on the current
state. The direction solves
$\bigl(H+p_{t,k}(1-p_{t,k})u_ku_k^\top\bigr)v_k=b_k$.
The Sherman--Morrison identity gives
\begin{equation}
v_k
=H^{-1}b_k
-\frac{p_{t,k}(1-p_{t,k})H^{-1}u_k\,u_k^\top H^{-1}b_k}
{1+p_{t,k}(1-p_{t,k})u_k^\top H^{-1}u_k}.
\label{eq:fringeb_sherman_morrison}
\end{equation}
This formula is exact when $H^{-1}$ is applied exactly. In the evaluated
spatially regularized variant, the two inverse applications are
approximated by iterative solves involving the fixed spatial operator.
Thus, only the unsmoothed direction is closed form; the smoothed variant
still has numerical solve error. These spatial iterations do not evaluate
the classifier, which is the computational distinction from the
categorical Fisher solve.

The Fisher norm used by the trust-region rule also becomes scalar:
\begin{equation}
\|v_k\|_{G_{\mathrm B}(x_k)}^2
=p_{t,k}(1-p_{t,k})(u_k^\top v_k)^2.
\label{eq:fringeb_norm}
\end{equation}
The KL and Euclidean caps, trapezoidal attribution quadrature, sign convention,
and realized-endpoint completeness relation are otherwise unchanged.

\subsection{Information Discarded by Aggregation}

The rank-one metric in Eq.~\eqref{eq:fringeb_metric} is exactly the
between-group component of the categorical pullback. The following
decomposition identifies the omitted information.

\begin{proposition}[Categorical metric decomposition]
\label{prop:fringeb_decomposition}
Let $h_j(x)=\nabla_x F^j(x)$ and define the conditional non-target weights and
mean gradient
\[
r_j(x)=\frac{p_j(x)}{1-p_t(x)},\qquad
\bar h_r(x)=\sum_{j\ne t}r_j(x)h_j(x).
\]
Then $u_t(x)=h_t(x)-\bar h_r(x)$ and
\begin{equation}
G(x)
=G_{\mathrm B}(x)
+\bigl(1-p_t(x)\bigr)\sum_{j\ne t}r_j(x)
\bigl(h_j(x)-\bar h_r(x)\bigr)
\bigl(h_j(x)-\bar h_r(x)\bigr)^\top.
\label{eq:fringeb_decomposition}
\end{equation}
Consequently, $G_{\mathrm B}(x)\preceq G(x)$ in the positive-semidefinite
order, with equality for binary classification or whenever the conditional
non-target logit gradients are identical.
\end{proposition}

\begin{proof}
The categorical pullback is the covariance of the logit gradients:
$G=\sum_jp_j(h_j-\bar h)(h_j-\bar h)^\top$, where
$\bar h=\sum_jp_jh_j$. Applying the law of total covariance to the partition
$\{t\}$ versus $\{j:j\ne t\}$ gives the between-group term
$p_t(1-p_t)(h_t-\bar h_r)(h_t-\bar h_r)^\top=G_{\mathrm B}$ and the
conditional within-rest covariance in Eq.~\eqref{eq:fringeb_decomposition}.
The latter is positive semidefinite.
\end{proof}

Thus, two categorical predictions with the same $p_t$ but different
allocations among the remaining classes are indistinguishable to FRInGe-B.
The specialization trades this within-rest information for a rank-one,
target-aware geometry. Removing that information changes the direction
and step control; the decomposition does not predict that either
specialization will yield better attributions. That is an empirical
question addressed by the comparisons below.

\subsection{FRInGe-B Algorithm}

Algorithm~\ref{alg:fringeb} makes the connection to the categorical
procedure explicit. The predictive state and Fisher action are replaced
by their binary forms, while the fixed explanation target, step
constraints, trapezoidal accumulation, and sign convention are retained.
The distinction between exact and iterative direction computations is
shown in the two branches.

\begin{algorithm}[H]
\caption{Target-vs-Rest FRInGe (FRInGe-B)}
\label{alg:fringeb}
\begin{algorithmic}[1]
\Require Input $x$, target $t$, logits $F$, class count $C$; $\lambda$, $\tau$,
$\eta_{\max}$, $\delta_{\mathrm{euc}}$; optional $L$,
$\gamma_{\mathrm{step}}$, $\gamma_{\mathrm{prior}}$
\Ensure Attributions $A$
\State $p_{t,0}\gets\softmax(F(x))_t$,
$\widetilde q_t^\star\gets(1/C,1-1/C)^\top$
\State Compute $T_{\mathrm B}$ from Eq.~\eqref{eq:step_sync} using
$d_{\mathrm{FR}}(\widetilde p_t(x),\widetilde q_t^\star)$
\State Build binary Fisher--Rao waypoints
$\{\widetilde s_j\}_{j=0}^{T_{\mathrm B}}$ from
$(\sqrt{p_{t,0}},\sqrt{1-p_{t,0}})^\top$ to
$\sqrt{\widetilde q_t^\star}$
\State $x_0\gets x$, $A_{\mathrm{path}}\gets0$
\For{$k=0,\dots,T_{\mathrm B}-1$}
    \State Compute $p_{t,k}$, $\delta_t(x_k)$, and
    $u_k\gets\nabla_x\delta_t(x_k)$
    \State Compute $\beta_k$ from Eq.~\eqref{eq:fringeb_beta} using
    $\widetilde s_{k+1}$
    \If{$\gamma_{\mathrm{step}}=\gamma_{\mathrm{prior}}=0$}
        \State $v_k\gets \beta_k u_k/
        \bigl(\lambda+p_{t,k}(1-p_{t,k})\|u_k\|_2^2\bigr)$
    \Else
        \State Compute $v_k$ from Eq.~\eqref{eq:fringeb_sherman_morrison}
    \EndIf
    \State $n_k^2\gets p_{t,k}(1-p_{t,k})(u_k^\top v_k)^2$
    \State $\eta_k\gets\min\!\left(\eta_{\max},
    \sqrt{2\tau/(n_k^2+\varepsilon)},
    \delta_{\mathrm{euc}}/(\|v_k\|_2+\varepsilon)\right)$
    \State $x_{k+1}\gets x_k-\eta_kv_k$
    \State $A_{\mathrm{path}}\gets A_{\mathrm{path}}+
    \frac12(\nabla F^t(x_k)+\nabla F^t(x_{k+1}))
    \odot(x_{k+1}-x_k)$
\EndFor
\State \Return $A=-A_{\mathrm{path}}$
\end{algorithmic}
\end{algorithm}

\subsection{Computational Cost}
\label{app:fringeb_cost}

\paragraph{Classifier and spatial work.}
Let $T_{\mathrm B}$ be the number of binary tracking steps and let $N$,
$T$, and $M$ have the meanings given in Appendix~\ref{app:complexity}.
Each binary step needs one log-odds gradient and one target-score gradient, but no
classifier-dependent Fisher JVP/VJP products. With the same leading-order
gradient accounting used for FRInGe, its classifier work is
\begin{align*}
\mathcal C_{\mathrm B}^{\mathrm{model}}
\approx 2T_{\mathrm B}(C_{\mathrm{fwd}}+C_{\mathrm{bwd}}),
\qquad
\frac{\mathcal C_{\mathrm B}^{\mathrm{model}}}
{\mathcal C_{\mathrm{IG}}^{\mathrm{model}}}
\approx\frac{2T_{\mathrm B}}{N},
\qquad
\frac{\mathcal C_{\mathrm B}^{\mathrm{model}}}
{\mathcal C_{\mathrm{FRInGe}}^{\mathrm{model}}}
\approx\frac{2T_{\mathrm B}}{T(M+2)}.
\end{align*}
At equal tracking-step counts, the leading classifier-work ratio of FRInGe to
FRInGe-B is therefore approximately $(M+2)/2$. These expressions do not
compare total runtimes: the methods can take different numbers of steps,
and their sequential steps and batching patterns differ.

For the spatially regularized binary direction, each step also needs two approximate
applications of $H^{-1}$, with $H$ defined above. If each application takes at
most $K_H$ iterations and applying the fixed spatial operator costs $C_L$,
the unbatched-equivalent spatial work is $O(2T_{\mathrm B}K_HC_L)$.
These fixed-operator iterations never evaluate the classifier.
The evaluated binary solver starts from zero at each waypoint, whereas
the categorical solver uses the previous direction as its initial guess.
Initialization and truncation can therefore affect the attained accuracy
even with equal iteration ceilings. Removing classifier-dependent Fisher
products does not by itself guarantee a wall-clock speedup over IG.

\subsection{Empirical Comparisons}
\label{app:fringeb_experiments}

The following studies serve different purposes. The pilot controls the
input cohort and operating point to compare measured quality and latency.
The full-cohort comparison assesses separately tuned pipelines over the
larger image collection. The gallery holds scalar settings fixed to
illustrate differences in spatial appearance. Their protocols should not
be merged into a single geometry-only or speed--quality claim.

\begin{center}
\small
\setlength{\tabcolsep}{4pt}
\begin{tabularx}{\linewidth}{@{}p{0.22\linewidth}p{0.23\linewidth}X@{}}
\toprule
Study & Images and architectures & Parameter and solver policy \\
\midrule
Paired quality--latency pilot & 64 images, Inception-v3 & Shared $\tau=0.05$, damping, spatial coefficients, and two-iteration ceilings; IG with 100/200/300 steps \\
Full-cohort benchmark comparison & Original cohort, six models; matched-target exclusions & Independently tuned methods and their retained numerical configurations \\
Matched-setting gallery & Three shared images, six models & FRInGe scalar settings for both methods; 20-iteration ceilings; no binary retuning \\
\bottomrule
\end{tabularx}
\end{center}

\subsubsection{Matched 64-Image Quality--Latency Pilot}
In Table ~\ref{tab:fringeb_controlled_runtime} we compare all methods in one run on the same 64 ImageNet images and top-1
targets using Inception-v3, timing attribution for one image at a time under
matched conditions. Both FRInGe variants use $\tau=0.05$, the same damping
and spatial regularization,
and a two-iteration ceiling for their respective solves; IG uses 100, 200, or
300 Gauss--Legendre integration steps. These settings define diagnostic operating points,
not the architecture-tuned configurations of the main benchmark.

\begin{table}[H]
\centering
\scriptsize
\setlength{\tabcolsep}{3.5pt}
\caption{Matched Inception-v3 runtime pilot ($n=64$). All methods were measured
in the same run. Latency is the median per-image attribution time with a 95\%
image-bootstrap interval, excluding initialization and metric evaluation.
Steps, Fisher products, and relative completeness error are means.
Steps denote IG integration steps or FRInGe tracking steps.
Completeness is measured with respect to each
method's own path endpoints. Speedup is the FRInGe median divided by each
method's median.}
\label{tab:fringeb_controlled_runtime}
\resizebox{\textwidth}{!}{%
\begin{tabular}{lccccc}
\toprule
Method & Steps & Fisher products & Latency (ms) [95\% CI] & Speedup vs FRInGe & Comp. error $\downarrow$ \\
\midrule
IG-100 & 100.0 & --- & $813.9\ [805.5,846.4]$ & 8.0$\times$ & 0.036 \\
IG-200 & 200.0 & --- & $1632.3\ [1606.1,1679.1]$ & 4.0$\times$ & 0.015 \\
IG-300 & 300.0 & --- & $2513.8\ [2431.2,2571.6]$ & 2.6$\times$ & 0.010 \\
FRInGe-B & 8.9 & 0.0 & $588.4\ [559.1,606.6]$ & 11.1$\times$ & 0.478 \\
FRInGe & 9.7 & 47.4 & $6529.1\ [6282.9,6774.5]$ & 1.0$\times$ & 0.055 \\
\bottomrule
\end{tabular}%
}
\end{table}

FRInGe-B has the lowest median latency in this pilot, consistent with
removing classifier-dependent Fisher products. However, it also has a
larger completeness residual than FRInGe and IG. The speed
advantage therefore does not establish equivalent path-integration
accuracy at the tested operating point. Step counts alone are not a
measure of computational cost, since the methods perform different work
per step.

\paragraph{Quality across IG integration budgets.}
Table~\ref{tab:fringeb_controlled_quality} tests whether increasing the
IG integration budget changes the quality comparison on the same images
and targets.

\begin{table}[H]
\centering
\footnotesize
\setlength{\tabcolsep}{1pt}
\renewcommand{\arraystretch}{1.15}
\caption{Paired attribution-quality pilot on the same 64 Inception-v3 images.
All methods come from the same run. Entries are means $\pm$ half-widths of
95\% image-bootstrap confidence intervals; the original percentile intervals
can be asymmetric. Bold and underlined entries have the best mean at the
displayed precision.}
\label{tab:fringeb_controlled_quality}
\begin{tabular*}{\textwidth}{@{\extracolsep{\fill}}lcccccc@{}}
\toprule
Method & MAS-Ins $\uparrow$ & MAS-Del $\downarrow$ & Ins-AUC $\uparrow$ & Del-AUC $\downarrow$ & Infidelity $\downarrow$ & Sparseness $\uparrow$ \\
\midrule
IG-100 & \statentry{0.692}{0.031} & \statentry{0.438}{0.065} & \statentry{0.651}{0.050} & \statentry{0.256}{0.051} & \statentry{0.0306}{0.0135} & \statentry{0.625}{0.012} \\
IG-200 & \statentry{0.693}{0.031} & \statentry{0.433}{0.062} & \statentry{0.653}{0.050} & \statentry{0.256}{0.051} & \statentry{0.0306}{0.0135} & \statentry{0.625}{0.012} \\
IG-300 & \statentry{0.691}{0.031} & \statentry{0.429}{0.062} & \statentry{0.652}{0.050} & \statentry{0.256}{0.051} & \statentry{0.0306}{0.0134} & \statentry{0.625}{0.012} \\
FRInGe-B & \best{\statentry{0.740}{0.029}} & \best{\statentry{0.370}{0.069}} & \best{\statentry{0.715}{0.049}} & \best{\statentry{0.213}{0.046}} & \best{\statentry{0.0202}{0.0118}} & \statentry{0.681}{0.012} \\
FRInGe & \statentry{0.721}{0.039} & \best{\statentry{0.370}{0.069}} & \statentry{0.649}{0.049} & \statentry{0.253}{0.046} & \best{\statentry{0.0202}{0.0119}} & \best{\statentry{0.718}{0.012}} \\
\bottomrule
\end{tabular*}
\end{table}

Paired bootstrap comparisons favor FRInGe-B over IG-300 on the reported
metrics. FRInGe is favored on MAS, Infidelity, and Sparseness,
while its Ins-AUC and Del-AUC differences are inconclusive. Comparisons
between IG step budgets also remain inconclusive. Together with the
small changes in IG means, this suggests that integration resolution
alone does not explain the observed method differences in this pilot.
The intervals are descriptive and unadjusted for multiple comparisons.

\paragraph{Direct FRInGe versus FRInGe-B comparison.}
Within-image comparisons between FRInGe-B and FRInGe on the same
64-image pilot yield paired intervals favoring FRInGe-B on both Ins-AUC and Del-AUC and
FRInGe on Sparseness; they include zero for both MAS scores and Infidelity.
Sparseness measures concentration rather than faithfulness. Thus, the observed
quality profile is mixed even when the two methods share the same regularizer,
step constraints, and two-iteration solver ceiling. Equal ceilings do not
imply equal direction accuracy, since the underlying linear systems and
initialization policies differ.

\subsubsection{Independently Tuned Architecture Comparisons}

Table~\ref{tab:fringeb_all_architectures} complements this controlled pilot
with the full-cohort benchmark results across all six architectures. Unlike the
pilot, these rows use the independently tuned configuration of each method;
they compare the two deployed pipelines and do not isolate the binary
predictive geometry. FRInGe-B has better insertion and deletion AUC sample
means on the four ResNets and Inception-v3, whereas FRInGe has both better
AUC means on VGG-19 and is consistently sparser. The MAS and Infidelity
comparisons are mixed. This architecture dependence rules out a blanket
quality-equivalence claim from the available results.

\begin{table}[H]
\centering
\scriptsize
\setlength{\tabcolsep}{0.8pt}
\renewcommand{\arraystretch}{1.08}
\caption{Full-cohort comparison of independently tuned FRInGe and FRInGe-B configurations. Entries are means $\pm$ half-widths of 95\% image-bootstrap
percentile intervals (2,000 resamples; seed 42). Bold and underlined entries
have the best mean within each architecture at the displayed precision,
without implying statistical superiority.}
\label{tab:fringeb_all_architectures}
\begin{tabular*}{\textwidth}{@{\extracolsep{\fill}}llcccccc@{}}
\toprule
Model & Method & MAS-Ins $\uparrow$ & MAS-Del $\downarrow$ & Ins-AUC $\uparrow$ & Del-AUC $\downarrow$ & Infidelity $\downarrow$ & Sparseness $\uparrow$ \\
\midrule
\multirow{2}{*}{ResNet-18}
 & FRInGe   & \statentry{0.598}{0.016} & \statentry{0.264}{0.010} & \statentry{0.492}{0.016} & \statentry{0.130}{0.008} & \best{\statentry{0.058}{0.008}} & \best{\statentry{0.706}{0.002}} \\
 & FRInGe-B & \best{\statentry{0.601}{0.014}} & \best{\statentry{0.259}{0.008}} & \best{\statentry{0.541}{0.015}} & \best{\statentry{0.095}{0.006}} & \statentry{0.059}{0.008} & \statentry{0.660}{0.003} \\
\addlinespace[1pt]
\multirow{2}{*}{ResNet-50}
 & FRInGe   & \best{\statentry{0.648}{0.014}} & \statentry{0.302}{0.012} & \statentry{0.581}{0.016} & \statentry{0.161}{0.009} & \best{\statentry{0.150}{0.050}} & \best{\statentry{0.702}{0.002}} \\
 & FRInGe-B & \statentry{0.635}{0.013} & \best{\statentry{0.295}{0.010}} & \best{\statentry{0.618}{0.015}} & \best{\statentry{0.122}{0.007}} & \statentry{0.151}{0.051} & \statentry{0.643}{0.003} \\
\addlinespace[1pt]
\multirow{2}{*}{ResNet-101}
 & FRInGe   & \best{\statentry{0.679}{0.014}} & \statentry{0.351}{0.015} & \statentry{0.641}{0.017} & \statentry{0.228}{0.012} & \best{\statentry{0.121}{0.022}} & \best{\statentry{0.706}{0.002}} \\
 & FRInGe-B & \statentry{0.660}{0.012} & \best{\statentry{0.342}{0.013}} & \best{\statentry{0.687}{0.015}} & \best{\statentry{0.165}{0.010}} & \statentry{0.123}{0.022} & \statentry{0.626}{0.003} \\
\addlinespace[1pt]
\multirow{2}{*}{ResNet-152}
 & FRInGe   & \best{\statentry{0.691}{0.013}} & \statentry{0.392}{0.017} & \statentry{0.681}{0.016} & \statentry{0.265}{0.013} & \best{\statentry{0.111}{0.020}} & \best{\statentry{0.703}{0.002}} \\
 & FRInGe-B & \statentry{0.669}{0.011} & \best{\statentry{0.375}{0.015}} & \best{\statentry{0.708}{0.015}} & \best{\statentry{0.205}{0.012}} & \statentry{0.113}{0.019} & \statentry{0.625}{0.003} \\
\addlinespace[1pt]
\multirow{2}{*}{Inception-v3}
 & FRInGe   & \best{\statentry{0.761}{0.009}} & \statentry{0.444}{0.020} & \statentry{0.709}{0.014} & \statentry{0.240}{0.012} & \best{\statentry{0.020}{0.005}} & \best{\statentry{0.723}{0.003}} \\
 & FRInGe-B & \statentry{0.743}{0.008} & \best{\statentry{0.416}{0.018}} & \best{\statentry{0.723}{0.013}} & \best{\statentry{0.208}{0.011}} & \statentry{0.021}{0.005} & \statentry{0.676}{0.003} \\
\addlinespace[1pt]
\multirow{2}{*}{VGG-19}
 & FRInGe   & \best{\statentry{0.588}{0.015}} & \best{\statentry{0.255}{0.008}} & \best{\statentry{0.503}{0.016}} & \best{\statentry{0.119}{0.007}} & \statentry{0.343}{0.064} & \best{\statentry{0.707}{0.003}} \\
 & FRInGe-B & \statentry{0.512}{0.015} & \statentry{0.281}{0.008} & \statentry{0.485}{0.015} & \statentry{0.130}{0.008} & \best{\statentry{0.341}{0.063}} & \statentry{0.672}{0.003} \\
\bottomrule
\end{tabular*}
\end{table}

\subsubsection{Qualitative Comparison under Matched Settings}

\begin{figure}[hp]
\centering
\includegraphics[width=\textwidth]{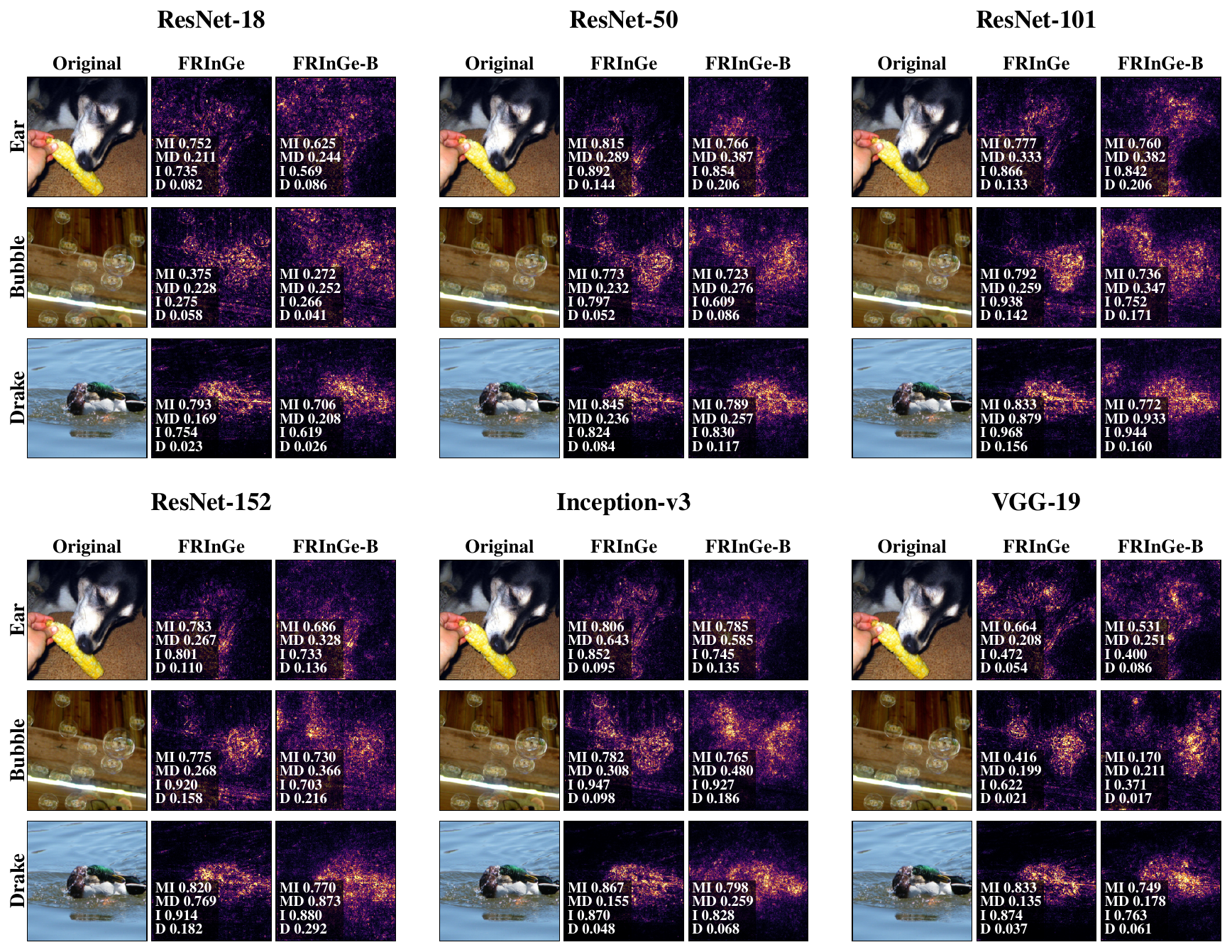}
\caption{FRInGe and FRInGe-B on six architectures using shared
FRInGe scalar settings and 20-iteration ceilings. Each model group
contains Original, FRInGe, and FRInGe-B columns, with ear, bubble, and
drake inputs in the same row order. All MAS scores are defined.
MI/MD denote MAS-Insertion/Deletion; I/D denote normalized insertion/deletion
AUC. Magnitude maps are independently normalized.}
\label{fig:fringeb_qualitative}
\end{figure}

Figure~\ref{fig:fringeb_qualitative} uses three shared examples,
ear, bubble, and drake, across all six architectures. The examples are
selected before attribution computation from a fixed shuffled ordering.
Eligibility requires the original target confidence to exceed both
evaluation-baseline confidences by more than $10^{-4}$ on every model.
No attribution map or method score enters selection. All displayed MAS
scores are therefore defined, although the gallery remains conditional
on baseline-confidence eligibility rather than representing a random
sample of the full benchmark.

Each pair explains the same original top-1 logit under identical
model-specific preprocessing. Both methods use the FRInGe scalar
coefficients in Table~\ref{tab:fringe_hparams}, including damping, the
two spatial coefficients, and step constraints, with a common
20-iteration solver ceiling. FRInGe-B is not retuned. This differs from
the two-iteration pilot and the main benchmark runs with a 25-iteration ceiling; it is a controlled illustration, not an exact
rerun of either quantitative study.

Metrics use the original signed tensors, while the displayed maps use
absolute-channel sums, 99th-percentile clipping, and independent
normalization, as in Appendix~\ref{app:qualitative}. Finite-budget solve
residuals are appreciable at these settings, so the panels compare the
evaluated numerical procedures, not isolated exact geometries. Three
shared images cannot establish a population-level quality ranking.

\subsection{Endpoint Diagnostics and Limitations}

FRInGe-B controls only the target probability. Even exact arrival at
$p_t(x_T)=1/C$ does not imply that the categorical prediction is uniform,
because the remaining probability mass may be concentrated on a small number
of non-target classes. The relevant binary endpoint quantities are
\begin{equation}
\left|p_t(x_T)-\frac1C\right|,
\qquad
\KL\!\left(\widetilde p_t(x_T)\,\|\,\widetilde q_t^\star\right),
\label{eq:fringeb_binary_endpoint}
\end{equation}
whereas the original categorical reference is assessed through
\begin{equation}
\KL\!\left(p(x_T)\,\|\,\tfrac1C\mathbf 1\right).
\label{eq:fringeb_full_endpoint}
\end{equation}
The first pair measures target neutralization; the latter measures proximity to
the original maximum-entropy reference. These quantities describe different
endpoint requirements; they are not interchangeable quality scores.
Realized-path completeness remains
$\sum_iA_i\approx F^t(x)-F^t(x_T)$ and is logically separate from either
endpoint criterion. Finally, for an arbitrary non-top target with
$p_t(x)<1/C$, the reference path increases rather than decreases its
probability. The present evaluation avoids this ambiguity by explaining the
top-1 predicted class.

\subsection{Scope and Modality-Specific Priors}

The spatial operator $L$ in Eq.~\eqref{eq:final_linear} was chosen for
the vision architectures evaluated here. It supplies an image-specific
prior; Gaussian smoothing is not required by the predictive Fisher--Rao
construction. Other modalities could use regularizers on suitable
continuous input representations while retaining a predictive reference
and geometric tracking. This separates the choice of input prior from
the choice of predictive geometry, including target-versus-rest
aggregation.

For example, an LLM extension could operate on token embeddings with a
language-appropriate prior and a specified predictive target. Such an
extension requires separate validation: embedding trajectories need not
correspond to discrete text, and the predictive target and reference
must be defined for the task. BlurIG's image-space Gaussian path has
no direct counterpart on discrete tokens, so adapting it would also
require a new corruption or interpolation path. The present image
experiments establish neither method's effectiveness in that setting.

\clearpage

\end{document}